\documentclass[10pt,twocolumn,letterpaper]{article}

\usepackage[pagenumbers]{wacv}
\usepackage[accsupp]{axessibility}  % Improves PDF readability for those with disabilities.
\usepackage{multirow}
\usepackage{array}
\usepackage{algorithm}
\usepackage{algpseudocode}

\definecolor{wacvblue}{rgb}{0.21,0.49,0.74}
\usepackage[pagebackref,breaklinks,colorlinks,allcolors=wacvblue]{hyperref}
\hypersetup{
  pdftitle={Beyond Output Faithfulness: Diagnosing and Mitigating Pathway Rerouting in Feature Attribution},
  pdfauthor={Siyu Zhang and Kenneth McMillan}
}

\def\wacvPaperID{3666} % *** Enter the WACV Paper ID here
\def\confName{WACV}
\def\confYear{2027}

\title{Beyond Output Faithfulness: Diagnosing and Mitigating Pathway Rerouting in Feature Attribution}

\author{Siyu Zhang \qquad Kenneth McMillan\\
The University of Texas at Austin\\
{\tt\small zhangsiyu@utexas.edu \qquad kenmcm@cs.utexas.edu}
}

\begin{document}
\maketitle
\begin{abstract}
    Insertion and deletion are standard metrics for evaluating feature 
    attributions by progressively adding or removing pixels. While they test 
    whether attributed features control a model output, they do not verify 
    whether they do so through the computation that supports the clean 
    prediction. We study this distinction with Ensemble Quantile Optimization (EQO), 
    a differentiable method that directly optimizes the insertion sequence. 
    On CNNs, Unconstrained EQO attains the highest insertion scores while departing 
    from clean representations and retaining its insertion score after channels
    supporting the clean prediction are ablated. Together, these results reveal 
    pathway rerouting that output-only metrics cannot detect. On a Vision 
    Transformer, target predictions from black inputs and attention deviation
    expose the same failure.
    To address this, we introduce ClipGrad, a layer-wise gradient
    filter that preserves clean activation trajectories during optimization.
    ClipGrad suppresses rerouting exploits while maintaining high output
    sensitivity across convolutional networks and Vision Transformers, and
    token intervention confirms this mitigation on the Transformer. These
    results show that output sensitivity is necessary but insufficient: a faithful
    attribution must control the output through internal computation used for the
    clean prediction.
\end{abstract}
    
\section{Introduction}
\label{sec:intro}

Feature attributions are commonly evaluated by \emph{insertion} and
\emph{deletion}: progressively adding pixels in the order prescribed by a map,
or removing them, and measuring the resulting change in target output
\cite{samek2016evaluating,petsiuk2018rise}. These tests answer a vital functional
question: does the ranked input evidence control the model's output? High
scores on these metrics are routinely taken as proof that the explanation
reflects the model's clean decision-making process.

This stronger conclusion does not follow automatically. A perturbed image can
recover the target prediction through alternate hidden features that were
irrelevant to the unperturbed input. The insertion score remains high even
though the evaluation trajectory has departed from the clean computation. We
call this divergence \emph{pathway rerouting}. It separates two questions about
explanation quality:
\begin{itemize}\itemsep1pt
  \item \textbf{Output faithfulness}, assessed here through \emph{output
  sensitivity}, measures whether attributed features control the prediction
  along an input perturbation trajectory.
  \item \textbf{Internal faithfulness} measures whether that control remains
  aligned with, and dependent on, the internal computational pathway that
  produced the original clean prediction.
\end{itemize}
Here, a \emph{computational pathway} refers to the sequence of intermediate
representations and feature detectors that support the clean prediction.

\begin{figure}[t]
\centering
\begin{subfigure}[b]{0.23\columnwidth}
    \includegraphics[width=\columnwidth]{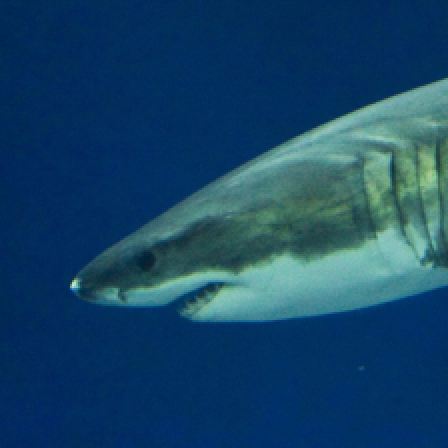}
    \caption{Input}
\end{subfigure}\hfill
\begin{subfigure}[b]{0.23\columnwidth}
    \includegraphics[width=\columnwidth]{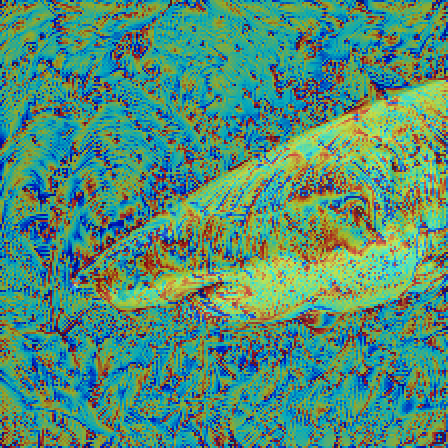}
    \caption{Unconstr.}
\end{subfigure}\hfill
\begin{subfigure}[b]{0.23\columnwidth}
    \includegraphics[width=\columnwidth]{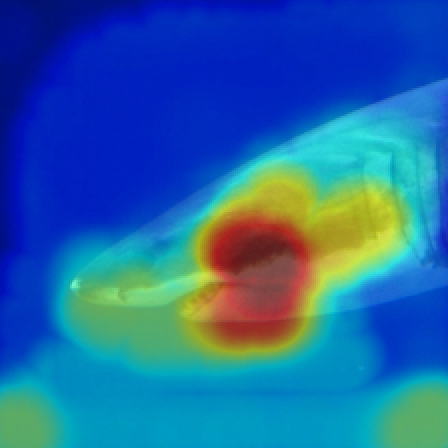}
    \caption{EP}
\end{subfigure}\hfill
\begin{subfigure}[b]{0.23\columnwidth}
    \includegraphics[width=\columnwidth]{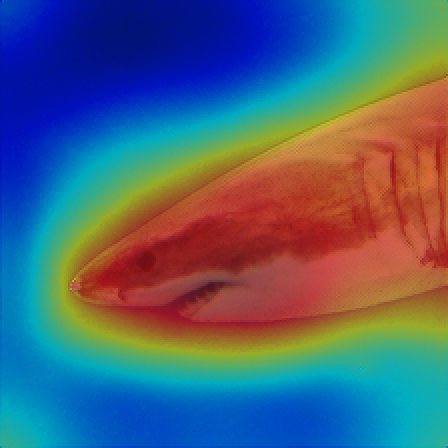}
    \caption{VM}
\end{subfigure}
\caption{\textbf{Equal outputs need not imply equal computational support.}
Unconstrained EQO achieves top insertion AUC with fragmented artifacts whose
score largely survives removal of clean-supporting features. ClipGrad (VM)
restores that dependence and produces stable, semantically grounded
attributions.}
\label{fig:motivation}
\end{figure}

We investigate this evaluation gap in three stages. First, we formulate
\emph{Ensemble Quantile Optimization} (EQO), a differentiable framework that
jointly optimizes the entire insertion curve via nested soft masks. As an
unconstrained diagnostic baseline, EQO achieves the highest insertion area under
the curve (AUC) across all evaluated architectures, yet produces fragmented,
noisy maps and induces target predictions from all-black inputs
(\cref{fig:motivation,tab:defense}). EQO directly demonstrates that output-only
metrics can reward misleading attributions.

Second, we formulate two complementary diagnostics of hidden computation to detect
rerouting. \emph{Activation-trajectory similarity} measures layer-wise geometric
alignment with clean internal representations. Complementing this, our
\emph{clean-support intervention probe} ($\rho$) measures how much of the
insertion score survives after greedily ablating channels that materially
support the clean prediction. Unconstrained EQO exhibits severe activation
deviation while largely preserving its insertion score after clean channels are
removed ($\rho \approx 0.79$--$0.82$, \cref{tab:activation,tab:rho}), showing
that its high score relies on alternate routes rather than the clean pathway.

Third, we propose \emph{ClipGrad}, a parameter-free, layer-wise gradient filter
that discourages EQO updates from moving intermediate activations away from
their clean values during optimization. Its Value Matching (VM) variant
prioritizes strict trajectory preservation and stability, while Inactivated
Binary Matching (IBM) maintains higher output sensitivity. Both variants
neutralize the rerouting exploit, restoring clean-pathway dependence and
yielding stable, coherent attribution maps. Together, they provide a
controllable tradeoff: VM most strongly preserves the clean pathway, while IBM
retains more of EQO's insertion gain (\cref{fig:frontier}).

The diagnostic separation holds across ImageNet and CUB-200-2011 with VGG16
and ResNet50. Focused ConvNeXt-Tiny and DeiT-Small studies further show how the
exploit and mitigation transfer to a GELU-based convolutional network and an
attention-based Transformer.

Our primary contributions are:
\begin{enumerate}\itemsep2pt
  \item \textbf{Pathway rerouting:} We introduce EQO to demonstrate
  that unconstrained optimization of output insertion metrics rewards
  pathway rerouting and induces severe black-input artifacts.
  \item \textbf{Internal diagnostics:} We propose a channel intervention probe
  ($\rho$) and a token-level analogue, paired with activation-trajectory
  metrics, to test internal computational faithfulness against the clean
  computation.
  \item \textbf{ClipGrad:} We develop
  ClipGrad to enforce layer-wise activation consistency during optimization,
  reconciling high functional output control with verified internal
  computational integrity.
\end{enumerate}

\section{Related Work}
\label{sec:related}

\paragraph{Attribution methods.}
Feature attribution methods fall broadly into gradient, representation, and
perturbation approaches. Gradient and relevance methods propagate sensitivity
or conservation scores from the output to input pixels
\cite{simonyan2013deep,sundararajan2017axiomatic,bach2015pixel}, with modified
backward paths and gradient aggregation designed to reduce noise and sharpen
visual localization
\cite{springenberg2014striving,smilkov2017smoothgrad,kapishnikov2021guided}.
Class Activation Mapping (CAM) methods aggregate class-discriminative feature
maps from intermediate convolutional representations
\cite{selvaraju2017grad,chattopadhay2018grad,wang2020score,jung2021towards}.
Perturbation and game-theoretic explanations estimate importance by measuring
output shifts under localized masking, randomized occlusion, Shapley values, or
optimized spatial masks
\cite{lundberg2017unified,petsiuk2018rise,fong2017interpretable,fong2019understanding,wagner2019interpretable}.
EQO extends this family to a nested sequence of retained-area levels. FGVis
also filters optimization gradients within clean activation bounds~\cite{wagner2019interpretable}; ClipGrad instead derives directional value-
and state-matching rules for that sequence.

\paragraph{Attribution evaluation and distribution shift.}
Insertion and deletion curves are the standard benchmarks for evaluating
attribution quality \cite{samek2016evaluating,petsiuk2018rise}, while parameter
randomization checks test whether maps depend on learned model weights
\cite{adebayo2018sanity}. Nieradzik \etal \cite{nieradzik2025reliable} show
that uncurated pixel removal can introduce out-of-distribution artifacts that
confound insertion/deletion rankings, and propose in-distribution adversarial
perturbations that make the evaluated inputs more realistic. We study what
happens after this input trajectory is chosen: whether high output scores are
produced by the same internal features that supported the clean prediction.
Our clean-support intervention measures this internal dependence.

\paragraph{Robustness of explanations.}
Attribution maps are known to be vulnerable to adversarial manipulation:
imperceptible input perturbations can dramatically alter saliency maps while
leaving the model's top-1 prediction unchanged
\cite{ghorbani2019interpretation,wang2020smoothed}. These studies show that an
\emph{explanation map} can change while the output label remains fixed. We
study a different failure: direct optimization of the evaluation metric can
achieve leading insertion scores through computational trajectories unsupported
by the clean input.

\paragraph{Counterfactual and contrastive explanations.}
Counterfactual and contrastive methods explain predictions relative to alternative
outcomes. Counterfactual Contrastive Explanations (CCE) generate content-aware
counterfactual inputs to answer ``why $P$ rather than $Q$?''
\cite{wang2023counterfactual}, while FG-VCE explains fine-grained
misclassifications using paired reference samples and Shapley-based saliency
partitions \cite{zhang2025towards}. These methods explain a prediction relative
to an alternative class. We instead ask whether features attributed to the
top-1 class preserve the computation that produced that prediction.

\paragraph{Internal representations and Transformer explainability.}
Mechanistic interpretability studies intermediate activations and learned
feature directions to understand neural computation
\cite{bau2017network,olah2020zoom,elhage2022toy,anthropic2023monosemanticity,cunningham2023sparse}. This
motivates our two internal diagnostics: activation-trajectory similarity
monitors intermediate state alignment, while clean-support ablation directly
probes computational reliance. For Vision Transformers, we benchmark against
gradient-, relevance-, and completeness-based methods including Transformer
Explainability, LeGrad, AttnLRP, and SLOC
\cite{chefer2021transformer,bousselham2025legrad,achtibat2024attnlrp,haddad2025soft},
extending our clean-support intervention from convolutional channels to patch
tokens.

\section{Optimizing Output Faithfulness with EQO}
\label{sec:method}

Insertion and deletion metrics evaluate whether the pixels highlighted by an
attribution map functionally control the model's output along a perturbation
trajectory \cite{samek2016evaluating,petsiuk2018rise}. They do not, however,
verify whether the perturbed input produces that output through the internal
computational pathway that governed the clean prediction. We test this gap by
directly optimizing the continuous insertion curve and then examining the
hidden computation that produces the score.

\subsection{Output-Based Evaluation}
Given an input image $x \in \mathbb{R}^{H\times W\times C}$, a frozen
classifier $f$, and a target class $y_t$, let $\phi(\tilde{x})$ and
$f_{y_t}(\tilde{x})$ denote the softmax probability and the logit of $y_t$ for an input $\tilde{x}$.
Unless stated otherwise, $y_t$ is the clean top-1 prediction on $x$. An
attribution map $M \in \mathbb{R}^{H\times W}$ assigns importance scores across
the pixel coordinate grid $\mathcal{P}$ ($|\mathcal{P}| = HW$). Let $u_q(M)$
and $l_q(M)$ denote the subsets containing the top and bottom $q$-fraction of
pixels ranked by $M$, respectively. For a baseline reference image $R$ and a
pixel replacement operator $x \otimes S$ that substitutes the pixel subset $S$
with $R$, the insertion and deletion Area Under the Curve (AUC) scores over
discrete quantile levels $\mathcal{Q}$ are defined as:
\begin{align}
V_{\mathrm{Ins}}(\phi, M, x)
  &= \frac{1}{|\mathcal{Q}|} \sum_{q \in \mathcal{Q}} \phi(x \otimes l_q(M)), \label{eq:ins_eval}\\
V_{\mathrm{Del}}(\phi, M, x)
  &= \frac{1}{|\mathcal{Q}|} \sum_{q \in \mathcal{Q}} \phi(x \otimes u_q(M)). \label{eq:del_eval}
\end{align}
Insertion progressively retains the highest-ranked pixels while replacing the
remaining bottom fraction with $R$, whereas deletion progressively removes the
highest-ranked pixels. Both metrics measure functional input--output
sensitivity along the evaluated perturbation trajectory. Neither
metric places any constraint on the internal hidden computation that produces
that output.

Here, $q$ denotes the fraction of the image replaced by the reference, meaning
each insertion step retains the complementary fraction $1-q$ of original
features. The choice of reference $R$, the masking operator, and the sampling
granularity jointly define the input trajectory evaluated by the metric. A high
insertion AUC demonstrates that the retained pixel subsets are sufficient to
drive target confidence along that specific trajectory. For example, localizing
a small but genuinely discriminative object part can represent a faithful
explanation even if the rest of the object is omitted. We study a different
failure: the optimized score remains high even
after the specific internal features that supported the clean prediction are
ablated, and high confidence can be induced entirely from content-free black
inputs.

\subsection{Unconstrained Output Optimization}
\label{sec:eqo}
To investigate what output-only optimization promotes, we directly optimize the
insertion curve. Ensemble Quantile Optimization (EQO) parameterizes the full
perturbation sequence via a collection of nested continuous retention maps
$\alpha_q \in [0,1]^{|\mathcal{P}|}$. Each $\alpha_q[p]$ interpolates between
the reference pixel at 0 and the original pixel at 1. For each replacement level
$q \in \mathcal{Q}$, EQO optimizes:
\begin{equation}
\begin{aligned}
\mathcal{L}_{\mathrm{ins}}(\alpha_q)
  &= -f_{y_t}\!\left(\alpha_q \odot x + (1 - \alpha_q) \odot R\right) \\
  &\quad + \beta \left|\sum_{p \in \mathcal{P}} \alpha_q[p] - (1 - q)|\mathcal{P}|\right|,
\end{aligned}
\label{eq:ins}
\end{equation}
where the second term penalizes deviations from the prescribed
retained area fraction $1-q$. To enforce monotonic spatial nesting
$l_{q_i} \subseteq l_{q_{i+1}}$, we parameterize non-negative increments
$\delta_{q_i} = \alpha_{q_i} - \alpha_{q_{i+1}}$ with boundary condition
$\alpha_{q_{k+1}} = \mathbf{0}$, and directly project gradient updates of
$\delta_{q_i}$ onto the feasible interval $[0, 1 - \alpha_{q_{i+1}}]$. This
guarantees that all masks remain strictly bounded in $[0,1]$ and nested
throughout optimization. The final attribution map is constructed by averaging
the optimized quantile sequence: $M = \frac{1}{k} \sum_{i=1}^k \alpha_{q_i}$.
The complete algorithmic workflow is presented in \cref{alg:fei}; optimization
hyperparameters are detailed in supplementary \cref{app:eqo_details}.

A single optimized mask covers only one retained-area level and does not define
a ranking over the full insertion curve. EQO instead optimizes several levels
as one nested sequence, placing every lower-retention mask inside the next larger one.
Averaging the sequence yields one pixel ranking for the standard discrete
insertion evaluator, keeping the differentiable objective separate from the
final evaluation. Ablations show that nested quantiles outperform direct area
regularization and that continuous masks provide a more effective optimization
surface than hard masks. Performance remains stable around the default
quantile count, penalty weight, and iteration budget (supplementary
\cref{app:ablations}).

We first analyze unconstrained EQO as a diagnostic baseline to characterize
the computational behavior rewarded by output metrics in the absence of
internal constraints. We refer to EQO integrated with our gradient filter as
\emph{Faithfulness-guided Ensemble Interpretation} (FEI); setting the
filter variant to \textsc{None} yields Unconstrained EQO. Its relation to the
target-score-only regimes of Extremal Perturbation and FGVis is detailed in
supplementary \cref{app:unconstrained_limit}.

\begin{algorithm}[t]
\caption{Faithfulness-guided Ensemble Interpretation (FEI)}
\label{alg:fei}
\begin{algorithmic}[1]
\Require Frozen model $f$, clean image $x$, reference $R$, quantiles $\mathcal{Q}=\{q_1<\cdots<q_k\}$, iterations $T$, step size $\eta$, area penalty $\beta$, ClipGrad variant
\Ensure Attribution map $M$
\State $\{h^\ell\}_{\ell=1}^L \gets \text{ForwardPass}(f, x)$
\State Initialize $\alpha_{q_{k+1}} \gets \mathbf{0}$, $\{\alpha_{q_i}\}_{i=1}^k \gets \mathbf{0}$, $\{\delta_{q_i}\}_{i=1}^k \gets \mathbf{0}$
\For{$i = k$ \textbf{downto} $1$}
  \For{$t = 1$ \textbf{to} $T$}
    \State $\alpha_{q_i} \gets \alpha_{q_{i+1}} + \delta_{q_i}$
    \State $\tilde{x} \gets \alpha_{q_i} \odot x + (1 - \alpha_{q_i}) \odot R$
    \State $\{\tilde{h}^{\ell}\}_{\ell=1}^L \gets \text{ForwardPass}(f, \tilde{x})$
    \State Compute $\mathcal{L}_{\text{ins}}(\alpha_{q_i})$ \Comment{\cref{eq:ins}}
    \For{$\ell = L$ \textbf{downto} $1$}
      \State $\gamma_h^\ell \gets \partial\mathcal{L}_{\text{ins}} / \partial\tilde{h}^\ell$
      \State $\tilde{\gamma}_h^\ell \gets \textsc{ClipGrad}(\gamma_h^\ell, h^\ell, \tilde{h}^\ell, \text{variant})$
    \EndFor
    \State $g_\delta \gets \text{Backpropagate } \{\tilde{\gamma}_h^\ell\} \text{ to } \delta_{q_i}$
    \State $\delta_{q_i} \gets \mathrm{clip}(\delta_{q_i} - \eta g_\delta,\, 0,\, 1 - \alpha_{q_{i+1}})$
  \EndFor
\EndFor
\State \Return $M = \frac{1}{k} \sum_{i=1}^k \alpha_{q_i}$
\end{algorithmic}
\end{algorithm}

\section{Diagnosing Internal Faithfulness}
\label{sec:diagnosis}

A model may base its prediction on object parts, background context, or visual
shortcuts that do not align with human annotations. We therefore evaluate
internal faithfulness against the model's own computation on the clean image.
Since an attribution explains that prediction, departing from the clean
computation is harmful when the output no longer depends on the features that
produced it, which $\rho$ tests, or when target evidence arises without image
content, which the black-input test (\cref{sec:defense}) detects.
We use two complementary diagnostics. \emph{Activation-trajectory similarity}
asks whether the perturbed image produces hidden representations similar to
those of the clean image. The \emph{clean-support intervention} asks whether
the insertion score still depends on the internal features that supported the
clean prediction.

\subsection{Dependence on Clean-Channel Support ($\rho$)}
\label{sec:causal}
For each clean input image $x$, we record the baseline target probability
$p_0 = \phi(x)$ and extract intermediate feature representations at the
penultimate feature block (e.g., VGG16 \texttt{features.29} with 512 channels;
ResNet50 \texttt{layer4} with 2048 channels). We then greedily construct an
image-specific clean-supporting channel set $\mathcal{S}_{\text{clean}}$ by
iteratively zeroing feature channels, at each step identifying the remaining
channel whose removal causes the steepest drop in clean target probability:
\begin{equation}
c^\star = \arg\min_{c \notin \mathcal{S}_{\text{clean}}} \mathrm{softmax}(f(x;\, \mathcal{S}_{\text{clean}} \cup \{c\} \text{ ablated}))_{y_t},
\label{eq:greedy_channel}
\end{equation}
repeating this selection until the resulting target probability $p_S$ falls below $\tau p_0$
with default threshold $\tau = 0.5$, or until no channels remain.

The penultimate feature block serves as the default measurement layer
because it represents the last high-dimensional, channel-structured layer
directly preceding the classification head. Sensitivity analysis across earlier
feature blocks and multiple stopping thresholds establishes consistency across
layers and intervention strengths (supplementary \cref{tab:rho_sensitivity,app:pathway}).

To determine whether an attribution map $M$ depends on the selected
clean-supporting channels, we zero-ablate the same set $\mathcal{S}_{\text{clean}}$
throughout the discrete insertion sweep and compute the clean-pathway
independence ratio:
\begin{equation}
\rho(M) = \frac{A_{-S}(M)}{A(M)},
\label{eq:rho}
\end{equation}
where $A(M)$ is the intact insertion AUC score and $A_{-S}(M)$ is the insertion
AUC measured with $\mathcal{S}_{\text{clean}}$ ablated at every evaluation step.
A high value of $\rho$ indicates that the attribution trajectory largely
preserves its output score after removing channels that support the clean
prediction, signaling pathway rerouting; a low value confirms that the
attribution remains dependent on those clean-supporting channels. We use zero
ablation because mean ablation can inject feature-like signals and produce
ratios above one that are difficult to interpret. Random-channel controls, threshold
sensitivity grids, and greedy selection details are reported in supplementary
\cref{app:pathway}.

The denominator in \cref{eq:rho} normalizes by the attribution map's own intact
insertion score, so $\rho$ measures the fraction of that map's functional output
sensitivity that survives removal of the selected clean-supporting pathway. For
example, $\rho=0.4$ means that 40\% of the map's original insertion score
remains after the intervention, whereas a value near one means that removing
the clean-supporting channels has little effect. We therefore report intact
insertion AUC alongside $\rho$.

The set $\mathcal{S}_{\text{clean}}$ is constructed from the original image
before any attribution method is evaluated, and the same set is used for every
method.

\subsection{Activation-Trajectory Similarity}
For layer-wise deterministic transformations:
\begin{equation}
h^{\ell+1} = f_\ell(h^\ell;\, \theta_\ell),
\label{eq:dynamics}
\end{equation}
where $\theta_\ell$ denotes the frozen parameters of layer $\ell$, we compare
the clean intermediate activation $h^\ell$ with the
activation $\tilde{h}^\ell$ generated along the insertion perturbation
trajectory. We report cosine similarity to compare feature direction and mean
squared error (MSE) to compare absolute activation values. Neural features are
represented as directions in hidden space
\cite{olah2020zoom,cunningham2023sparse,elhage2022toy}; the two measures capture
complementary changes in which directions are active and how strongly they are
expressed. Activation similarity measures the state
that ClipGrad directly constrains, while the independently constructed $\rho$
intervention tests which features support the output. We diagnose pathway rerouting
when high insertion coincides with large activation deviation, high $\rho$,
and target induction from an all-black input (\cref{sec:defense}). Spatial
coherence provides an additional check on map stability.

\section{Pathway-Constrained EQO with ClipGrad}
\label{sec:clipgrad}

ClipGrad addresses this failure by constraining EQO's optimization path. During the backward pass,
\emph{ClipGrad} filters each intermediate gradient coordinate. If an update
would move a perturbed activation $\tilde{h}^\ell$ farther from its clean value
$h^\ell$, ClipGrad sets that gradient coordinate to zero; the remaining
coordinates continue to optimize insertion. The backbone weights remain
frozen throughout.

Let $\gamma_h = \partial\mathcal{L}_{\mathrm{ins}} / \partial\tilde{h}^\ell$
denote the gradient of the insertion loss with respect to intermediate
activations at layer $\ell$. Derived from a saliency-weighted activation
matching objective (supplementary \cref{app:clipgrad_details}), our primary variant,
\textbf{Value Matching (VM)}, enforces strict bidirectional preservation by
suppressing gradients that push $\tilde{h}^\ell[j]$ away from $h^\ell[j]$ in
either direction:
\begin{equation}
\widetilde{\gamma}_h[j] =
\begin{cases}
0 & (\gamma_h[j] \le 0 \land \tilde{h}^\ell[j] > h^\ell[j]) \\
  & \quad \lor (\gamma_h[j] > 0 \land \tilde{h}^\ell[j] \le h^\ell[j]), \\
\gamma_h[j] & \text{otherwise.}
\end{cases}
\label{eq:vm}
\end{equation}
Under gradient descent, a negative gradient raises the corresponding
activation and a positive gradient lowers it. VM blocks either update when it
would move $\tilde{h}^\ell[j]$ farther from $h^\ell[j]$. It changes only the
backward optimization direction, without clamping forward activations or
modifying the frozen model.

\textbf{Inactivated Binary Matching (IBM)} is a relaxed alternative that
suppresses increases in neurons that were inactive on the clean input:
\begin{equation}
\widetilde{\gamma}_h[j] =
\begin{cases}
0 & \gamma_h[j] \le 0 \land h^\ell[j] \le 0, \\
\gamma_h[j] & \text{otherwise.}
\end{cases}
\label{eq:ibm}
\end{equation}
Thus, IBM blocks only updates that would increase a unit whose activation on
the clean image is non-positive; all other directions remain unchanged.
For smooth activations such as GELU, IBM treats positive outputs as active and
non-positive outputs as inactive.

VM applies the stricter two-sided rule and most strongly preserves the clean
activation trajectory. IBM relaxes the constraint and therefore retains more
of EQO's insertion gain. IVM and AVM apply the one-sided value rules defined in
supplementary \cref{app:clipgrad_details} and test how constraint direction
interacts with architecture.

% Declared early so the full-width qualitative comparison can use the first
% available page top rather than queueing behind the main benchmark table.
\begin{figure*}[t]
  \centering
  \setlength{\tabcolsep}{2pt}
  \scriptsize
  \begin{tabular}{cccccc}
  Input & FEI$_{\mathrm{IBM}}$ & FEI$_{\mathrm{VM}}$ & Unconstr. EQO & EP & Lift-CAM \\
  \includegraphics[width=.15\linewidth]{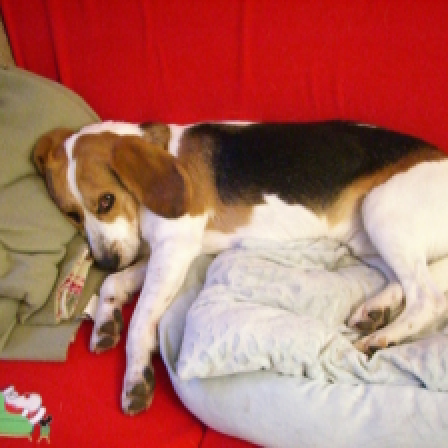} &
  \includegraphics[width=.15\linewidth]{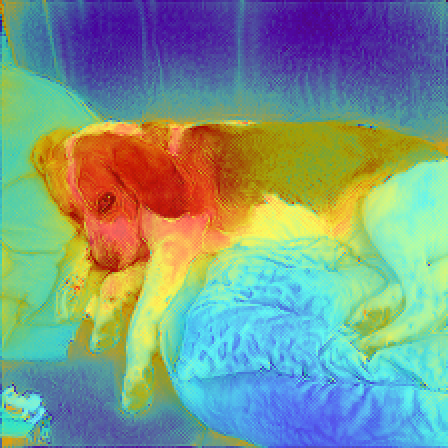} &
  \includegraphics[width=.15\linewidth]{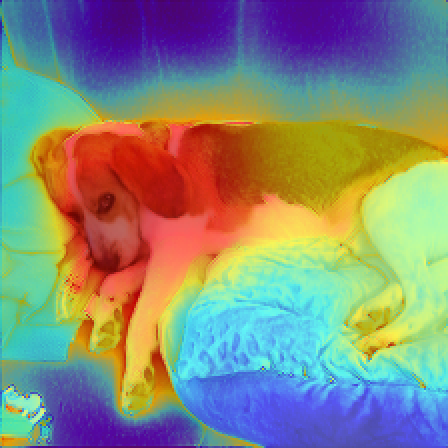} &
  \includegraphics[width=.15\linewidth]{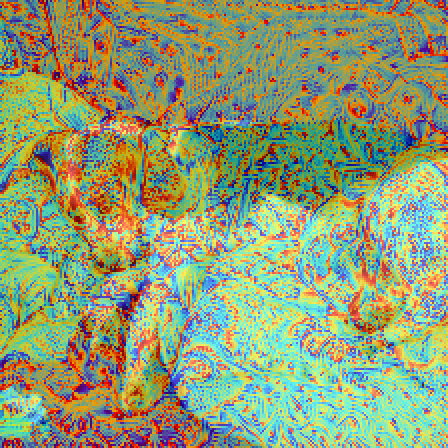} &
  \includegraphics[width=.15\linewidth]{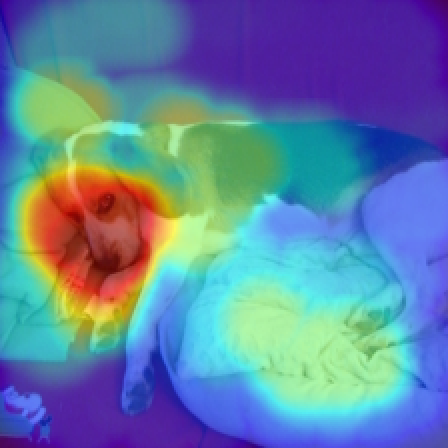} &
  \includegraphics[width=.15\linewidth]{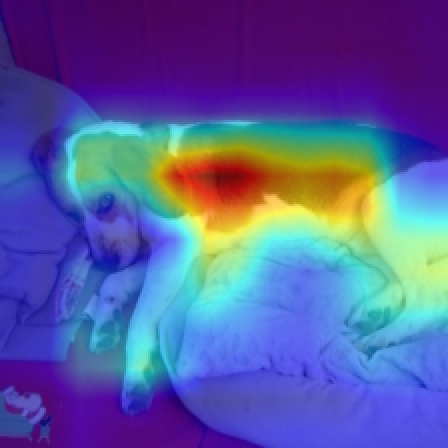} \\
  \includegraphics[width=.15\linewidth]{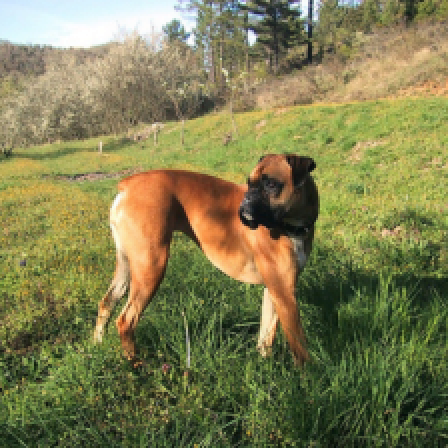} &
  \includegraphics[width=.15\linewidth]{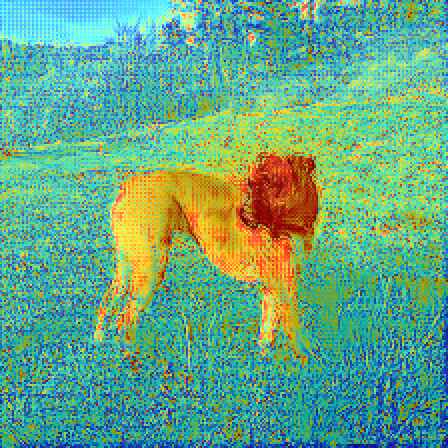} &
  \includegraphics[width=.15\linewidth]{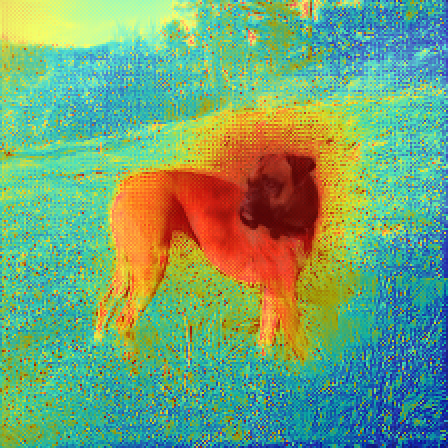} &
  \includegraphics[width=.15\linewidth]{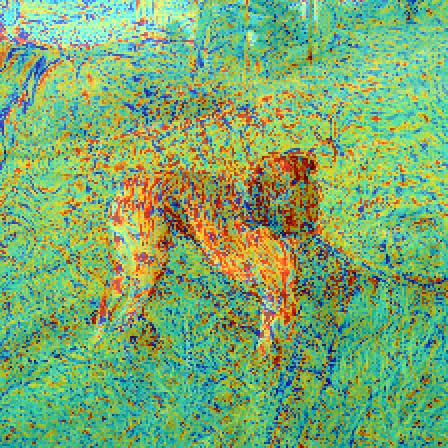} &
  \includegraphics[width=.15\linewidth]{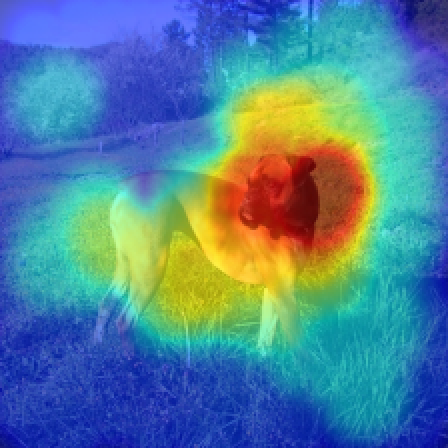} &
  \includegraphics[width=.15\linewidth]{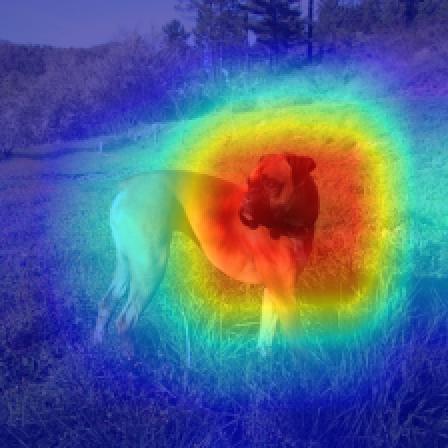}
  \end{tabular}
  \caption{Qualitative ImageNet comparison on VGG16 (top) and ResNet50
  (bottom). Unconstrained EQO produces fragmented maps, while the constrained
  FEI variants remain spatially coherent.}
  \label{fig:qualitative_main}
  \end{figure*}
  
  \section{Experiments}
  \label{sec:experiments}
  
  We organize the experiments around three questions:
  \begin{itemize}\itemsep2pt
    \item \textbf{RQ1:} Can direct optimization achieve high insertion scores by
    rerouting computation away from the clean prediction pathway?
    \item \textbf{RQ2:} Do activation similarity and clean-support intervention
    identify the same failure?
    \item \textbf{RQ3:} Can ClipGrad prevent rerouting while retaining strong
    output sensitivity across architectures?
  \end{itemize}
  Additional experiments test consistency across references and datasets and
  transfer to ConvNeXt and Vision Transformers.
  
  \subsection{Experimental Setup}
  \label{sec:setup}
  
  \paragraph{Architectures and Datasets.} We evaluate VGG16 \cite{simonyan2014very} and
  ResNet50 \cite{he2016deep} across ImageNet \cite{Imagenet} and CUB-200-2011 \cite{wah2011caltech},
  representing broad categorization and fine-grained visual classification, respectively.
  To evaluate architectural transfer beyond convolutional networks, we further benchmark
  DeiT-Small \cite{touvron2021training} as a representative Vision Transformer.
  
  Ensemble Quantile Optimization (EQO) uses Adam \cite{kingma2014adam} for 100
  iterations at each quantile replacement fraction in
  $\mathcal{Q} = \{0.1, 0.3, 0.5, 0.7, 0.9\}$. The initial learning rate is
  $\eta = 0.1$, and the area-constraint penalty is $\beta = 0.1$.
  For the primary CNN benchmarks, ClipGrad is applied across all intermediate
  convolutional layers in the backbone. The DeiT-Small transfer study evaluates only the primary VM variant,
  applied to the outputs of the attention modules in all 12 blocks.
  
  In our primary benchmark, FEI optimization resamples
  a random solid-color RGB reference at each gradient step to prevent the optimizer from
  overfitting to a single static background, whereas the standard discrete insertion and deletion
  metrics evaluate the resulting maps using a constant zero reference tensor. Main benchmark metrics are computed over
  $n = 3{,}000$ validation images per architecture; targeted sensitivity analyses specify exact
  sample sizes in context. All implementation details, hardware specifications,
  and ablation studies are detailed in supplementary \cref{app:eqo_details,app:runtime,app:ablations}.
  On one RTX~3090, FEI$_{\mathrm{VM}}$ takes 1.58\,s per map, versus 6.34\,s
  for RISE and 33.50\,s for Extremal Perturbation (supplementary \cref{tab:runtime}).
  
  \paragraph{Controlled Benchmark Protocol.} The comparison spans gradient,
  relevance, CAM, perturbation, and counterfactual methods. Within each condition,
  we recompute all maps with the same images, clean top-1 targets, resolution,
  perturbation schedule, and reference. The complete protocol appears in
  supplementary \cref{app:baseline_protocol}.
  
  \subsection{RQ1: Output Optimization Exposes the Gap}
  \label{sec:defense}
  
  \paragraph{Direct Optimization Exposes Pathway Rerouting.} Across both datasets and
 backbones, Unconstrained EQO attains the highest insertion AUC in our study
  (\cref{tab:external}). When generation and evaluation share a zero or a blur
  reference, it still exceeds VM by 0.194 and 0.260 (supplementary
  \cref{tab:reference_ci}). Following the black-input test of
  FGVis~\cite{wagner2019interpretable}, we evaluate all-black-input target
  induction (complete rule in supplementary \cref{app:black}). Unconstrained EQO
  induces the requested category for 92.8\%--100\% of CNN targets and
  99.5\% of DeiT-Small targets (\cref{tab:defense}). The canvas contains no
  evidence from a clean image, so these predictions show that unconstrained
  optimization can create target-driving patterns through the perturbation
  process itself. High insertion therefore does not establish that an
  attribution reflects the computation used for the clean image. Cascading
  parameter randomization confirms that FEI's maps depend on learned weights
  (supplementary \cref{sec:sanity}).
  
  Deletion gives a complementary result. Although Unconstrained EQO has the
  highest insertion AUC in every setting, its deletion AUC is also higher than
  every ClipGrad variant and every comparison method, with one tie (Extremal
  Perturbation on CUB/VGG16; \cref{tab:external}). Optimizing insertion alone
  therefore does not produce a ranking that also removes decisive evidence early;
  ClipGrad improves deletion while retaining strong insertion.
  
  \begin{table*}[t]
  \caption{Insertion and deletion performance (mean AUC; insertion $\uparrow$,
  deletion $\downarrow$). Bold and underline mark the best and second-best
  standard or constrained methods. The final row reports Unconstrained EQO as an
  output-optimized diagnostic.}
  \label{tab:external}
  \centering\small\setlength{\tabcolsep}{1.8pt}
  \begin{tabular}{c|cc|cc|cc|cc}
  \toprule
  & \multicolumn{4}{c|}{CUB-200-2011} & \multicolumn{4}{c}{ImageNet}\\
  \cmidrule(lr){2-5}\cmidrule(lr){6-9}
  & \multicolumn{2}{c|}{VGG16} & \multicolumn{2}{c|}{ResNet50}
  & \multicolumn{2}{c|}{VGG16} & \multicolumn{2}{c}{ResNet50}\\
  \cmidrule(lr){2-3}\cmidrule(lr){4-5}\cmidrule(lr){6-7}\cmidrule(lr){8-9}
  & Ins. $\uparrow$ & Del. $\downarrow$ & Ins. $\uparrow$ & Del. $\downarrow$ & Ins. $\uparrow$ & Del. $\downarrow$ & Ins. $\uparrow$ & Del. $\downarrow$\\
  \midrule
  FG-VCE~\cite{zhang2025towards}& \underline{0.738} & 0.094 & 0.615 & 0.106 & 0.487 & 0.114 & 0.566 & 0.149\\
  Grad-CAM~\cite{selvaraju2017grad} & 0.728 & 0.079 & 0.612 & 0.098 & 0.453 & 0.102 & 0.563 & 0.139\\
  Extremal Perturbation~\cite{fong2019understanding} & 0.727 & 0.095 & 0.595 & 0.098 & 0.473 & 0.130 & 0.516 & 0.149\\
  Lift-CAM~\cite{jung2021towards} & 0.724 & 0.077 & 0.610 & 0.097 & 0.483 & 0.097 & 0.554 & 0.141\\
  Grad-CAM++~\cite{chattopadhay2018grad} & 0.718 & 0.093 & 0.610 & 0.096 & 0.477 & 0.119 & 0.553 & 0.142\\
  Score-CAM~\cite{wang2020score}& 0.716 & 0.082 & 0.611 & 0.094 & 0.486 & 0.101 & 0.554 & 0.145\\
  IIA~\cite{barkan2023visual} & 0.709 & 0.091 & 0.607 & 0.098 & 0.437 & 0.109 & 0.549 & 0.134\\
  RISE~\cite{petsiuk2018rise} & 0.697 & 0.082 & 0.583 & 0.085 & 0.431 & 0.103 & 0.480 & 0.125\\
  Saliency~\cite{simonyan2013deep} & 0.498 & \underline{0.043} & 0.339 & 0.082 & 0.210 & 0.076 & 0.257 & 0.125\\
  SmoothGrad~\cite{smilkov2017smoothgrad} & 0.638 & 0.047 & 0.514 & 0.060 & 0.339 & \underline{0.075} & 0.395 & 0.117\\
  Integrated Gradients~\cite{sundararajan2017axiomatic} & 0.097 & \textbf{0.033} & 0.118 & \underline{0.050} & 0.098 & \textbf{0.039} & 0.154 & \textbf{0.065}\\
  \midrule
  FEI$_{\mathrm{IBM}}$ & \textbf{0.753} & 0.051 & \textbf{0.647} & \textbf{0.044} & \textbf{0.566} & 0.081 & \textbf{0.626} & \underline{0.093}\\
  FEI$_{\mathrm{AVM}}$ & 0.732 & 0.055 & \underline{0.636} & 0.054 & \underline{0.557} & 0.080 & \underline{0.601} & 0.113\\
  FEI$_{\mathrm{VM}}$ & 0.724 & 0.072 & 0.634 & 0.054 & 0.512 & 0.117 & 0.590 & 0.121\\
  FEI$_{\mathrm{IVM}}$ & 0.713 & 0.079 & 0.615 & 0.059 & 0.489 & 0.130 & 0.561 & 0.135\\
  \midrule
  Unconstrained EQO & 0.778 & 0.095 & 0.772 & 0.112 & 0.729 & 0.185 & 0.849 & 0.207\\
  \bottomrule
  \end{tabular}
  \end{table*}
  
  \begin{figure*}[t]
  \centering
  \begin{subfigure}[t]{0.485\textwidth}
  \centering
  \includegraphics[width=\linewidth]{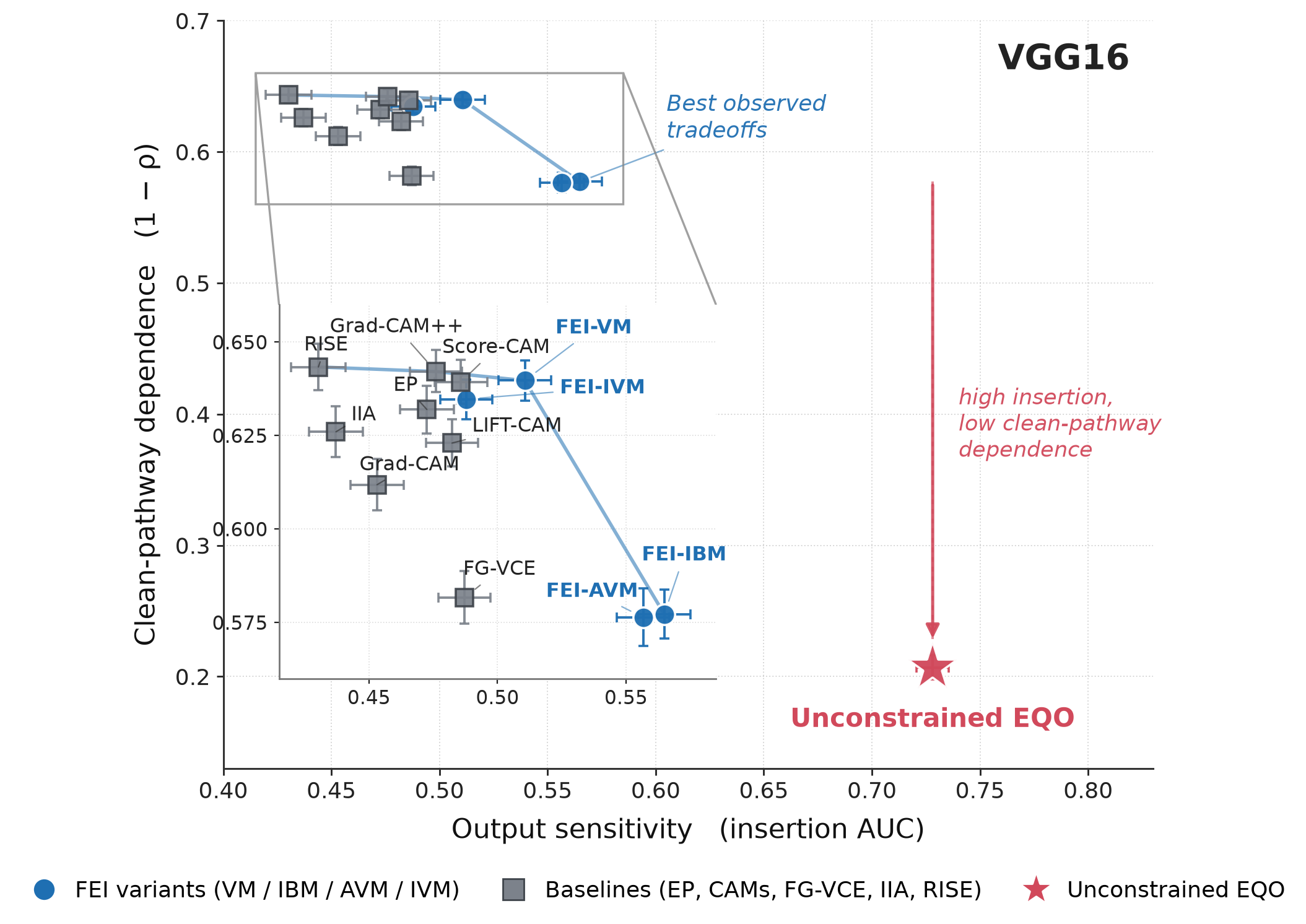}
  \caption{VGG16}
  \label{fig:frontier-vgg16}
  \end{subfigure}
  \hfill
  \begin{subfigure}[t]{0.485\textwidth}
  \centering
  \includegraphics[width=\linewidth]{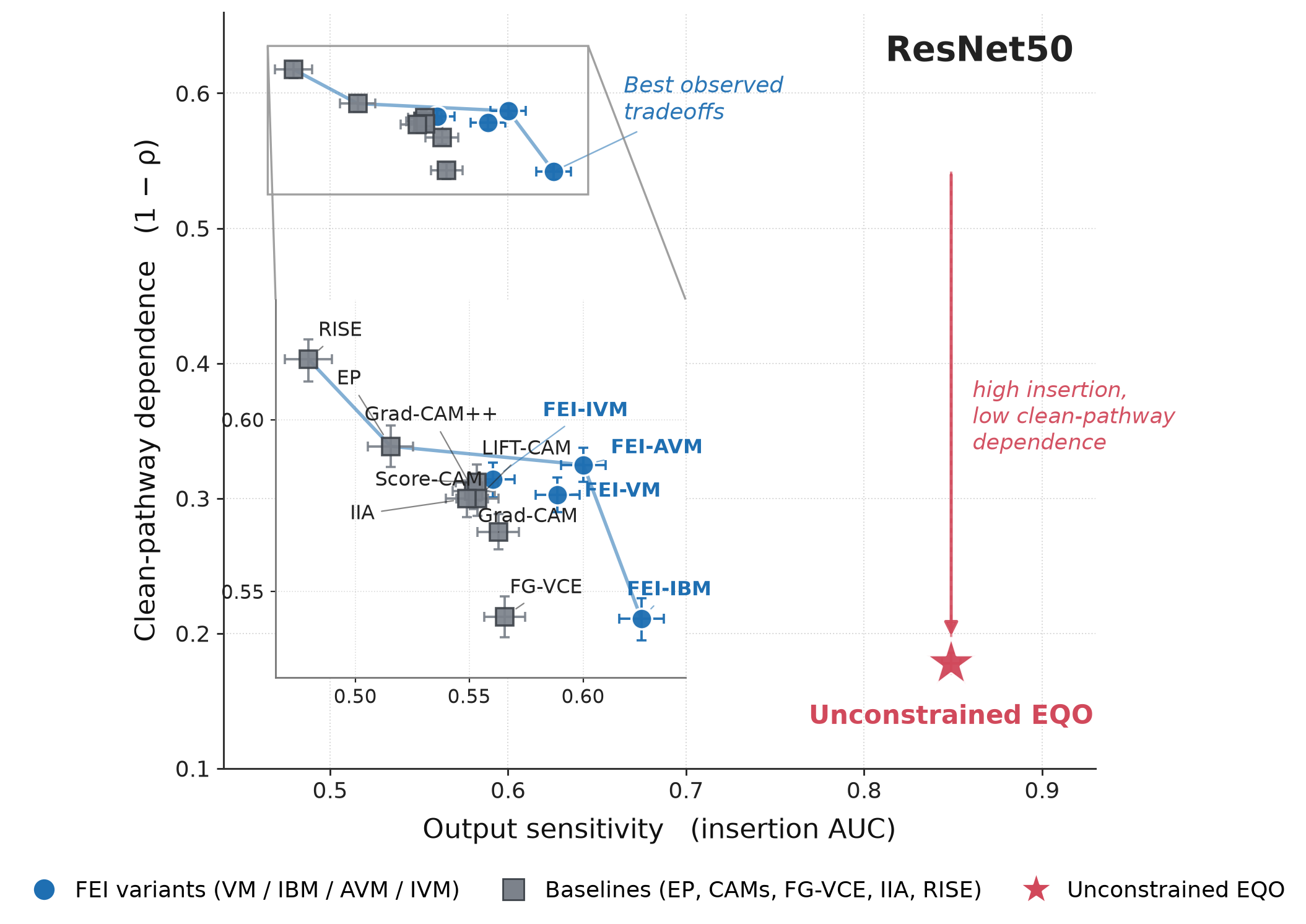}
  \caption{ResNet50}
  \label{fig:frontier-resnet50}
  \end{subfigure}
  \caption{ImageNet tradeoff between insertion AUC and clean-pathway dependence
  ($1-\rho$); higher values on both axes are better. Across both backbones,
  ClipGrad restores clean-pathway dependence while retaining part of
  Unconstrained EQO's insertion gain.}
  \label{fig:frontier}
  \end{figure*}
  
\begin{table}[t]
\caption{All-black-input target-induction rate (\%; lower is better); dashes
mean not evaluated. FGVis introduced this check~\cite{wagner2019interpretable};
$^\dagger$ applies its activation-bound gradient rule in our EQO optimizer,
which produces RGB attributions. Protocol in supplementary \cref{app:black}.}
  \label{tab:defense}
  \centering\small\setlength{\tabcolsep}{2pt}
  \begin{tabular}{lcccccc}
  \toprule
Model & FGVis$^\dagger$ & IBM & IVM & AVM & VM & Unconstr.\\
  \midrule
  AlexNet~\cite{krizhevsky2012imagenet}    & 0.1\% & 0.4\% & 0.2\% & 93.8\% & 0.1\% & 92.8\%\\
  ResNet50   & 0.1\% & 0.1\% & 0\%   & 0.1\%  & 0.1\% & 97.1\%\\
  VGG16      & 0.1\% & 0.1\% & 0.1\% & 95.2\% & 0.1\% & 94.7\%\\
  GoogLeNet~\cite{szegedy2015going}  & 0.1\% & 0\%   & 0.2\% & 0\%    & 0\%   & 97.9\%\\
  DeiT-Small & --    & --    & --    & --     & 0.1\% & 99.5\%\\
  ConvNeXt-T & --    & 0.4\% & 0.3\% & 20.7\% & 0.1\% & 100\%\\
  \bottomrule
  \end{tabular}
  \end{table}
  
  \subsection{RQ2: Complementary Internal Diagnostics}
  \label{sec:internal_evidence}

  \paragraph{Dependence on Clean-Channel Support.} On ImageNet, unconstrained EQO has
  $\rho = 0.793$ on VGG16 and $0.822$ on ResNet50, versus $0.357$--$0.424$ and
  $0.382$--$0.458$ for every other method in \cref{tab:rho}. A size-matched
  random-channel control leaves AUC nearly unchanged for all methods ($\rho \approx 0.995$;
  supplementary \cref{tab:rho_random_control}), showing that the drop comes
  from removing clean-supporting channels rather than from zeroing the same
  number of arbitrary channels. The complete layer and threshold analysis appears
  in supplementary \cref{tab:rho_sensitivity}.
  
  \begin{table}[t]
  \caption{ImageNet pathway independence $\rho$ ($n=3000$; $\uparrow$ worse).
  Saliency, SmoothGrad, and Integrated Gradients (low intact insertion) are
  listed in supplementary \cref{tab:rho_ci}.}
  \label{tab:rho}
  \centering\small\setlength{\tabcolsep}{6pt}
  \begin{tabular}{lcc}
  \toprule
  Method & VGG16 & ResNet50\\
  \midrule
  Grad-CAM & .388 & .433\\
  Grad-CAM++ & .358 & .421\\
  Score-CAM & .361 & .418\\
  Lift-CAM & .377 & .423\\
  RISE & .357 & .382\\
  Extremal Perturbation & .368 & .408\\
  IIA & .374 & .423\\
  FG-VCE & .418 & .457\\
  \midrule
  FEI$_{\mathrm{VM}}$ & .360 & .422\\
  FEI$_{\mathrm{IBM}}$ & .423 & .458\\
  FEI$_{\mathrm{AVM}}$ & .424 & .413\\
  FEI$_{\mathrm{IVM}}$ & .365 & .417\\
  \midrule
  Unconstrained EQO & .793 & .822\\
  \bottomrule
  \end{tabular}
  \end{table}
  
  \paragraph{Activation-Trajectory Similarity.} At the final feature block, unconstrained EQO has
  substantially larger MSE and lower cosine similarity to the clean trajectory than the
  constrained variants and comparison methods (\cref{tab:activation}). VM gives the closest
  VGG16 match, while the constrained variants remain close on ResNet50. This result describes
  the representation change; the clean-support intervention separately shows that the changed
  trajectory is less dependent on features supporting the clean prediction. Layerwise results
  are in supplementary \cref{app:activation}.
  
  \begin{table}[t]
  \caption{ImageNet final-block activation match (MSE $\downarrow$, cos. $\uparrow$).
  All methods: supplementary \cref{tab:supp_layerwise_resnet,tab:supp_layerwise_vgg}.}
  \label{tab:activation}
  \centering\small\setlength{\tabcolsep}{2.5pt}
  \begin{tabular}{lcccc}
  \toprule
  & \multicolumn{2}{c}{MSE $\downarrow$} &
  \multicolumn{2}{c}{Cos. $\uparrow$}\\
  \cmidrule(lr){2-3}\cmidrule(lr){4-5}
  Method & VGG16 & RN50 & VGG16 & RN50\\
  \midrule
  Grad-CAM++ & .573 & .495 & .640 & .708\\
  Lift-CAM & .648 & .500 & .594 & .704\\
  RISE & .731 & .588 & .561 & .654\\
  Extremal Perturbation & .671 & .577 & .583 & .661\\
  Integrated Gradients & 1.173 & .817 & .260 & .485\\
  \midrule
  FEI$_{\mathrm{VM}}$ & \textbf{.545} & .412 & \textbf{.688} & .758\\
  FEI$_{\mathrm{IBM}}$ & .629 & .415 & .658 & .757\\
  FEI$_{\mathrm{AVM}}$ & .952 & \textbf{.399} & .628 & \textbf{.766}\\
  \midrule
  Unconstrained EQO & 3.619 & 1.494 & .344 & .592\\
  \bottomrule
  \end{tabular}
  \end{table}
  
  \paragraph{Qualitative Comparison.} In \cref{fig:qualitative_main},
  Unconstrained EQO breaks attribution mass into fragmented patches on both
  backbones, whereas VM and IBM produce coherent maps. Additional examples appear
  in supplementary \cref{app:qual}.
  
  \paragraph{Reading the Evidence Together.} Black-input induction, activation
  similarity, and $\rho$ expose complementary parts of the same failure.
  Unconstrained EQO fails all three and produces fragmented maps,
  whereas VM and IBM retain strong insertion without these failures. AVM shows why the
  checks must be read together: on VGG16, its $\rho=0.424$ is close to the comparison
  methods, yet it induces 95.2\% of black-input targets
  (\cref{tab:defense,tab:rho,fig:qualitative_main}).
  CUB, stability, architecture, and qualitative results appear in supplementary
  \cref{app:cub_intervention,app:stability,app:arch,app:qual}.
  
  \subsection{RQ3: ClipGrad Controls the Tradeoff}
  \label{sec:frontier}
  
  \paragraph{Balancing Output Sensitivity and Clean-Pathway Dependence.}
  \Cref{fig:frontier} plots insertion AUC against clean-pathway dependence,
  measured as $1-\rho$; the desirable region is the upper right. Unconstrained
  EQO lies far to the right but low because its high score survives removal of
  clean-supporting channels. ClipGrad moves EQO upward: VM restores
  clean-pathway dependence to the range of existing methods, while IBM retains
  more of EQO's insertion gain. Because every EQO variant uses the same insertion
  optimizer, this movement isolates the pathway constraint on both backbones.
  
  \subsection{Transfer Across Modern Architectures}
  \label{sec:vit}
  
  \paragraph{Vision Transformers.}
  For DeiT-Small, $\rho_{\mathrm{token}}$ replaces channel ablation with masking
  clean-supporting patch tokens. Unconstrained EQO versus VM gives 99.5\% versus
  0.1\% black-input induction and MSE/cosine $0.260/0.826$ versus
  $0.116/0.918$ (\cref{tab:defense,tab:deit}). VM also reduces the token ratio
  from $0.453$ to $0.397$, although AttnLRP ($0.467$) and SLOC ($0.462$) show that
  it separates methods less strongly. Black-input induction and
  activation deviation give the clearest separation
  (supplementary \cref{app:deit_probe}).
  
  \begin{table}[t]
  \caption{DeiT-Small output scores, last-attention similarity, and token
  intervention. Mean $\rho_{\mathrm{token}}$ uses a common $n=500$ set.}
  \label{tab:deit}
  \centering\small\setlength{\tabcolsep}{1.5pt}
  \begin{tabular}{lccccc}
  \toprule
  Method & Ins. $\uparrow$ & Del. $\downarrow$ & MSE $\downarrow$ & Cos. $\uparrow$ & \shortstack{$\rho_{\mathrm{token}}$\\$\downarrow$}\\
  \midrule
  Trans. Exp.~\cite{chefer2021transformer} & .488 & .171 & .186 & .860 & \textbf{.364}\\
  Ext. Pert.~\cite{fong2019understanding} & .545 & .243 & .175 & .871 & .387\\
  LeGrad~\cite{bousselham2025legrad} & .489 & .170 & .186 & .860 & .370\\
  AttnLRP~\cite{achtibat2024attnlrp} & .518 & \textbf{.061} & .193 & .861 & .467\\
  SLOC~\cite{haddad2025soft} & .596 & .175 & .189 & .858 & .462\\
  \midrule
  FEI$_{\mathrm{VM}}$ & \textbf{.609} & .165 & \textbf{.116} & \textbf{.918} & .397\\
  Unconstrained EQO  & .844 & .184 & .260 & .826 & .453\\
  \bottomrule
  \end{tabular}
  \end{table}
  
On a focused $n=500$ ConvNeXt-Tiny~\cite{liu2022convnet} study, Unconstrained
EQO reaches 0.791 insertion and $\rho=0.865$, while VM gives $\rho=0.492$.
In the separate 999-target black-input test, VM reduces induction from 100\%
to 0.1\% (supplementary \cref{tab:convnext,app:arch}).

\section{Conclusion and Limitations}
\label{sec:conclusion}

EQO shows that leading insertion scores can coexist with rerouting from the
clean prediction pathway, which the clean-support intervention and activation
comparison expose. Across CNNs and Vision Transformers, ClipGrad preserves strong output control
while constraining EQO toward clean internal states. Faithful attribution
requires both output control and computational integrity. Benchmarks can
therefore pair insertion with a clean-pathway check, such as $\rho$ or
black-input induction, before reading a high score as faithful.\footnote{Code:
\url{https://github.com/zhangsiyu1103/fei}}

\paragraph{Limitations.} Internal faithfulness, as measured here, rests on
assumptions that limit its generality. It takes the clean computation as the
reference (\cref{sec:diagnosis}), yet every insertion step changes hidden
states, so our measures rank methods rather than certify them: $\rho$ sets no
threshold at which a deviation changes what an attribution explains. ClipGrad constrains activations relative to their
clean values as a proxy for the pathway, which is why $\rho$ tests dependence
separately. In
turn, $\rho$ depends on its intervention: greedy channel selection at one
block, a stopping threshold, and zero ablation, which can create unnatural
internal states. Unconstrained EQO stays separated across the tested VGG16
layers and thresholds (supplementary \cref{tab:rho_sensitivity}), but mean
replacement can give ratios above one, and activation patching is untested.
On DeiT-Small, the token probe does not separate Unconstrained EQO from AttnLRP
or SLOC, and activation similarity covers selected layers. FGVis, whose gradient rule
differs from ClipGrad's (supplementary \cref{app:unconstrained_limit}), is
compared only on black-input induction. The exploit is shown for
insertion with zero, blur, and solid-color references; in-distribution
perturbations~\cite{nieradzik2025reliable} or ROAD
imputation~\cite{rong2022consistent} may limit rerouting. All experiments use
single-label image classification.
\clearpage
{
    \small
    \bibliographystyle{ieeenat_fullname}
    \bibliography{main}
}
% Continue numbering and resolve all references within this document.
\clearpage
\maketitlesupplementary
\appendix
\section*{Use of LLM}
OpenAI Codex and Anthropic Claude Code were used for experiment scripting,
implementation checks, and manuscript editing. The authors verified the
resulting code, experimental outputs, and text.

\section{EQO and ClipGrad Implementation}
\label{app:eqo_details}

All attribution maps are computed post hoc with frozen model parameters.  EQO
optimizes the five replacement fractions from largest to smallest, equivalently
the retained-area levels from smallest to largest, while maintaining
nested soft masks through projected non-negative increments.  The target is
the model's clean top-1 class.  \Cref{alg:fei} gives the complete procedure;
setting the ClipGrad variant to \textsc{None} produces unconstrained EQO.
Optimization maximizes the pre-softmax target logit $f_{y_t}$; insertion and
deletion are scored with the softmax probability $\phi$.  It uses Adam with the
settings reported in \cref{sec:setup}, and the learning rate follows cosine
annealing over the iterations of each quantile level.

\section{Clean-Support Intervention Details}
\label{app:rho}
\label{app:pathway}
The clean-support intervention measures how much insertion AUC survives after
removing channels that support the clean prediction. For each image, the
target class is the model's own clean
argmax, with baseline probability
$p_0=\mathrm{softmax}(f(x))_{y_t}$. We build
$\mathcal{S}_{\text{clean}}$ by steepest-descent greedy ablation at the
penultimate feature block: VGG16 \texttt{features.29} (512 channels) and
ResNet50 \texttt{layer4} (2048 channels). Starting with
$\mathcal{S}_{\text{clean}}=\emptyset$, we evaluate every remaining
channel $c$ by additionally zeroing $c$ on top of the already selected set,
select
\[
c^\star=\arg\min_{c\notin\mathcal{S}_{\text{clean}}}
\mathrm{softmax}(f(x;\mathcal{S}_{\text{clean}}\cup\{c\}\ \mathrm{ablated}))_{y_t},
\]
commit $c^\star$ to $\mathcal{S}_{\text{clean}}$, and repeat until the
current probability satisfies $p_S\leq\tau p_0$ with default
$\tau=0.5$, or until no channels remain. The selection metric is always
target-class softmax probability, not logits or margin. For ResNet50,
zeroing a channel is implemented by zeroing its global-average-pooling
value; for VGG16, it zeros the channel's post-MaxPool spatial block in the
flattened feature vector. We then reuse the same
$\mathcal{S}_{\text{clean}}$ for every attribution method, zero-ablate this
subset during the insertion sweep, and compute
\begin{equation}
\rho=\frac{\text{AUC}_{\text{ablated}}}{\text{AUC}_{\text{intact}}}.
\end{equation}
A high value means the attribution's insertion score survives removal of the
clean-supporting subset. We use zero ablation because mean ablation can
inject feature-like signal and was observed to produce ill-conditioned
ratios ($\rho>1$). The diagnostic is independent of the activation MSE/cosine
objective used by ClipGrad. Per-method results and confidence intervals appear
in \cref{tab:rho_ci}.

\subsection{Random size-matched control}
\paragraph{Random size-matched channel subsets.}
The main $\rho$ diagnostic ablates a greedily constructed clean-supporting
channel subset $\mathcal{S}_{\text{clean}}$. To test whether the observed
score drops are due to ablating the same number of channels, we run a
size-matched random control. For each image, we sample a random channel
subset with $|\mathcal{S}|=|\mathcal{S}_{\text{clean}}|$, compute
$\rho=\text{AUC}_{\text{ablated}}/\text{AUC}_{\text{intact}}$, and repeat
three times with different random seeds, reporting the mean across seeds
(standard deviations are below $0.0015$ for all methods). Random subsets
leave the external score almost unchanged for all methods
($\rho\approx0.995$ on average), while the greedy subset substantially
reduces AUC for the baseline methods. The effect
therefore depends on selecting channels that support the clean prediction,
not on the number of channels ablated.

\begin{table}[t]
\centering\small
\caption{Random size-matched control for the $\rho$ diagnostic on
VGG16/ImageNet. For each image, random subsets match
$|\mathcal{S}_{\text{clean}}|$ and are averaged over three random seeds.
Higher $\rho$ means the score survives ablation; the random column is a control,
while lower clean-support $\rho$ indicates stronger pathway dependence.}
\label{tab:rho_random_control}
\begin{tabular}{lcc}
\toprule
Method & Random $\rho$ & Clean $\rho$ $\downarrow$ \\
\midrule
Grad-CAM & $0.994$ & $0.388$ \\
Grad-CAM++ & $0.994$ & $0.358$ \\
Score-CAM & $0.995$ & $0.361$ \\
Lift-CAM & $0.995$ & $0.377$ \\
RISE & $0.994$ & $0.357$ \\
Extremal Perturbation & $0.993$ & $0.368$ \\
IIA & $0.995$ & $0.374$ \\
FG-VCE & $0.994$ & $0.418$ \\
\midrule
FEI$_{\text{VM}}$ & $0.994$ & $0.360$ \\
FEI$_{\text{IBM}}$ & $0.995$ & $0.423$ \\
FEI$_{\text{IVM}}$ & $0.994$ & $0.365$ \\
FEI$_{\text{AVM}}$ & $0.996$ & $0.424$ \\
Unconstrained EQO & $1.001$ & $0.793$ \\
\bottomrule
\end{tabular}
\end{table}

Under greedy ablation unconstrained EQO remains separated from the other methods
($\rho=0.793$): the clean-supporting subset still affects its score, but far
less than for the baseline methods.

\begin{table}[t]
\centering\small\setlength{\tabcolsep}{0.8pt}
\caption{Clean-pathway independence $\rho$ on ImageNet ($n=3000$). Mean and
95\% confidence interval; higher is worse. The last three comparison methods
have low intact insertion AUC (\cref{tab:external}) and are omitted from
main-paper \cref{tab:rho}.}
\label{tab:rho_ci}
\begin{tabular}{lcc}
\toprule
Method & VGG16 & ResNet50\\
\midrule
Grad-CAM & $.388\,[.381,.395]$ & $.433\,[.427,.438]$\\
Grad-CAM++ & $.358\,[.352,.363]$ & $.421\,[.416,.426]$\\
Score-CAM & $.361\,[.355,.366]$ & $.418\,[.413,.424]$\\
Lift-CAM & $.377\,[.371,.383]$ & $.423\,[.418,.428]$\\
RISE & $.357\,[.350,.363]$ & $.382\,[.377,.389]$\\
Extremal Perturbation & $.368\,[.362,.374]$ & $.408\,[.402,.414]$\\
IIA & $.374\,[.367,.381]$ & $.423\,[.418,.428]$\\
FG-VCE & $.418\,[.411,.425]$ & $.457\,[.451,.463]$\\
Saliency & $.339\,[.332,.346]$ & $.350\,[.344,.357]$\\
SmoothGrad & $.347\,[.340,.353]$ & $.388\,[.382,.394]$\\
Integrated Gradients & $.357\,[.351,.364]$ & $.335\,[.329,.341]$\\
\midrule
FEI$_{\mathrm{VM}}$ & $.360\,[.355,.366]$ & $.422\,[.417,.427]$\\
FEI$_{\mathrm{IBM}}$ & $.423\,[.416,.429]$ & $.458\,[.452,.464]$\\
FEI$_{\mathrm{AVM}}$ & $.424\,[.416,.431]$ & $.413\,[.408,.418]$\\
FEI$_{\mathrm{IVM}}$ & $.365\,[.360,.371]$ & $.417\,[.413,.423]$\\
\midrule
Unconstrained EQO & $.793\,[.784,.802]$ & $.822\,[.815,.830]$\\
\bottomrule
\end{tabular}
\end{table}

\subsection{Layer and Threshold Sensitivity}

\Cref{tab:rho_sensitivity} provides the complete $3\times3$ grid evaluating
$\rho$ across three feature layers and three stopping thresholds.

\begin{table*}[t]
\caption{Clean-support intervention across layers and stopping thresholds on
VGG16/ImageNet ($n=1000$ per cell). F15, F22, and F29 are VGG16 feature
layers; 25, 50, and 75 are target-probability stopping ratios. Higher $\rho$
indicates weaker dependence on the selected clean-supporting channels.}
\label{tab:rho_sensitivity}
\centering\small\setlength{\tabcolsep}{2.3pt}
\begin{tabular}{lccccccccc}
\toprule
& \multicolumn{3}{c}{F15} & \multicolumn{3}{c}{F22} & \multicolumn{3}{c}{F29}\\
\cmidrule(lr){2-4}\cmidrule(lr){5-7}\cmidrule(lr){8-10}
Method & 25 & 50 & 75 & 25 & 50 & 75 & 25 & 50 & 75\\
\midrule
EP & .277 & .427 & .577 & .276 & .424 & .581 & .229 & .373 & .510\\
FEI$_{\mathrm{AVM}}$ & .346 & .498 & .645 & .354 & .508 & .658 & .267 & .423 & .563\\
FEI$_{\mathrm{IBM}}$ & .322 & .487 & .645 & .333 & .494 & .655 & .262 & .422 & .572\\
FEI$_{\mathrm{IVM}}$ & .253 & .416 & .582 & .261 & .422 & .594 & .216 & .368 & .518\\
Unconstrained EQO & \textbf{.820} & \textbf{.867} & \textbf{.906} & \textbf{.835} & \textbf{.878} & \textbf{.916} & \textbf{.726} & \textbf{.799} & \textbf{.852}\\
FEI$_{\mathrm{VM}}$ & .244 & .407 & .576 & .252 & .414 & .588 & .209 & .361 & .511\\
FG-VCE & .350 & .499 & .634 & .353 & .498 & .640 & .279 & .426 & .557\\
Grad-CAM & .313 & .460 & .602 & .309 & .457 & .607 & .246 & .390 & .522\\
Grad-CAM++ & .250 & .408 & .567 & .257 & .413 & .580 & .214 & .360 & .505\\
IIA & .302 & .451 & .594 & .299 & .443 & .594 & .235 & .377 & .511\\
Lift-CAM & .287 & .443 & .592 & .290 & .444 & .601 & .233 & .380 & .518\\
RISE & .282 & .428 & .574 & .276 & .420 & .570 & .223 & .362 & .494\\
Score-CAM & .266 & .420 & .575 & .266 & .418 & .581 & .218 & .362 & .503\\
\bottomrule
\end{tabular}
\end{table*}

\section{Matched-Reference Robustness}
\label{app:reference}

The reference experiment uses 500 fixed VGG16/ImageNet images.  For each
image and each condition, we generate separate unconstrained and VM maps using either
the zero reference or a Gaussian-blurred version of that image, then evaluate the
insertion curve with that same reference.  This matched design isolates
whether the unconstrained--VM contrast persists under a different perturbation
semantics.  All 2,000 image--reference--method evaluations are valid.  The
  paired within-reference differences compare the two methods under identical
perturbation semantics.  Zero and blur define different insertion paths, so
the experiment estimates the unconstrained--VM gap separately for each.

\begin{table*}[t]
\caption{Matched-reference insertion on VGG16/ImageNet ($n=500$). Mean and
95\% confidence interval; the gap is paired within each reference.}
\label{tab:reference_ci}
\centering\small\setlength{\tabcolsep}{1.5pt}
\begin{tabular}{lccc}
\toprule
Reference & Unconstr. & VM & Paired gap\\
\midrule
Zero & $.682\,[.664,.699]$ & $.487\,[.463,.512]$ & $.194\,[.171,.217]$\\
Blur & $.888\,[.875,.901]$ & $.628\,[.604,.652]$ & $.260\,[.236,.283]$\\
\bottomrule
\end{tabular}
\end{table*}

\section{Cross-Dataset Intervention Results}
\label{app:cub_intervention}

\Cref{tab:cub_all} reports the 16-method CUB-200-2011/VGG16 intervention
evaluation on 500 fixed images.  Unconstrained EQO is again separated by the joint pattern of
high insertion and high $\rho$.  The remaining methods occupy different
output--pathway operating points.

\begin{table}[t]
\caption{16-method CUB-200-2011/VGG16 intervention evaluation ($n=500$). Mean and
95\% confidence interval.}
\label{tab:cub_all}
\centering\small\setlength{\tabcolsep}{2pt}
\begin{tabular}{lcc}
\toprule
Method & Insertion & $\rho$\\
\midrule
EP & $.734\,[.712,.756]$ & $.459\,[.442,.476]$\\
FEI$_{\mathrm{AVM}}$ & $.738\,[.717,.760]$ & $.471\,[.452,.489]$\\
FEI$_{\mathrm{IBM}}$ & $.764\,[.744,.785]$ & $.459\,[.444,.474]$\\
FEI$_{\mathrm{IVM}}$ & $.724\,[.702,.746]$ & $.419\,[.406,.433]$\\
FEI$_{\mathrm{VM}}$ & $.735\,[.713,.757]$ & $.410\,[.397,.424]$\\
Unconstrained EQO & $.783\,[.764,.803]$ & $.736\,[.710,.761]$\\
FG-VCE & $.750\,[.729,.771]$ & $.540\,[.522,.557]$\\
Grad-CAM & $.743\,[.722,.764]$ & $.482\,[.465,.498]$\\
Grad-CAM++ & $.726\,[.704,.748]$ & $.418\,[.405,.432]$\\
IIA & $.724\,[.702,.746]$ & $.462\,[.444,.479]$\\
Integrated Gradients & $.094\,[.079,.109]$ & $.383\,[.367,.398]$\\
Lift-CAM & $.739\,[.718,.760]$ & $.456\,[.441,.472]$\\
RISE & $.709\,[.686,.733]$ & $.442\,[.426,.458]$\\
Saliency & $.510\,[.486,.534]$ & $.400\,[.381,.419]$\\
Score-CAM & $.729\,[.706,.751]$ & $.417\,[.403,.431]$\\
SmoothGrad & $.648\,[.626,.670]$ & $.422\,[.404,.439]$\\
\bottomrule
\end{tabular}
\end{table}

\section{ResNet50 Output--Pathway Tradeoff}
\label{app:resnet_frontier}

The ResNet50 frontier is shown beside VGG16 in main-paper
\cref{fig:frontier}. Its numerical coordinates come from the common insertion
evaluation in \cref{tab:external} and the clean-support intervention in
\cref{tab:rho_ci}.

\section{Reference-Resampling Stability}
\label{app:stability}

We regenerate VM and IBM maps for 100 fixed VGG16/ImageNet images under three
independent random solid-color reference sequences.  Stability is the mean
pairwise intersection-over-union of the top-20\% attribution support across
the three runs.  Spatial coherence is summarized by the fraction of selected
pixels in the largest connected component; a map is called fragmented when
this fraction is below 0.25.  VM and IBM have three independently generated
maps per image and therefore support the across-reference IoU measurement.
The single unconstrained map per image supplies the fragmentation comparison.

\begin{table}[t]
\caption{Reference-resampling stability with 95\% confidence intervals
($n=100$).}
\label{tab:stability_ci}
\centering\small\setlength{\tabcolsep}{5pt}
\begin{tabular}{lcc}
\toprule
Method & Support IoU & Largest component\\
\midrule
FEI$_{\mathrm{VM}}$ & $.913\,[.904,.923]$ & $.757\,[.734,.779]$\\
FEI$_{\mathrm{IBM}}$ & $.842\,[.833,.851]$ & $.612\,[.587,.637]$\\
Unconstrained EQO & -- & $.024\,[.021,.028]$\\
\bottomrule
\end{tabular}

\vspace{3pt}
\begin{tabular}{lc}
\toprule
Method & Fragmented maps\\
\midrule
FEI$_{\mathrm{VM}}$ & $0/300$\\
FEI$_{\mathrm{IBM}}$ & $23/300$\\
Unconstrained EQO & $100/100$\\
\bottomrule
\end{tabular}
\end{table}

\section{DeiT Token-Intervention Protocol}
\label{app:deit_probe}

For DeiT-Small, the clean-support intervention is defined over patch tokens
rather than CNN channels.  We use the same clean top-1 target, greedily select
tokens whose masking most reduces that target, and compute the ratio of
ablated to intact insertion AUC on 500 fixed images.  Candidate token effects
share a common prefix computation within each greedy step; this changes only
runtime, not the selected set.  All 3,500 method--image evaluations for the seven
reported methods are valid.  The paired unconstrained-minus-VM difference is
$0.056\,[0.030,0.083]$, establishing that VM changes the token dependence of
the same optimizer. Across methods, however, $\rho_{\mathrm{token}}$ is less
discriminative than the CNN channel probe: AttnLRP and SLOC have ratios of
$0.467$ and $0.462$, and Transformer Explainability has a lower ratio than
VM (paired difference $-0.033\,[-0.048,-0.018]$). Accordingly, the black-input and
attention-trajectory evidence carry the transformer diagnosis, while the token
probe corroborates the paired mitigation effect.

We also measure last-attention activation matching: MSE and cosine similarity
between the clean and perturbed outputs of the final attention module (all
tokens, after the output projection), averaged along the zero-reference
insertion trajectory as in \cref{tab:activation}.  VM is applied to the
outputs of the attention modules in all 12 blocks, so this measurement reads
one of the activations VM constrains; the independently constructed token
intervention provides the complementary clean-anchored comparison.

\begin{table}[t]
\caption{DeiT-Small token-intervention ratio on 500 images. Mean and 95\%
confidence interval; lower is better.}
\label{tab:deit_rho_ci}
\centering\small\setlength{\tabcolsep}{6pt}
\begin{tabular}{lc}
\toprule
Method & $\rho_{\mathrm{token}}$\\
\midrule
Transformer Explainability & $.364\,[.347,.381]$\\
Extremal Perturbation & $.387\,[.370,.404]$\\
LeGrad & $.370\,[.352,.388]$\\
AttnLRP & $.467\,[.440,.495]$\\
SLOC & $.462\,[.441,.482]$\\
\midrule
FEI$_{\mathrm{VM}}$ & $.397\,[.381,.413]$\\
Unconstrained EQO & $.453\,[.423,.483]$\\
\bottomrule
\end{tabular}
\end{table}

\begin{table}[t]
\caption{DeiT-Small last-attention activation-matching results.
MSE $\downarrow$ and cosine similarity $\uparrow$.}
\label{tab:deit_activation}
\centering\small\setlength{\tabcolsep}{6pt}
\begin{tabular}{lcc}
\toprule
Method & MSE $\downarrow$ & Cos. $\uparrow$\\
\midrule
Transformer Explainability & .186 & .860\\
Extremal Perturbation & .175 & .871\\
LeGrad & .186 & .860\\
AttnLRP & .193 & .861\\
SLOC & .189 & .858\\
\midrule
FEI$_{\mathrm{VM}}$ & \textbf{.116} & \textbf{.918}\\
Unconstrained EQO & .260 & .826\\
\bottomrule
\end{tabular}
\end{table}

\section{Runtime Protocol}
\label{app:runtime}

The runtime comparison in \cref{tab:runtime} measures attribution computation
only, excluding model construction, checkpoint loading, data loading, and the
clean target-selection forward pass.  CUDA is synchronized at the timing
boundaries, one warm-up image precedes measurement, and throughput is averaged
over 100 VGG16/ImageNet images on an RTX~3090.  Each perturbation method uses
the largest batch supported by its implementation and available memory.  FEI
includes all five quantiles and 100 optimization steps per quantile; EP and
RISE include their complete mask generation/evaluation procedures.  The
reported values are therefore throughput rather than batch-one latency.

\begin{table}[t]
\caption{Mean seconds per attribution for complete perturbation-based
pipelines on an RTX 3090 over 100 images, using maximum-feasible-batch
throughput.}
\label{tab:runtime}
\centering\small\setlength{\tabcolsep}{9pt}
\begin{tabular}{lccc}
\toprule
Method & FEI$_{\mathrm{VM}}$ & EP & RISE\\
\midrule
Seconds $\downarrow$ & \textbf{1.58} & 33.50 & 6.34\\
\bottomrule
\end{tabular}
\end{table}

\section{Focused ConvNeXt Transfer}
\label{app:convnext}

We evaluate ConvNeXt-Tiny~\cite{liu2022convnet} on 500 ImageNet validation
images as a focused transfer study. ConvNeXt retains a convolutional residual
topology while replacing the ReLU--BatchNorm design of the primary CNN
benchmarks with GELU and LayerNorm. ClipGrad is applied to the outputs of all
18 GELU modules. VM, AVM, and IVM use their original value-based rules; IBM
uses the sign state $h>0$ as positive and $h\leq0$ as non-positive. The
clean-image metrics use the same 500 images, clean top-1 targets, EQO settings
(except $\eta=0.05$), and common evaluator. The all-black rates use the complete category-level
protocol in \cref{app:black}.

\begin{table}[t]
\caption{Focused ConvNeXt-Tiny/ImageNet transfer study ($n=500$). Metrics:
Ins.$\uparrow$, Del.$\downarrow$, MSE$\downarrow$, Cos.$\uparrow$, and
$\rho\downarrow$. Black-input rates appear in \cref{tab:defense}.}
\label{tab:convnext}
\centering\small\setlength{\tabcolsep}{4pt}
\begin{tabular}{lrrrrr}
\toprule
Method & Ins. & Del. & MSE & Cos. & $\rho$\\
\midrule
Grad-CAM & .588 & .229 & .050 & .728 & .553\\
RISE & .541 & .240 & .052 & .713 & .514\\
\midrule
FEI$_{\mathrm{VM}}$  & .513 & .171 & .049 & .737 & \textbf{.492}\\
FEI$_{\mathrm{IBM}}$ & \textbf{.606} & \textbf{.119} & .058 & .685 & .573\\
FEI$_{\mathrm{AVM}}$ & .588 & .130 & \textbf{.043} & \textbf{.766} & .524\\
FEI$_{\mathrm{IVM}}$ & .518 & .133 & .051 & .723 & .501\\
\midrule
Unconstrained EQO & .791 & .279 & .149 & .294 & .865\\
\bottomrule
\end{tabular}
\end{table}

\Cref{tab:convnext} reproduces the central separation. Unconstrained EQO has
the highest insertion AUC, 100\% black-input target induction, the greatest
activation deviation, and the weakest clean-channel dependence. Each ClipGrad
rule sharply reduces the black-input and clean-support failures. IBM provides
the strongest constrained insertion/deletion point, VM provides the lowest
$\rho$, and AVM provides the closest activation trajectory. AVM's 20.7\%
black-input rate also shows that activation similarity and content-free-input
testing identify different aspects of pathway behavior.

The full per-method results appear in \cref{tab:convnext}.

\section{Common Evaluator Protocol}
\label{app:baseline_protocol}

All values in our comparison tables are recomputed under one evaluator.
Methods use the same fixed image
indices, input resolution, clean top-1 target, perturbation fractions, and
reference within each reported condition.  This is necessary because
insertion/deletion values change with the selected images, target definition,
reference, preprocessing, and curve discretization.  The resulting values form
controlled within-table comparisons under a shared evaluation protocol.

For FG-VCE~\cite{zhang2025towards}, the common table evaluates the method's
saliency-partition output as a single map for the clean top-1 target.  This
keeps the evaluator, target, and perturbation path identical across the map
quality comparison.

\section{All-Black-Input Target-Induction Rule}
\label{app:black}
Following the black-input test of FGVis~\cite{wagner2019interpretable}, we run
each attribution method on an
all-black RGB canvas, normalized by the model's standard preprocessing. For
each model and method, we optimize the same canvas separately toward every
output category except the category predicted from the unperturbed canvas.
This leaves 999 non-trivial target trials for each ImageNet model. During
optimization, the replacement reference is a zero tensor in normalized
model-input space; this reference is distinct from the black RGB input. We
then apply the method's final full perturbation. A trial counts as target
induction exactly when the model's top-1 prediction equals the requested
category. The rate therefore measures category coverage after the full
attribution pipeline, not a threshold on the rendered heatmap.

\section{Relation Between Constrained and Unconstrained EQO}
\label{app:unconstrained_limit}
Unconstrained EQO is the insertion optimizer used by the constrained FEI
variants with ClipGrad disabled.  It characterizes the behavior promoted by
the output objective before a clean-pathway constraint is introduced.  The
paired unconstrained and constrained configurations therefore isolate the
effect of ClipGrad while holding the optimizer fixed.

Extremal Perturbation and FGVis also optimize a target score together with
spatial, sparsity, or foreground constraints
\cite{fong2019understanding,wagner2019interpretable}. Removing those auxiliary
terms places them in the same target-score-only regime as Unconstrained EQO.
The parameterizations remain different: EQO optimizes a nested insertion
sequence rather than a single mask. This shared regime connects the diagnostic
to the broader family of optimization-based attribution methods.

The activation-bound rule of FGVis, applied in our EQO optimizer for
\cref{tab:defense}, zeros every gradient of a unit whose perturbed activation
lies outside the interval between zero and its clean value, including updates
that would move it back. VM instead blocks only updates that move an
activation farther from its clean value (\cref{eq:vm}). The two rules are
compared on black-input induction, where both suppress target induction.

\section{Sanity Check}
\label{sec:sanity}

\begin{figure}[t]
\centering
\setlength\tabcolsep{1pt}
\footnotesize
\begin{tabular}{cccc}
Input& Original & Linear.6& Linear.3\\
\includegraphics[width=.23\columnwidth]{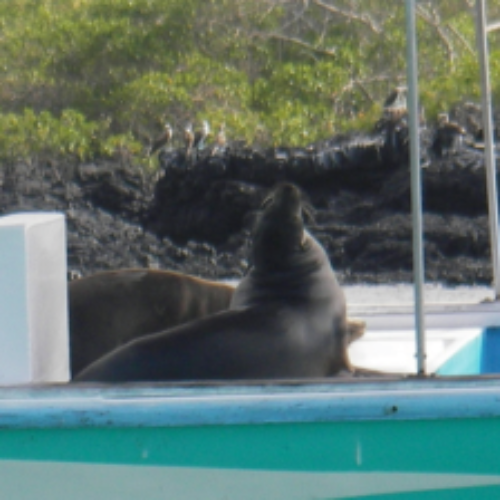} & 
\includegraphics[width=.23\columnwidth]{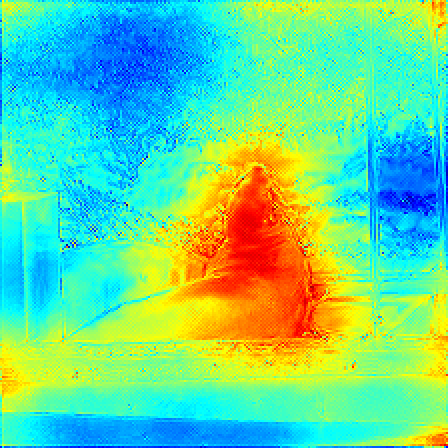} & 
\includegraphics[width=.23\columnwidth]{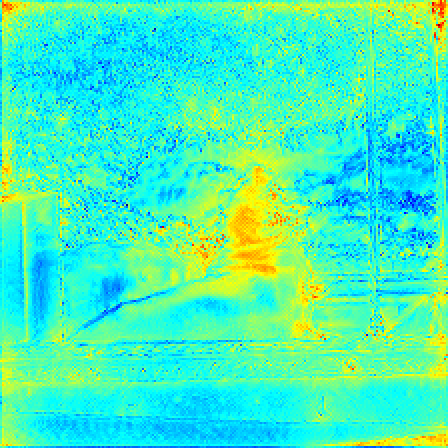}  &
\includegraphics[width=.23\columnwidth]{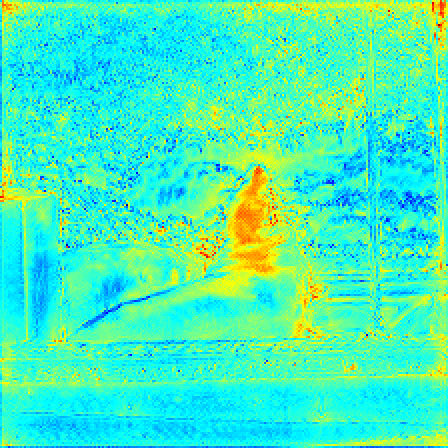}  \\
\end{tabular}

\begin{tabular}{cccc}
Linear.0 &  Feature.28&Feature.26 &Feature.24\\
\includegraphics[width=.23\columnwidth]{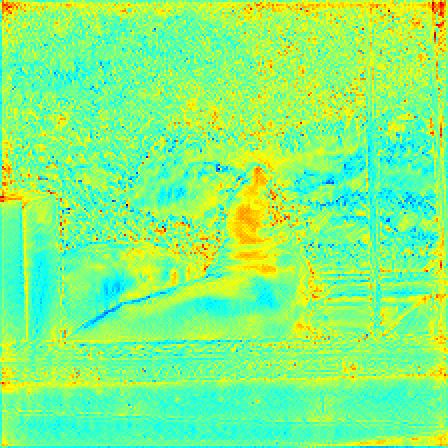}  &
\includegraphics[width=.23\columnwidth]{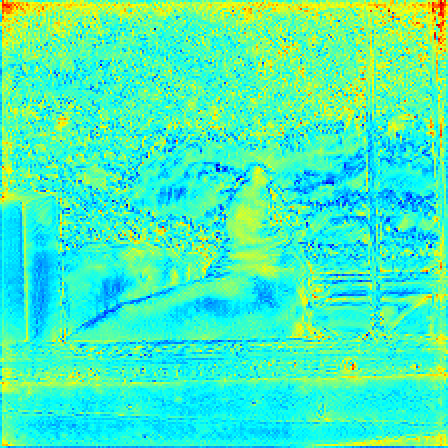}  &
\includegraphics[width=.23\columnwidth]{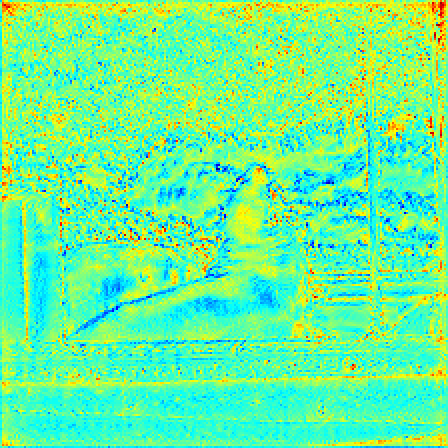}  &
\includegraphics[width=.23\columnwidth]{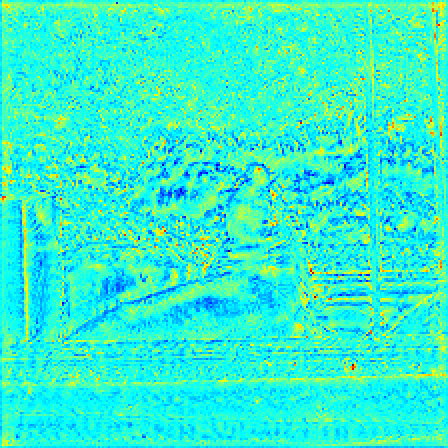}  \\
\end{tabular}
\caption{\textbf{Cascading randomization sanity check.} Example: \textit{Sea Lion}, VGG16, FEI$_\text{IBM}$.} 
\label{fig:fig-sanity}
\end{figure}

Following Adebayo \etal~\cite{adebayo2018sanity}, we examine parameter
sensitivity through cascading randomization (\cref{fig:fig-sanity}). Layers
are progressively randomized from top to bottom, and the FEI$_{\text{IBM}}$
map gradually loses semantic structure as learned features are removed.

High-level object structure disappears before low-level edge patterns because
the corresponding low-level features remain intact until later in the
randomization sequence.  This progression supports dependence on learned
parameters at multiple depths.

The gradual degradation from semantic structure to noise supports that 
FEI depends on learned parameters rather than exploiting 
input artifacts or edge priors.

\begin{table}[t]
\centering
\caption{Ablation study on ImageNet/VGG16. Rows use FEI$_\text{VM}$ unless noted.}
\label{tab:ablation}
\small
\setlength{\tabcolsep}{3.5pt}
\begin{tabular}{lcccc}
\toprule
Configuration & Ins.$\uparrow$ & Del.$\downarrow$ & MSE$\downarrow$ & Cosine.$\uparrow$ \\
\midrule
\textbf{FEI$_\text{VM}$ (full method)} & 0.512 &0.117 & 0.545 & 0.688 \\
\midrule
\multicolumn{5}{l}{\textit{Quantile levels }} \\
\quad $(0.1,0.5,0.9)$ & 0.509 & 0.107 & 0.568 & 0.683 \\
\quad $\{0.1,0.2,\ldots,0.9\}$ & 0.532 & 0.125 & 0.547 & 0.690 \\
\midrule
\multicolumn{5}{l}{\textit{Regularization schedule ($\beta$)}} \\
\quad Constant ($\beta=1.0$) & 0.513 & 0.113 & 0.550 & 0.688 \\
\quad Constant ($\beta=10^{-2}$) & 0.500 & 0.131 & 0.543 & 0.680 \\
\quad Constant ($\beta=10^{-3}$) & 0.339 & 0.148 & 0.785 & 0.526 \\
\midrule
\multicolumn{5}{l}{\textit{Gradient clipping scope}} \\
\quad Early layers only (1-4) & 0.550 & 0.114 & 0.610 & 0.664 \\
\quad Early layers only (1-8) & 0.544 & 0.113 & 0.584 & 0.673 \\
\quad Early layers only (1-16) & 0.525 & 0.113 & 0.559 & 0.681 \\
\quad Later layers only (16-30) & 0.430 & 0.069 & 0.717 & 0.603 \\ 
\midrule
\multicolumn{5}{l}{\textit{Area Regularization}} \\
\quad L1($\beta=10^{-1}$) & 0.328 & 0.226 & 0.794 & 0.486 \\
\quad L1($\beta=10^{-2}$) & 0.464 & 0.147 & 0.576 & 0.648 \\
\quad L1($\beta=10^{-3}$) & 0.213 & 0.102 & 1.016 & 0.392 \\
\quad No Constraint & 0.127 & 0.156 & 1.118 & 0.290 \\
\midrule
\multicolumn{5}{l}{\textit{Alternative optimization targets (VM)}} \\
\quad EQO + hard mask (STE) & 0.128 & 0.228 & 1.110 & 0.284 \\
\quad EQO deletion optimized & 0.145 & 0.489 & 1.043 & 0.278 \\
\midrule
\multicolumn{5}{l}{\textit{Optimization Step}} \\
% \quad epochs(500) & 0.512 & 0.122 & 0.539 & 0.683 \\
\quad epochs(300) & 0.518 & 0.129 & 0.521 & 0.693 \\
\quad epochs(200) & 0.516 & 0.125 & 0.527 & 0.692 \\
\quad epochs(50) & 0.491 & 0.088 & 0.624 & 0.663 \\
\bottomrule
\end{tabular}
\end{table}

\begin{figure*}[t]
\centering
\setlength{\tabcolsep}{1pt}
\caption{Layer clipping-scope ablation.}
\label{fig:clipping_ablation}

\begin{tabular}{@{}cccccc@{}}
\toprule
Original & FEI$_{\text{VM}}$ & FEI$_{\text{VM}}$(1--4) & FEI$_{\text{VM}}$(1--8) & FEI$_{\text{VM}}$(1--16) & FEI$_{\text{VM}}$(17--30) \\
\midrule

% Row 1
\includegraphics[width=0.155\textwidth]{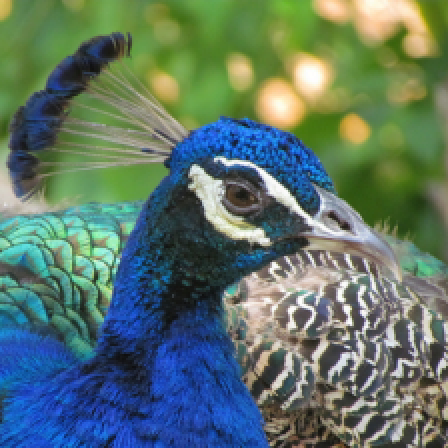} &
\includegraphics[width=0.155\textwidth]{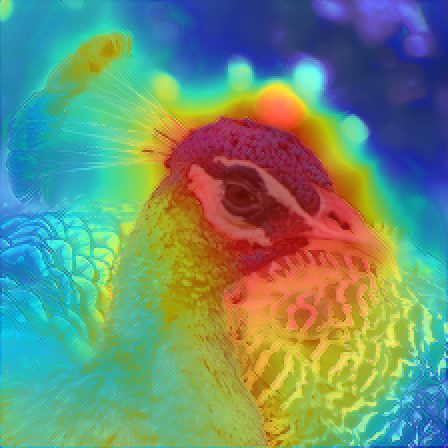} &
\includegraphics[width=0.155\textwidth]{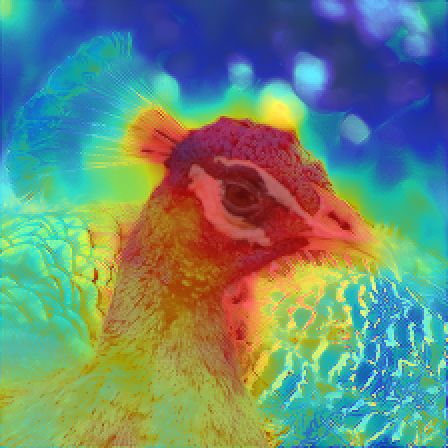} &
\includegraphics[width=0.155\textwidth]{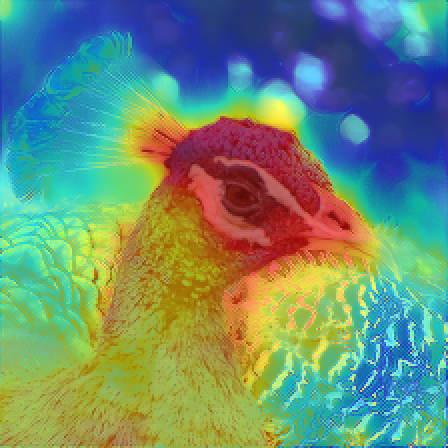} &
\includegraphics[width=0.155\textwidth]{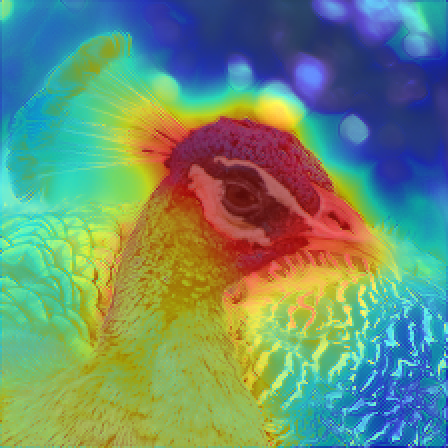} &
\includegraphics[width=0.155\textwidth]{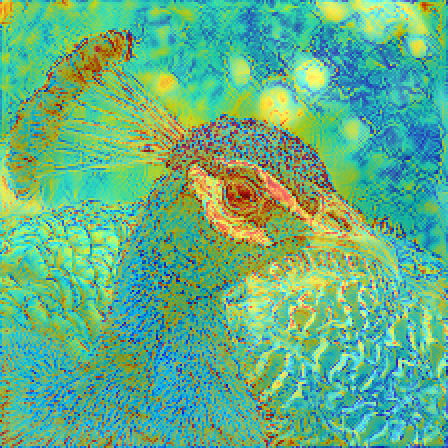} \\[2mm]

% Row 2
\includegraphics[width=0.155\textwidth]{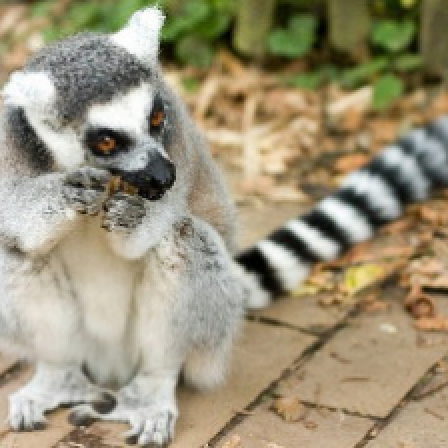} &
\includegraphics[width=0.155\textwidth]{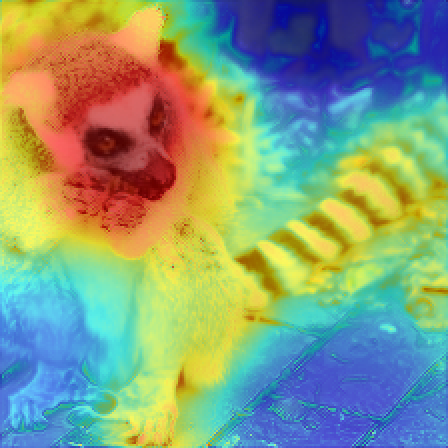} &
\includegraphics[width=0.155\textwidth]{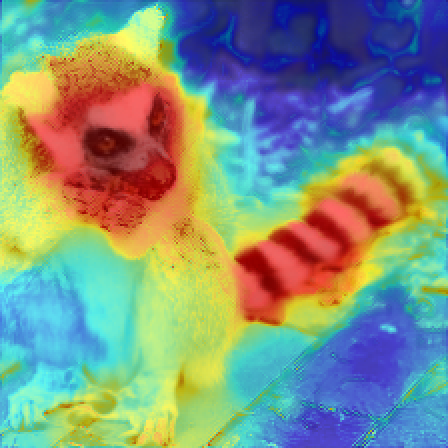} &
\includegraphics[width=0.155\textwidth]{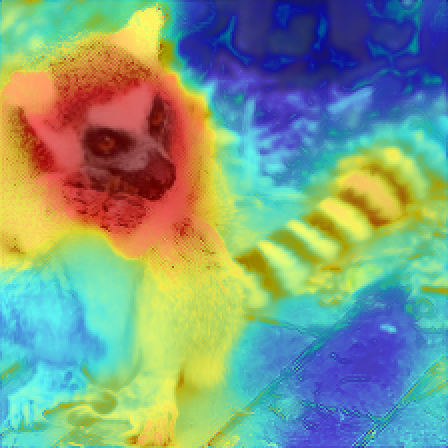} &
\includegraphics[width=0.155\textwidth]{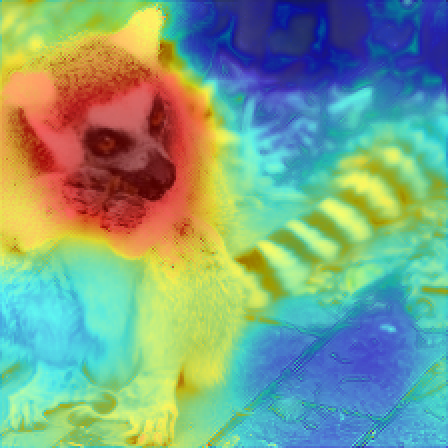} &
\includegraphics[width=0.155\textwidth]{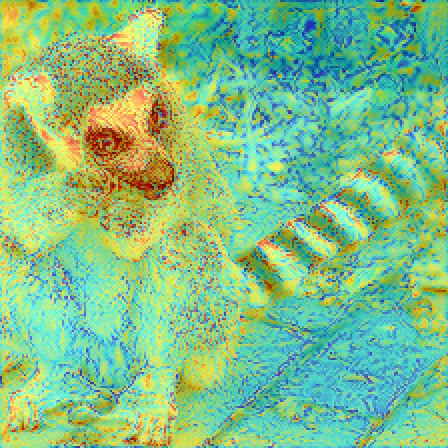} \\[2mm]

% Row 3
\includegraphics[width=0.155\textwidth]{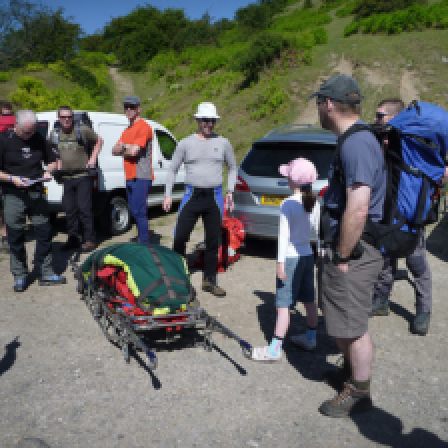} &
\includegraphics[width=0.155\textwidth]{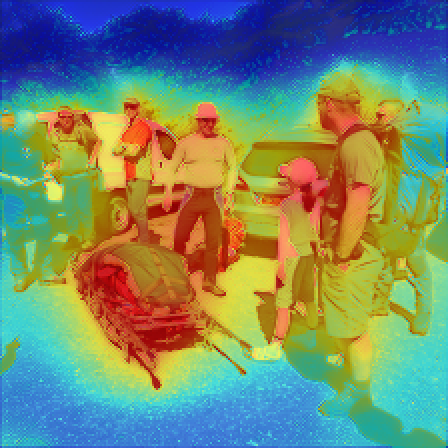} &
\includegraphics[width=0.155\textwidth]{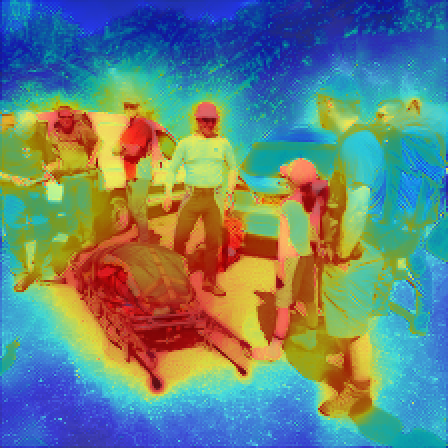} &
\includegraphics[width=0.155\textwidth]{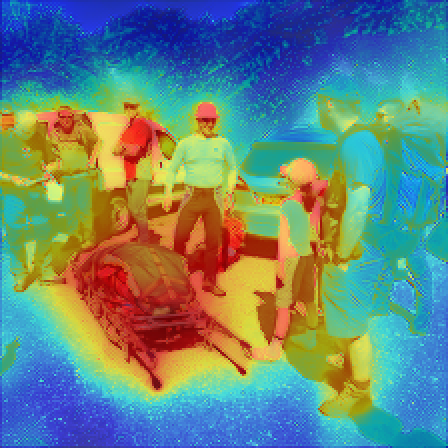} &
\includegraphics[width=0.155\textwidth]{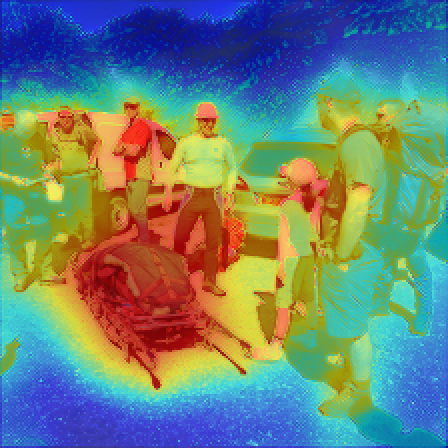} &
\includegraphics[width=0.155\textwidth]{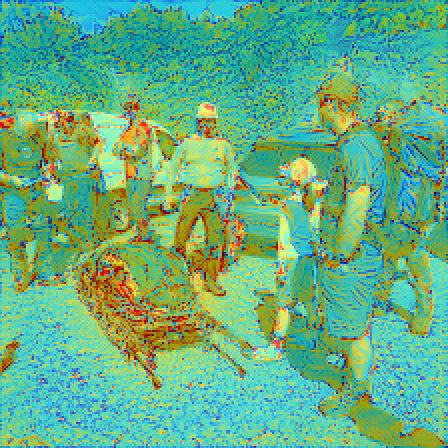} \\
% Row 4
\includegraphics[width=0.155\textwidth]{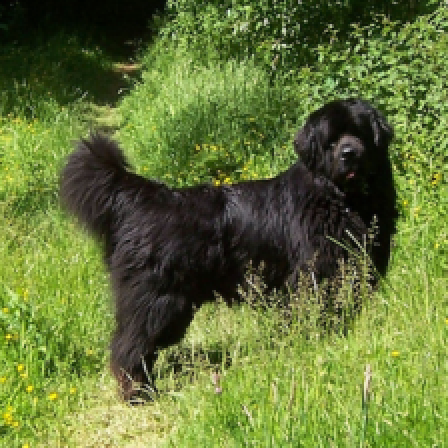} &
\includegraphics[width=0.155\textwidth]{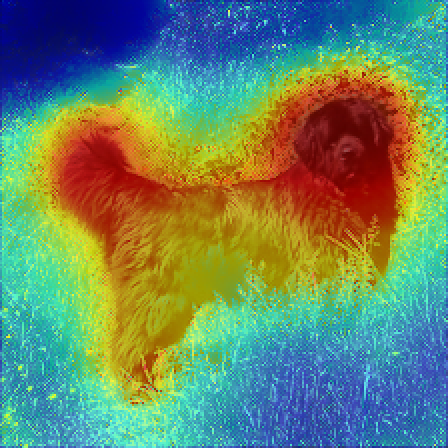} &
\includegraphics[width=0.155\textwidth]{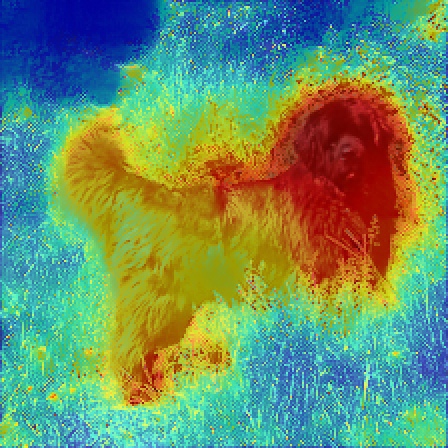} &
\includegraphics[width=0.155\textwidth]{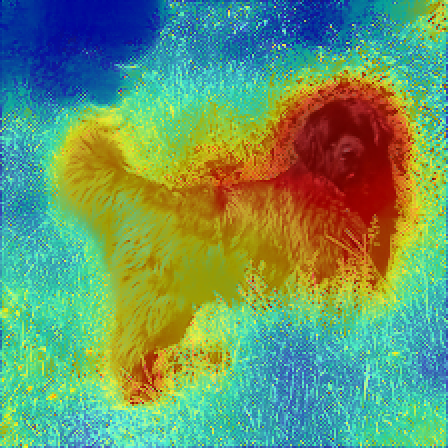} &
\includegraphics[width=0.155\textwidth]{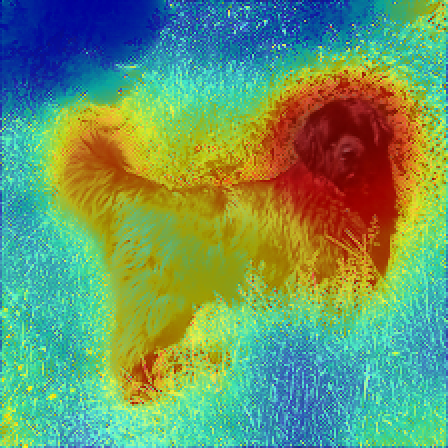} &
\includegraphics[width=0.155\textwidth]{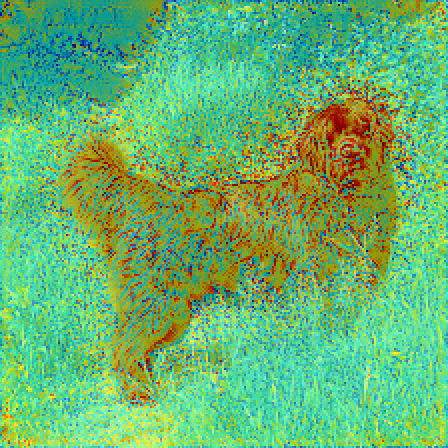} \\
\bottomrule
\end{tabular}

\end{figure*}

\section{Ablations and Implementation Checks}
\label{app:ablations}
\label{sec:ablation}

Unless otherwise stated, ablations are run on ImageNet with VGG16 using
FEI$_\text{VM}$, 100 iterations per quantile,
$\mathcal{Q}=\{0.1,0.3,0.5,0.7,0.9\}$, $\beta=0.1$, and ClipGrad across all
convolutional layers.  \Cref{tab:ablation} reports insertion, deletion, MSE, and cosine
similarity for optimization and implementation choices.  The clean-support
intervention is evaluated separately in \cref{app:pathway}.

\subsection{Ensemble Quantile Optimization vs. Area Regularization}

This experiment asks whether the nested quantile construction contributes
beyond a direct area penalty.  We replace it with
$\mathcal{L}=-f_{y_t}(\tilde{x})+\beta\|\alpha\|_1$ and also test optimization
without an area constraint.  Across the tested weights, direct area penalties
give lower insertion and larger activation deviation than the default EQO
configuration.  Removing the area constraint gives the weakest compact-map
behavior.  The nested quantile construction therefore supplies both curve
coverage and an explicit retained-area target.

We also vary the mask relaxation and output objective while retaining VM.
The straight-through hard mask reduces insertion and activation matching, so
the continuous masks provide a more effective optimization surface.  Direct
deletion optimization yields high deletion AUC and low insertion AUC, whereas
optimizing insertion and evaluating deletion produces the intended
reveal-and-remove comparison.  Numerical results for both variants are in
\cref{tab:ablation}.

\subsection{Quantile Level Selection}

This experiment tests sensitivity to curve discretization.  Three, five, and
nine levels produce insertion scores within 0.023, MSE within 0.023, and cosine
similarity within 0.007.  Five levels provide the middle computational budget
while covering the insertion curve from 10\% to 90\% retained area.

\subsection{Regularization Strength}

The weight $\beta$ controls the retained-area penalty in \cref{eq:ins}.
Results are stable from $\beta=0.1$ to $1.0$; insertion and activation matching
degrade as the weight falls to $10^{-2}$ and $10^{-3}$.  We use $\beta=0.1$ as
the smallest tested value in the stable range.

\subsection{Optimization Iterations}

This experiment tests convergence with 50, 100, 200, and 300 iterations per
quantile.  Fifty iterations reduces insertion and activation matching, while
200 and 300 provide only small changes beyond 100.  We use 100 iterations at
the beginning of this saturation region.

\subsection{Gradient Clipping Scope}

This experiment localizes where activation constraints matter.  Extending
ClipGrad from layers 1--4 through layers 1--16 progressively improves MSE and
cosine similarity while reducing insertion slightly.  Applying ClipGrad only
to layers 16--30 gives weaker insertion and activation matching than applying
it to the early stack.  The visualizations in \cref{fig:clipping_ablation}
show the same pattern: configurations covering the early or complete
convolutional stack remain coherent, whereas late-only maps are diffuse. These results are consistent
with early deviations propagating forward, and motivate clipping the full
convolutional stack as the default operating point.

\begin{figure*}[ht]
\centering
  \begin{subfigure}[b]{0.98\textwidth}
    \centering
    \includegraphics[width=\linewidth]{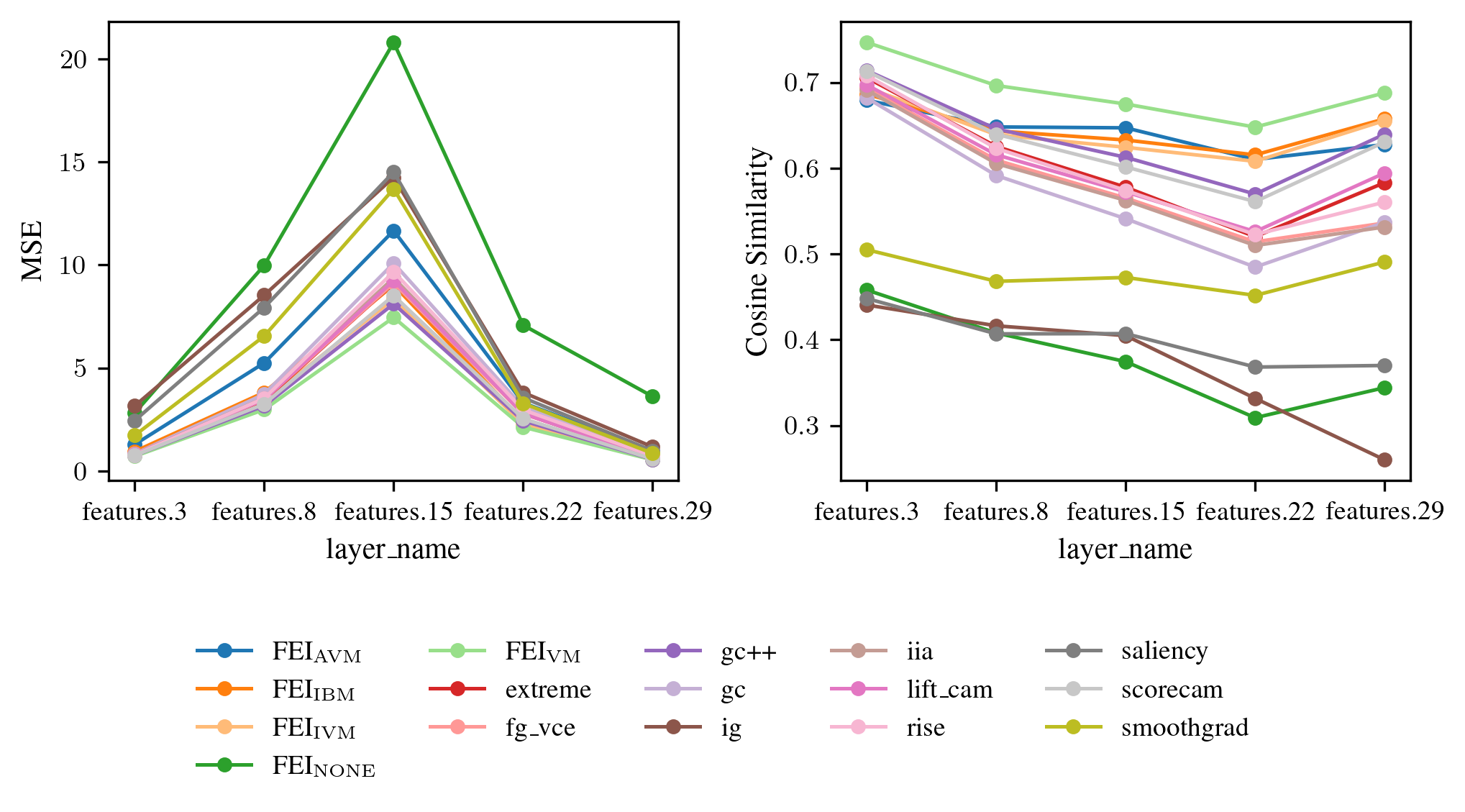}
    \caption{VGG16-ImageNet}
  \end{subfigure}
  
  \begin{subfigure}[b]{0.98\textwidth}
    \centering
    \includegraphics[width=\linewidth]{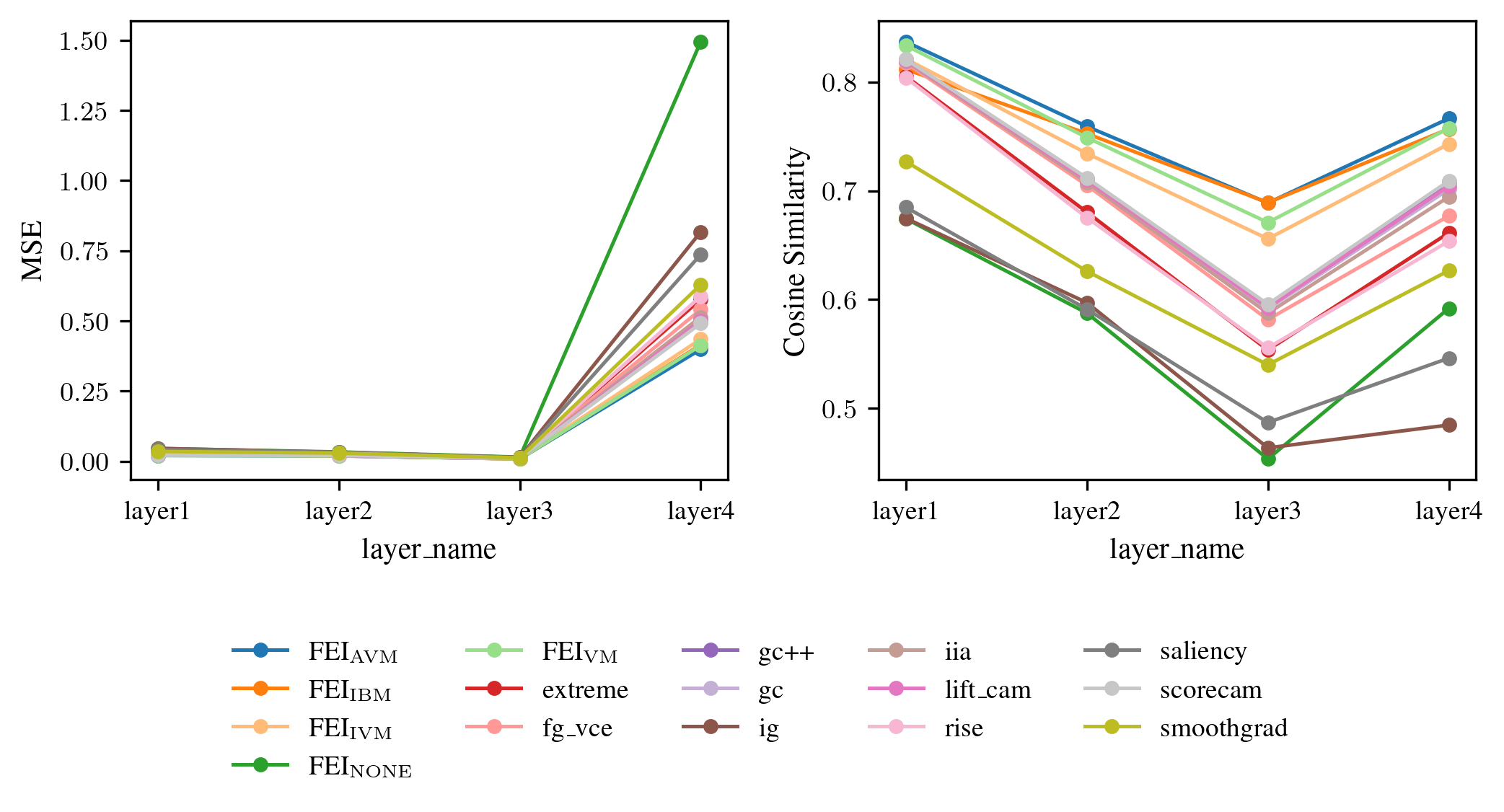}
    \caption{ResNet50-ImageNet}
  \end{subfigure}
\caption{Layer-wise internal preservation across architectures and datasets.
MSE$\downarrow$, cosine similarity$\uparrow$. Continued on next page.}
\label{fig:supp_layerwise_internal}
\end{figure*}

\begin{figure*}[ht]
\ContinuedFloat
\centering
  \begin{subfigure}[b]{0.98\textwidth}
    \centering
    \includegraphics[width=\linewidth]{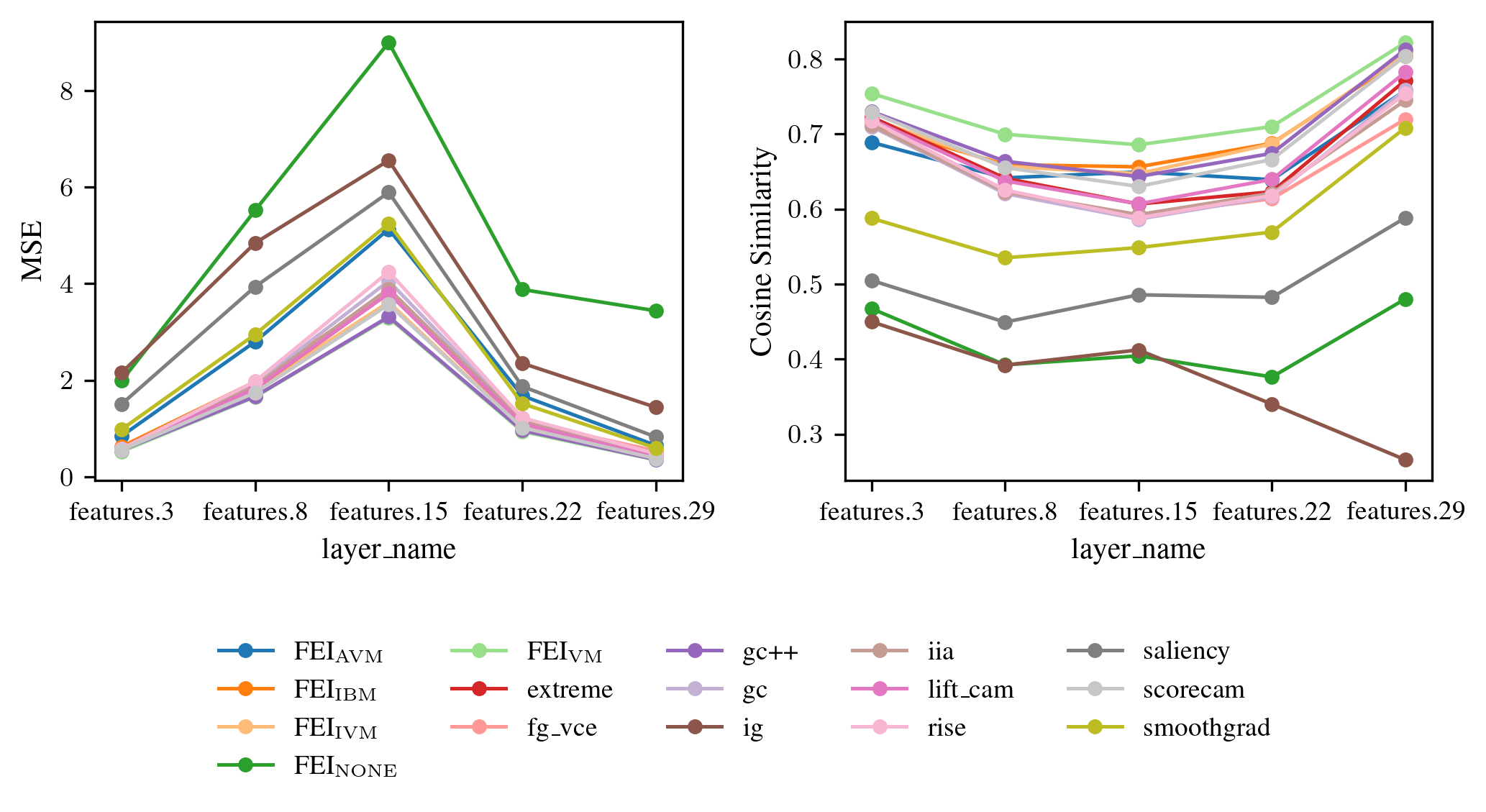}
    \caption{VGG16-CUB-200-2011}
  \end{subfigure}
  
  \begin{subfigure}[b]{0.98\textwidth}
    \centering
    \includegraphics[width=\linewidth]{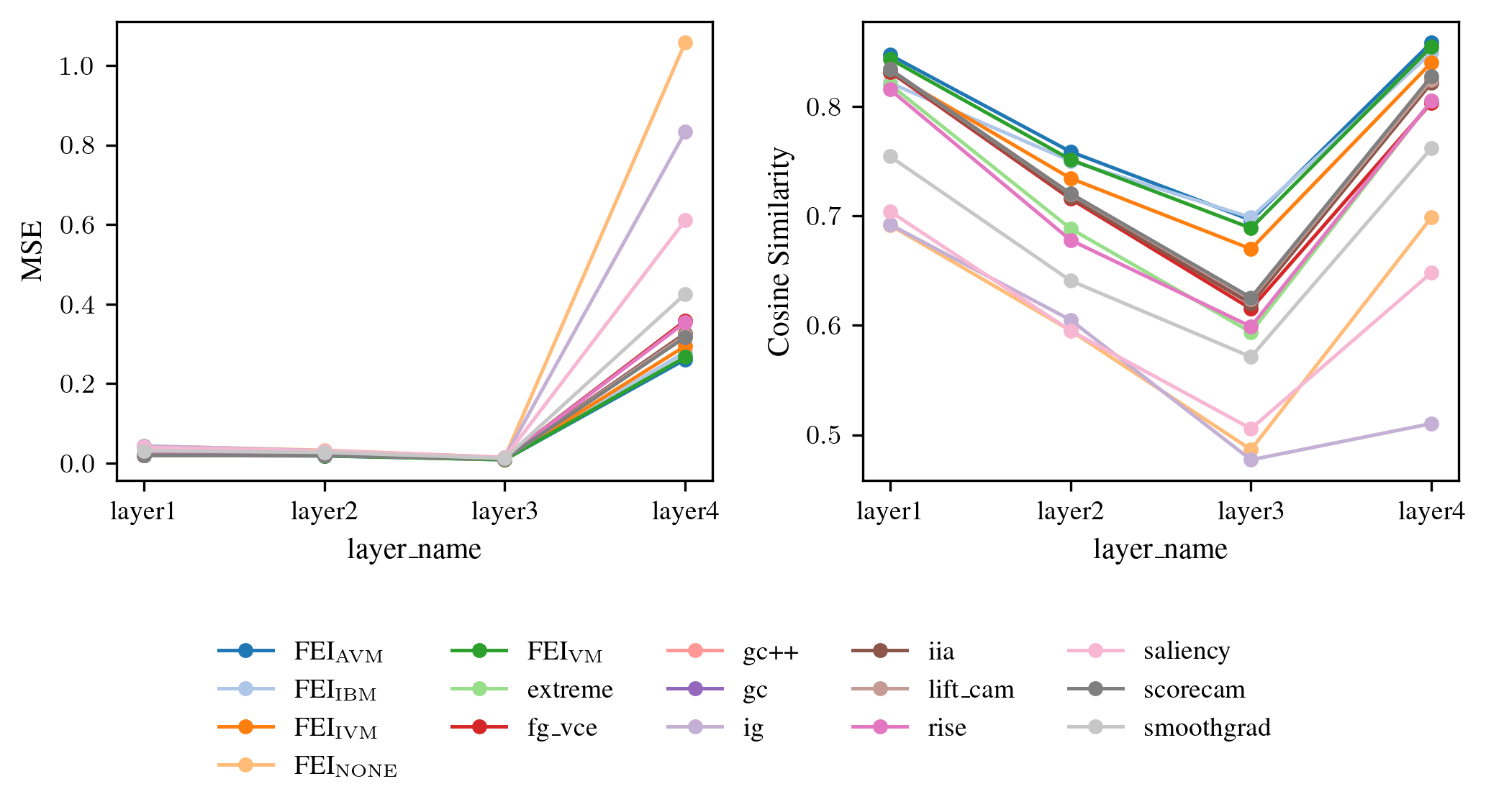}
    \caption{ResNet50-CUB-200-2011}
  \end{subfigure}
\caption{(Continued) Layer-wise internal preservation.}
\end{figure*}

\begin{table*}[t]
\centering
\caption{Layer-wise internal activation matching on ResNet50. MSE$\downarrow$,
Cos.$\uparrow$.}
\label{tab:supp_layerwise_resnet}
\small
\setlength{\tabcolsep}{4pt}
\renewcommand{\arraystretch}{1.02}
\textbf{ImageNet}\\[2pt]
\begin{tabular}{lcccccccc}
\toprule
\multirow{2}{*}{Method} &
\multicolumn{4}{c}{MSE $\downarrow$} &
\multicolumn{4}{c}{Cos. $\uparrow$} \\
\cmidrule(lr){2-5} \cmidrule(lr){6-9}
& L1 & L2 & L3 & L4 & L1 & L2 & L3 & L4 \\
\midrule
Grad-CAM++ & .021 & .021 & .011 & .495 & .820 & .711 & .595 & .708 \\
Score-CAM & .021 & .021 & .011 & .491 & .821 & .712 & .596 & .709 \\
Lift-CAM & .021 & .021 & .011 & .500 & .820 & .710 & .593 & .704 \\
Grad-CAM & .021 & .021 & .011 & .505 & .820 & .709 & .591 & .701 \\
FG-VCE & .022 & .021 & .011 & .540 & .817 & .705 & .582 & .677 \\
IIA & .022 & .021 & .011 & .514 & .819 & .708 & .588 & .694 \\
RISE & .024 & .025 & .012 & .588 & .804 & .675 & .556 & .654 \\
EP & .023 & .024 & .012 & .577 & .805 & .680 & .554 & .661 \\
\midrule
SmoothGrad & .035 & .029 & .012 & .629 & .727 & .626 & .540 & .627 \\
Saliency & .042 & .032 & .013 & .736 & .685 & .591 & .487 & .546 \\
Int. Grad. & .046 & .032 & .014 & .817 & .675 & .597 & .464 & .485 \\
\midrule
$\text{FEI}_{\text{VM}}$ & \textbf{.021} & \underline{.019} & \underline{.009} & \underline{.412} & \underline{.833} & \underline{.749} & .671 & \underline{.758} \\
$\text{FEI}_{\text{IVM}}$ & .022 & .020 & .009 & .435 & .821 & .734 & .656 & .743 \\
$\text{FEI}_{\text{IBM}}$ & .024 & .019 & \textbf{.008} & .415 & .812 & .753 & \textbf{.689} & .757 \\
$\text{FEI}_{\text{AVM}}$ & \underline{.021} & \textbf{.019} & .009 & \textbf{.399} & \textbf{.837} & \textbf{.759} & \underline{.689} & \textbf{.766} \\
\midrule
Unconstr. EQO & .045 & .033 & .015 & 1.494 & .674 & .588 & .454 & .592 \\
\bottomrule
\end{tabular}

\vspace{6pt}
\textbf{CUB-200-2011}\\[2pt]
\begin{tabular}{lcccccccc}
\toprule
\multirow{2}{*}{Method} &
\multicolumn{4}{c}{MSE $\downarrow$} &
\multicolumn{4}{c}{Cos. $\uparrow$} \\
\cmidrule(lr){2-5} \cmidrule(lr){6-9}
& L1 & L2 & L3 & L4 & L1 & L2 & L3 & L4 \\
\midrule
Grad-CAM++ & .020 & .020 & .011 & .320 & .834 & .720 & .625 & .826 \\
Score-CAM & .021 & .020 & .011 & .317 & .834 & .721 & .625 & .828 \\
Lift-CAM & .021 & .020 & .011 & .322 & .834 & .720 & .624 & .825 \\
Grad-CAM & .021 & .020 & .011 & .326 & .834 & .719 & .623 & .822 \\
FG-VCE & .021 & .020 & .011 & .358 & .832 & .716 & .615 & .804 \\
IIA & .021 & .020 & .011 & .326 & .833 & .717 & .620 & .822 \\
RISE & .023 & .024 & .012 & .354 & .816 & .678 & .599 & .805 \\
EP & .022 & .023 & .012 & .358 & .820 & .688 & .594 & .805 \\
\midrule
SmoothGrad & .032 & .027 & .012 & .426 & .754 & .641 & .571 & .762 \\
Saliency & .039 & .031 & .014 & .611 & .704 & .595 & .506 & .648 \\
Int. Grad. & .044 & .031 & .015 & .833 & .692 & .605 & .477 & .510 \\
\midrule
$\text{FEI}_{\text{VM}}$ & \underline{.019} & \underline{.019} & \underline{.009} & \underline{.267} & \underline{.844} & \underline{.752} & \underline{.689} & \underline{.855} \\
$\text{FEI}_{\text{IVM}}$ & .021 & .020 & .009 & .295 & .831 & .734 & .670 & .840 \\
$\text{FEI}_{\text{IBM}}$ & .023 & .020 & \textbf{.009} & .280 & .822 & .750 & \textbf{.699} & .849 \\
$\text{FEI}_{\text{AVM}}$ & \textbf{.020} & \textbf{.019} & .009 & \textbf{.261} & \textbf{.847} & \textbf{.759} & .696 & \textbf{.859} \\
\midrule
Unconstr. EQO & .043 & .033 & .015 & 1.058 & .692 & .595 & .486 & .699 \\
\bottomrule
\end{tabular}
\end{table*}

\begin{table*}[t]
\centering
\caption{Layer-wise internal activation matching on VGG16. MSE$\downarrow$,
Cos.$\uparrow$. Best in \textbf{bold}, second best \underline{underlined}.}
\label{tab:supp_layerwise_vgg}
\small
\setlength{\tabcolsep}{3pt}
\renewcommand{\arraystretch}{1.02}
\textbf{ImageNet}\\[2pt]
\begin{tabular}{lcccccccccc}
\toprule
\multirow{2}{*}{Method} &
\multicolumn{5}{c}{MSE $\downarrow$} &
\multicolumn{5}{c}{Cos. $\uparrow$} \\
\cmidrule(lr){2-6} \cmidrule(lr){7-11}
& F3 & F8 & F15 & F22 & F29
& F3 & F8 & F15 & F22 & F29 \\
\midrule
Grad-CAM++ & \underline{0.762} & \underline{3.165} & \underline{8.116} & 2.438 & 0.573 & \underline{.714} & .646 & .613 & .570 & .640 \\
Score-CAM & 0.763 & 3.254 & 8.520 & 2.544 & 0.588 & .713 & .640 & .602 & .561 & .631 \\
Lift-CAM & 0.799 & 3.483 & 9.245 & 2.786 & 0.648 & .697 & .616 & .572 & .526 & .594 \\
Grad-CAM & 0.834 & 3.729 & 10.087 & 3.100 & 0.755 & .682 & .592 & .541 & .485 & .537 \\
FG-VCE & 0.805 & 3.472 & 9.132 & 2.773 & 0.742 & .693 & .609 & .565 & .514 & .536 \\
IIA & 0.813 & 3.538 & 9.372 & 2.915 & 0.765 & .691 & .605 & .563 & .510 & .531 \\
RISE & 0.784 & 3.531 & 9.672 & 2.970 & 0.731 & .708 & .623 & .574 & .523 & .561 \\
EP & 0.784 & 3.423 & 9.123 & 2.790 & 0.671 & .706 & .625 & .578 & .520 & .583 \\
\midrule
SmoothGrad & 1.733 & 6.550 & 13.692 & 3.284 & 0.862 & .505 & .468 & .473 & .452 & .491 \\
Saliency & 2.451 & 7.936 & 14.510 & 3.561 & 0.984 & .448 & .407 & .407 & .368 & .370 \\
Int. Grad. & 3.164 & 8.533 & 14.252 & 3.816 & 1.173 & .441 & .416 & .405 & .332 & .260 \\
\midrule
$\text{FEI}_{\text{VM}}$ & \textbf{0.717} & \textbf{2.986} & \textbf{7.455} & \textbf{2.134} & \textbf{0.545} & \textbf{.747} & \textbf{.696} & \textbf{.675} & \textbf{.648} & \textbf{.688} \\
$\text{FEI}_{\text{IVM}}$ & 0.840 & 3.418 & 8.344 & \underline{2.284} & \underline{0.567} & .694 & .639 & .624 & .608 & .656 \\
$\text{FEI}_{\text{IBM}}$ & 0.953 & 3.817 & 9.030 & 2.417 & 0.629 & .686 & .644 & .633 & \underline{.616} & \underline{.658} \\
$\text{FEI}_{\text{AVM}}$ & 1.291 & 5.247 & 11.666 & 3.313 & 0.952 & .679 & \underline{.648} & \underline{.647} & .610 & .628 \\
\midrule
Unconstr. EQO & 2.823 & 9.981 & 20.796 & 7.081 & 3.619 & .458 & .408 & .374 & .309 & .344 \\
\bottomrule
\end{tabular}

\vspace{6pt}
\textbf{CUB-200-2011}\\[2pt]
\begin{tabular}{lcccccccccc}
\toprule
\multirow{2}{*}{Method} &
\multicolumn{5}{c}{MSE $\downarrow$} &
\multicolumn{5}{c}{Cos. $\uparrow$} \\
\cmidrule(lr){2-6} \cmidrule(lr){7-11}
& F3 & F8 & F15 & F22 & F29
& F3 & F8 & F15 & F22 & F29 \\
\midrule
Grad-CAM++ & \underline{0.549} & \underline{1.664} & \underline{3.315} & \underline{0.960} & \underline{0.355} & \underline{.730} & \underline{.664} & .643 & .674 & \underline{.813} \\
Score-CAM & 0.553 & 1.743 & 3.567 & 1.017 & 0.376 & .729 & .655 & .630 & .666 & .804 \\
Lift-CAM & 0.568 & 1.830 & 3.799 & 1.093 & 0.410 & .719 & .637 & .607 & .639 & .783 \\
Grad-CAM & 0.585 & 1.929 & 4.055 & 1.179 & 0.461 & .710 & .621 & .587 & .616 & .759 \\
FG-VCE & 0.580 & 1.852 & 3.818 & 1.145 & 0.535 & .710 & .623 & .591 & .614 & .720 \\
IIA & 0.579 & 1.879 & 3.892 & 1.156 & 0.491 & .712 & .623 & .592 & .621 & .745 \\
RISE & 0.577 & 1.976 & 4.245 & 1.220 & 0.478 & .718 & .625 & .588 & .618 & .754 \\
EP & 0.564 & 1.827 & 3.804 & 1.133 & 0.436 & .723 & .641 & .607 & .623 & .772 \\
\midrule
SmoothGrad & 0.982 & 2.949 & 5.245 & 1.516 & 0.596 & .588 & .535 & .549 & .569 & .708 \\
Saliency & 1.507 & 3.937 & 5.890 & 1.871 & 0.824 & .505 & .449 & .486 & .482 & .588 \\
Int. Grad. & 2.157 & 4.842 & 6.552 & 2.351 & 1.438 & .450 & .392 & .412 & .340 & .266 \\
\midrule
$\text{FEI}_{\text{VM}}$ & \textbf{0.525} & \textbf{1.654} & \textbf{3.295} & \textbf{0.938} & \textbf{0.349} & \textbf{.754} & \textbf{.700} & \textbf{.686} & \textbf{.710} & \textbf{.822} \\
$\text{FEI}_{\text{IVM}}$ & 0.579 & 1.824 & 3.617 & 1.000 & 0.371 & .720 & .657 & .647 & .687 & .807 \\
$\text{FEI}_{\text{IBM}}$ & 0.625 & 1.968 & 3.817 & 1.061 & 0.402 & .713 & .659 & \underline{.656} & \underline{.688} & .804 \\
$\text{FEI}_{\text{AVM}}$ & 0.845 & 2.799 & 5.129 & 1.679 & 0.649 & .689 & .641 & .649 & .639 & .758 \\
\midrule
Unconstr. EQO & 1.991 & 5.518 & 8.996 & 3.878 & 3.437 & .467 & .392 & .404 & .376 & .480 \\
\bottomrule
\end{tabular}
\end{table*}

\section{Architectural Sensitivity}
\label{app:arch}
\label{sec:archi}

ClipGrad applies the same local gradient rule across architectures, while the
measured behavior of asymmetric variants reflects the interaction among
network topology, activation function, and normalization. We compare
sequential ReLU networks, multi-branch ReLU networks, and GELU--LayerNorm
ConvNeXt to characterize this interaction.

\subsection{Constraint Propagation in Sequential vs. Multi-Branch Architectures}

In a sequential architecture, each layer receives the preceding layer's
output directly. A residual block also carries an identity branch:
\begin{equation}
h_{\ell+1} = \text{ReLU}(f_{\ell}^{\text{residual}}(h_{\ell}) + h_{\ell})
\end{equation}
The identity and residual branches expose the next block to a mixture of
constrained and bypassed signals.  This structural difference motivates the
architecture-stratified results below.

\subsection{AVM: Asymmetric Constraint on Activation Decreases}

FEI$_{\text{AVM}}$ blocks decreases below the clean activation while
permitting increases (\cref{eq:avm}).  On the sequential AlexNet and VGG16
models, AVM produces edge-emphasized maps and all-black-input target-induction
rates of 93.8\% and 95.2\%, respectively (\cref{tab:defense}).  These results
are consistent with activation increases propagating across consecutive
layers when decreases are blocked.

On ResNet50 and GoogLeNet, AVM has target-induction rates of 0.1\% and 0\%,
respectively (\cref{tab:defense}), whereas it reaches 20.7\% on residual
ConvNeXt-Tiny. Residual structure can limit how strongly a one-sided
constraint determines the combined activation, but does not determine the
outcome alone: smooth GELU activations and LayerNorm alter how the directional
constraint propagates. The three designs together show that asymmetric
constraint behavior emerges from the complete computational block rather than
from branch structure in isolation.

\subsection{IBM/IVM: Asymmetric Constraint on Activation Increases}

FEI$_{\text{IBM}}$ and FEI$_{\text{IVM}}$ constrain activation increases
(\cref{eq:ibm,eq:ivm}).  IVM caps increases above the clean value, while IBM
applies the binary version to units inactive on the clean input.  Both yield
all-black-input target-induction rates between 0\% and 0.4\% across the tested
ReLU architectures (\cref{tab:defense}). On ConvNeXt-Tiny, their sign- and
value-based forms remain effective at 0.4\% and 0.3\%, respectively. Their maps are more localized on the sequential models
and more diffuse on the multi-branch models.

\subsection{VM: Bidirectional Constraint}

FEI$_{\text{VM}}$ blocks movement away from the clean value in both
directions (\cref{eq:vm}). It gives a 0\%--0.1\% all-black-input
target-induction rate on the four ReLU convolutional architectures and 0.1\%
on ConvNeXt-Tiny (\cref{tab:defense}), making it the most consistent variant under
this check.

\subsection{Practical Guidance}

The results support VM as the architecture-stable default and IBM as the
output-oriented operating point with the highest constrained insertion scores
(\cref{tab:external,tab:convnext}). AVM reveals how a directional constraint
interacts with the full block design, changing sharply across both topology
and activation-normalization regimes.

\section{Layer-wise Activation-Matching Analysis}
\label{app:layerwise}
\label{app:layer_wise}
\label{app:activation}

\paragraph{Rationale and scope.}
We motivate activation preservation from three related
observations.  First, if layerwise activations remain close to their clean
values, each deterministic downstream transformation receives a similar
state, biasing the perturbed computation toward the clean trajectory.  Second,
mechanistic work represents features as directions in overlapping hidden
spaces \cite{olah2020zoom,cunningham2023sparse,elhage2022toy}; large deviations
can therefore change which task-relevant directions are active.  Third,
adversarial-robustness studies show that small representation changes can
accompany large semantic changes \cite{ilyas2019adversarial,xu2019interpreting}.
These points motivate activation matching. MSE and cosine similarity establish
geometric preservation of the internal trajectory, while the independently
constructed clean-support intervention establishes functional dependence.

The aggregate final-block comparison appears in \cref{tab:activation}; the
fuller layerwise and cross-dataset values follow.

\Cref{fig:supp_layerwise_internal} shows the same broad pattern across the four
architecture--dataset combinations. Gradient-based methods and unconstrained
EQO have higher MSE and lower cosine similarity than the constrained variants
and several comparison methods.

\textbf{ResNet50-specific patterns.} The early-layer MSE profiles are tightly
clustered in ResNet50, and \cref{tab:supp_layerwise_resnet} supplies the values
that are difficult to distinguish in the plot.  This compression may reflect
the combined effects of residual connections and normalization; the present
measurements do not isolate either mechanism.

At Layer4, MSE values are roughly an order of magnitude larger and the methods
separate more clearly.  Constrained FEI variants maintain lower MSE, while
unconstrained EQO has the largest deviation.  This observation is compatible
with the final feature block carrying more class-discriminative information;
the activation measurements alone do not isolate the cause.

\textbf{Cross-dataset consistency.} The same descriptive pattern appears on CUB-200-2011 and ImageNet (\cref{tab:supp_layerwise_resnet}): early-layer MSE values are compressed, Layer4 differences are larger, and the constrained variants more closely match clean activations.

\textbf{Implications for evaluation.} The compressed scale in ResNet50's early layers can obscure method differences in \cref{fig:supp_layerwise_internal}; \cref{tab:supp_layerwise_resnet} provides the numerical values. Together with the clean-support intervention, these measurements establish consistent internal preservation across layers.

\section{Derivation of Selective Gradient Clipping (ClipGrad) Variants}
\label{app:derivation}
\label{app:clipgrad_details}

ClipGrad can be derived from an insertion objective combined with an
activation-matching objective.

\subsection{Value Matching (VM) Variants}
We define the total loss as:
\begin{equation}
\label{eq:total_loss}
\mathcal{L}_{\text{total}} = \mathcal{L}_{\text{ins}} + \mathcal{L}_{\text{int}}
\end{equation}

The external gradient is:
\begin{equation}
\label{eq:external_gradient}
\gamma_h = \frac{\partial \mathcal{L}_{\text{ins}}}{\partial \tilde{h}^\ell}
\end{equation}

The internal loss is designed to match the relevant parts of internal activations, where relevance is measured by saliency:
\begin{equation}
\label{eq:internal_loss_general}
\mathcal{L}_{\text{int}} = \beta_{\text{int}} \sum_{j} s^\ell[j] \cdot d(h^\ell[j], \tilde{h}^\ell[j])
\end{equation}

Here $s^\ell[j]$ is the relevance of unit $j$ at layer $\ell$,
$d(h^\ell[j],\tilde h^\ell[j])$ measures deviation from its clean activation,
and $\beta_{\text{int}}$ balances the two objectives.  We use the $L_1$
distance $d(a,b)=|a-b|$ and approximate relevance by
$s^\ell[j]\approx|\gamma_h[j]|$.

Here $s^\ell$ is treated as a stop-gradient saliency estimate: it
determines which coordinates matter for matching, but gradients are not
propagated through $s^\ell$ itself. This is what allows the saliency factor
to cancel cleanly below.

This yields the internal loss:
\begin{equation}
\label{eq:internal_loss}
\mathcal{L}_{\text{int}} = \beta_{\text{int}} \sum_{j} |\gamma_h[j]| \cdot |h^\ell[j] - \tilde{h}^\ell[j]|
\end{equation}

The gradient of the internal loss is:
\begin{equation}
\label{eq:internal_gradient}
\frac{\partial \mathcal{L}_{\text{int}}}{\partial \tilde{h}^\ell[j]} = \beta_{\text{int}} \cdot |\gamma_h[j]| \cdot \text{sign}(\tilde{h}^\ell[j] - h^\ell[j])
\end{equation}

The total gradient is:
\begin{equation}
\label{eq:total_gradient}
\tilde{\gamma}_h[j] = \gamma_h[j] + \beta_{\text{int}} \cdot |\gamma_h[j]| \cdot \text{sign}(\tilde{h}^\ell[j] - h^\ell[j])
\end{equation}

\textbf{Derivation of Value Matching (VM)}

Setting $\beta_{\text{int}} = 1$ and expanding by cases:
\begin{equation}
\label{eq:vm_cases}
\tilde{\gamma}_h[j] = \begin{cases}
\gamma_h[j] + \gamma_h[j] \cdot (+1) = 2\gamma_h[j] & \text{if } \gamma_h[j] > 0 \\
&\land \tilde{h}^\ell[j] > h^\ell[j] \\
\gamma_h[j] + |\gamma_h[j]| \cdot (-1) = 2\gamma_h[j] & \text{if } \gamma_h[j] \leq 0 \\
&\land \tilde{h}^\ell[j] \leq h^\ell[j] \\
\gamma_h[j] + \gamma_h[j] \cdot (-1) = 0 & \text{if } \gamma_h[j] > 0\\
& \land \tilde{h}^\ell[j] \leq h^\ell[j] \\
\gamma_h[j] + |\gamma_h[j]| \cdot (+1) = 0 & \text{if } \gamma_h[j] \leq 0 \\
&\land \tilde{h}^\ell[j] > h^\ell[j]
\end{cases}
\end{equation}

Absorbing the factor of 2 into the learning rate, we obtain the VM clipping rule:
\begin{equation}
\tilde{\gamma}_h[j] =
\begin{cases}
0 & \text{if } (\gamma_h[j] \leq 0 \land \tilde{h}^\ell[j] > h^\ell[j]) \\
  & \quad \lor (\gamma_h[j] > 0 \land \tilde{h}^\ell[j] \leq h^\ell[j]) \\
\gamma_h[j] & \text{otherwise.}
\end{cases}
\end{equation}

\textbf{Analysis.} VM clips the gradient whenever the insertion objective
would move $\tilde{h}^\ell[j]$ farther from its clean value $h^\ell[j]$.
Although the objective in \cref{eq:internal_loss_general} is
relevance-weighted, cancellation produces the binary directional rule in
\cref{eq:vm_cases}.  The rule applies a uniform clean-activation filter to all
coordinates.

In gradient descent ($\tilde{h}^\ell \leftarrow \tilde{h}^\ell - \alpha \tilde{\gamma}_h$):
\begin{itemize}
 \item When $\gamma_h[j] > 0$ and $\tilde{h}^\ell[j] \le h^\ell[j]$: The optimizer wants to decrease $\tilde{h}^\ell[j]$, but it is already at or below $h^\ell[j]$. Clipping prevents further suppression.
 	\item When $\gamma_h[j] \leq 0$ and $\tilde{h}^\ell[j] > h^\ell[j]$: The optimizer wants to increase $\tilde{h}^\ell[j]$, but it is already above $h^\ell[j]$. Clipping prevents further amplification.
\end{itemize}

\textbf{Derivation of Activated Value Matching (AVM)}

AVM uses only Case 3 from the VM derivation:
\begin{equation}
\label{eq:avm}
\tilde{\gamma}_h[j] =
\begin{cases}
0 & \text{if } \gamma_h[j] > 0 \land \tilde{h}^\ell[j] \leq h^\ell[j] \\
\gamma_h[j] & \text{otherwise.}
\end{cases}
\end{equation}

\textbf{Analysis.} AVM prevents suppression of activations below their original values. It blocks gradient descent when $\gamma_h[j] > 0$ (wants to decrease) and $\tilde{h}^\ell[j] \le h^\ell[j]$ (already at or below original). This allows increases but prevents further suppression of already-suppressed neurons.

\textbf{Derivation of Inactivated Value Matching (IVM)}

IVM uses only Case 4 from the VM derivation:
\begin{equation}
\label{eq:ivm}
\tilde{\gamma}_h[j] =
\begin{cases}
0 & \text{if } \gamma_h[j] \leq 0 \land \tilde{h}^\ell[j] > h^\ell[j] \\
\gamma_h[j] & \text{otherwise.}
\end{cases}
\end{equation}

\textbf{Analysis.} IVM prevents increases above clean activation values. It
blocks gradient descent when $\gamma_h[j]\leq0$ would increase a coordinate
that already satisfies $\tilde{h}^\ell[j]>h^\ell[j]$.  It allows decreases
while preventing further amplification.

\subsection{Binary Matching (BM) Variants}

The \textbf{Binary Matching} variants are derived from a binary-aware distance function that matches only the \textit{sign} of the activation rather than its exact value.

\textbf{Binary Distance Function}

Instead of the $L_1$ distance used for VM, we define a binary-aware distance function:
\begin{align}
    \label{eq:binary_distance}
d(h^\ell[j], \tilde{h}^\ell[j]) = &(1 - B(h^\ell[j]))(\tilde{h}^\ell[j] - h^\ell[j]) + \\
&B(h^\ell[j])(h^\ell[j] - \tilde{h}^\ell[j])
\end{align}

where $B(h^\ell[j])$ is a binary indicator:
\begin{equation}
\label{eq:binary_indicator}
B(h^\ell[j]) = \begin{cases}
0 & \text{if } h^\ell[j] \le 0 \text{ (inactive)} \\
1 & \text{if } h^\ell[j] > 0 \text{ (active)}
\end{cases}
\end{equation}

The total gradient becomes:
\begin{equation}
\label{eq:binary_total_gradient}
\tilde{\gamma}_h[j] = \gamma_h[j] + |\gamma_h[j]| \cdot (1 - 2B(h^\ell[j]))
\end{equation}

\textbf{Derivation of Binary Matching (BM)}

Setting $\beta_{\text{int}} = 1$, we have
\begin{equation}
\label{eq:bm_full_cases}
\tilde{\gamma}_h[j] = \begin{cases}
\gamma_h[j] + |\gamma_h[j]| = 0 & \text{if } \gamma_h[j] \le 0 \land h^\ell[j] \le 0 \\
\gamma_h[j] + |\gamma_h[j]| = 2\gamma_h[j] & \text{if } \gamma_h[j] > 0 \land h^\ell[j] \le 0 \\
\gamma_h[j] - |\gamma_h[j]| = 2\gamma_h[j] & \text{if } \gamma_h[j] \le 0 \land h^\ell[j] > 0 \\
\gamma_h[j] - |\gamma_h[j]| = 0 & \text{if } \gamma_h[j] > 0 \land h^\ell[j] > 0
\end{cases}
\end{equation}

Absorbing the factor of 2 into the learning rate, we obtain the \textbf{Binary Matching (BM)} rule:
\begin{equation}
\label{eq:bm_rule}
\tilde{\gamma}_h[j] = \begin{cases}
0 & \text{if } (\gamma_h[j] \le 0 \land h^\ell[j] \le 0) \\
&\lor (\gamma_h[j] > 0 \land h^\ell[j] > 0) \\
\gamma_h[j] & \text{otherwise}
\end{cases}
\end{equation}

\textbf{Analysis.} BM clips the gradient in two scenarios:
\begin{itemize}
    \item When $\gamma_h[j] \le 0$ and $h^\ell[j] \le 0$, clipping prevents an increase of a clean-inactive unit.
    \item When $\gamma_h[j] > 0$ and $h^\ell[j] > 0$, clipping prevents a decrease of a clean-active unit.
\end{itemize}

BM enforces binary state preservation: inactive neurons stay inactive, active neurons stay active.

\textbf{Derivation of Inactivated Binary Matching (IBM)}

IBM uses only the first case from the BM derivation (inactive neurons):
\begin{equation}
\tilde{\gamma}_h[j] =
\begin{cases}
0 & \text{if } \gamma_h[j] \leq 0 \land h^\ell[j] \leq 0 \\
\gamma_h[j] & \text{otherwise.}
\end{cases}
\end{equation}

\textbf{Analysis.} IBM prevents activation of units that are inactive on the
clean input.  When $\gamma_h[j]\leq0$ would increase a coordinate with
$h^\ell[j]\leq0$, the gradient is clipped.  IBM is therefore a relaxed version
of IVM that constrains only clean-inactive units.
Its positive-gradient gating is related in spirit to Guided Backpropagation
\cite{springenberg2014striving}, but IBM applies the gate inside attribution
optimization to preserve clean activation state rather than to produce a
single backward visualization.

\textbf{Derivation of Activated Binary Matching (ABM)}

ABM uses only the second case from the BM derivation (active neurons):
\begin{equation}
\label{eq:abm_rule}
\tilde{\gamma}_h[j] = \begin{cases}
0 & \text{if } \gamma_h[j] > 0 \land h^\ell[j] > 0 \\
\gamma_h[j] & \text{otherwise}
\end{cases}
\end{equation}

\textbf{Analysis.}
ABM prevents decreases of units that are active on the clean input.  When
$\gamma_h[j]>0$ would decrease a coordinate with $h^\ell[j]>0$, the gradient
is clipped.

\textbf{Choice of binary variant.}
Binary Matching constrains the active or inactive state of each unit rather
than its exact value.  BM applies both directions, ABM constrains clean-active
units, and IBM constrains clean-inactive units.  We use IBM as the one-sided
binary operating point because it leaves clean-active units free to vary.  In
the experiments, this looser rule retains more insertion than VM while the
clean-support and stability measurements quantify the corresponding tradeoff.

\subsection{Summary}

The ClipGrad variants provide different levels of constraint on activation modifications:

\paragraph{Value Matching Family (L1 Distance):}
\begin{itemize}
    \item \textbf{VM (Value Matching)}: Strict bidirectional constraint that prevents both suppression and amplification away from original values. Combines AVM and IVM. This variant provides the strongest activation-preservation proxy.
    \item \textbf{AVM (Activated Value Matching)}: One-sided constraint that prevents suppression below clean values while allowing amplification.
    \item \textbf{IVM (Inactivated Value Matching)}: One-sided constraint that prevents amplification above clean values while allowing suppression.
\end{itemize}

\paragraph{Binary Matching Family (Binary Distance):}
\begin{itemize}
    \item \textbf{BM (Binary Matching)}: Strict binary state preservation that prevents both activation of inactive neurons and deactivation of active neurons. Combines IBM and ABM.
    \item \textbf{IBM (Inactivated Binary Matching)}: One-sided constraint that prevents activation of clean-inactive units ($h^\ell[j] \le 0$).
    \item \textbf{ABM (Activated Binary Matching)}: One-sided constraint that prevents deactivation of clean-active units ($h^\ell[j] > 0$).
\end{itemize}

\paragraph{Key Difference.}
VM variants preserve the actual activation values using an $L_1$ distance, while BM variants enforce state preservation by checking only whether the neuron was originally active or inactive.

\paragraph{Saliency Collapse and Interpretation.}
The derivation begins with the relevance-weighted objective in
\cref{eq:internal_loss_general}.  The relevance factor cancels against the
external-gradient magnitude, yielding uniform directional clipping rules.
Gradient magnitudes still determine the retained updates.  ClipGrad therefore
filters update direction using clean activation values; the clean-support
intervention separately measures pathway dependence.

% Page 1: VGG16 part 1 (6 rows with caption)
\begin{figure*}[!t]
\centering
\setlength\tabcolsep{1pt}

\begin{tabular}{@{}c@{}ccccc@{}}
\multicolumn{6}{c}{\textbf{VGG16-ImageNet}} \\
& Brittany spaniel & indri & library & assault rifle & chain saw \\
\rotatebox{90}{Original Image} &
\includegraphics[width=.18\linewidth]{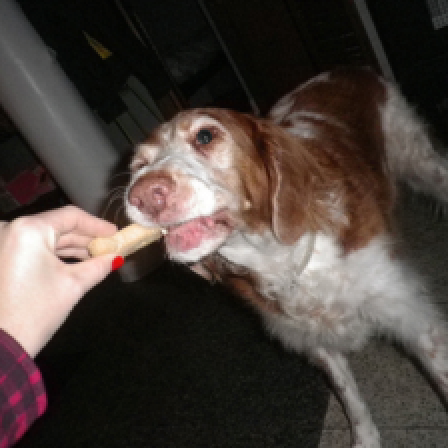} &
\includegraphics[width=.18\linewidth]{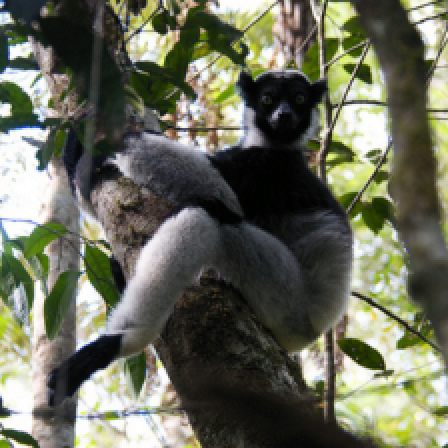} &
\includegraphics[width=.18\linewidth]{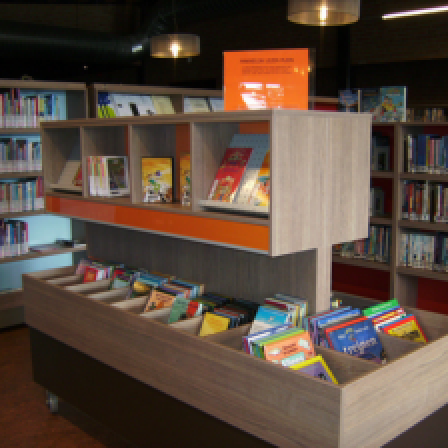} &
\includegraphics[width=.18\linewidth]{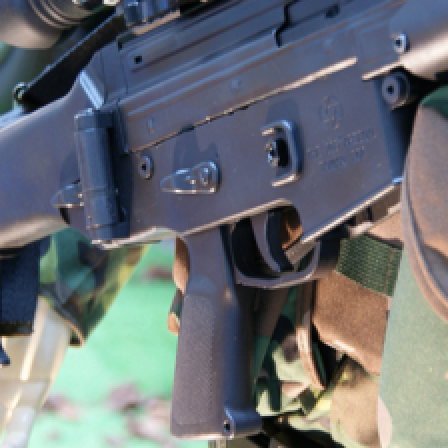} &
\includegraphics[width=.18\linewidth]{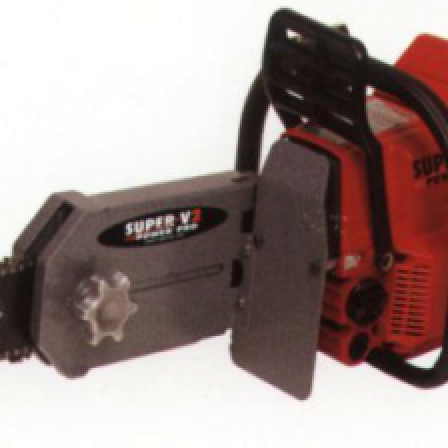}\\[2pt]

\rotatebox{90}{$\text{FEI}_{\text{IBM}}$} &
\includegraphics[width=.18\linewidth]{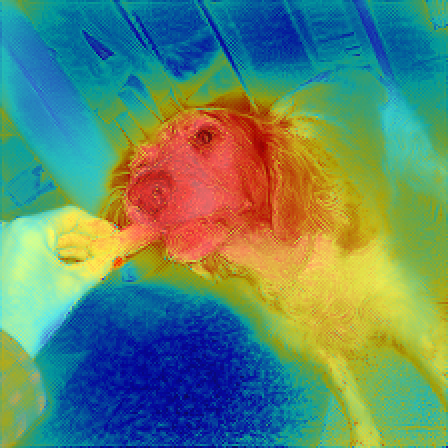} &
\includegraphics[width=.18\linewidth]{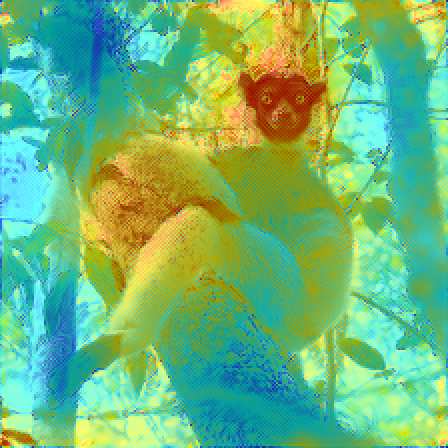} &
\includegraphics[width=.18\linewidth]{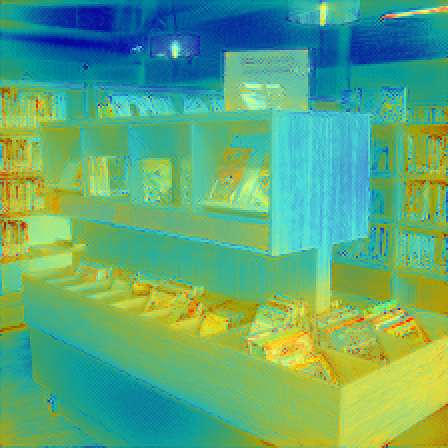} &
\includegraphics[width=.18\linewidth]{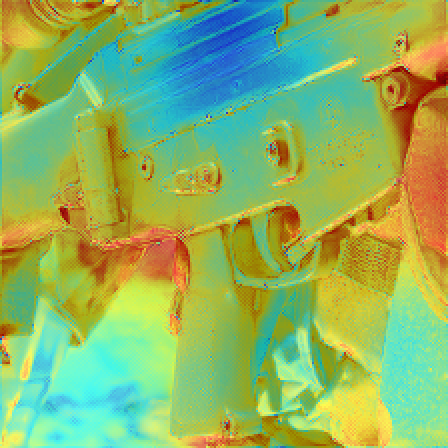} &
\includegraphics[width=.18\linewidth]{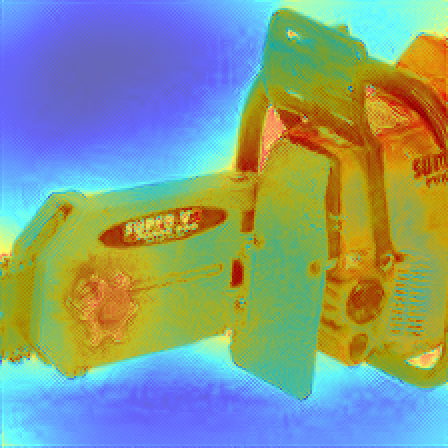} \\[2pt]

\rotatebox{90}{$\text{FEI}_{\text{IVM}}$} &
\includegraphics[width=.18\linewidth]{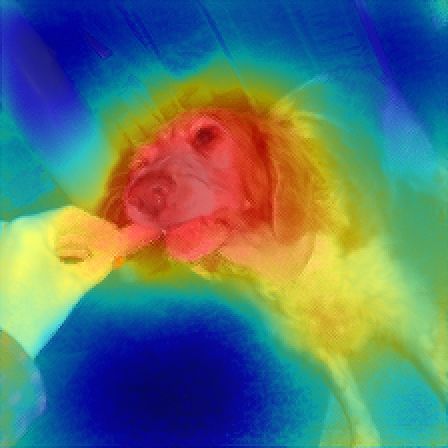} &
\includegraphics[width=.18\linewidth]{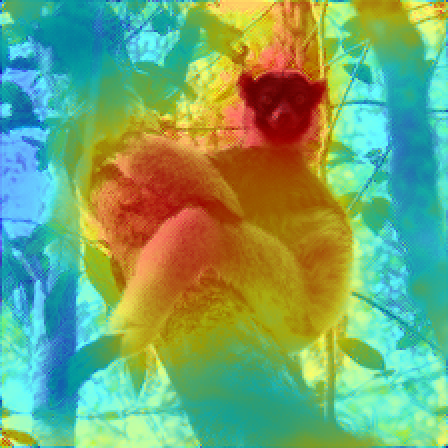} &
\includegraphics[width=.18\linewidth]{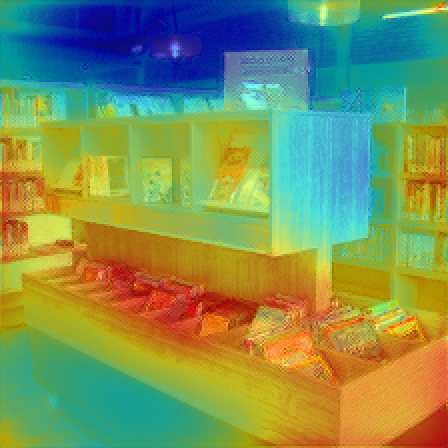} &
\includegraphics[width=.18\linewidth]{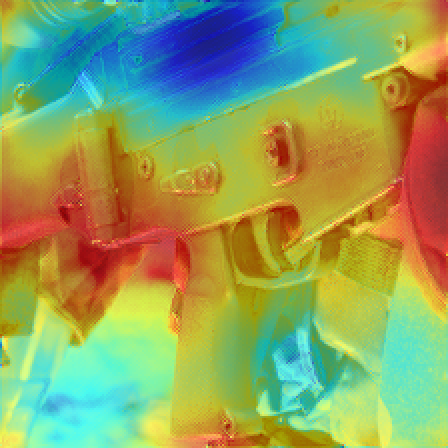} &
\includegraphics[width=.18\linewidth]{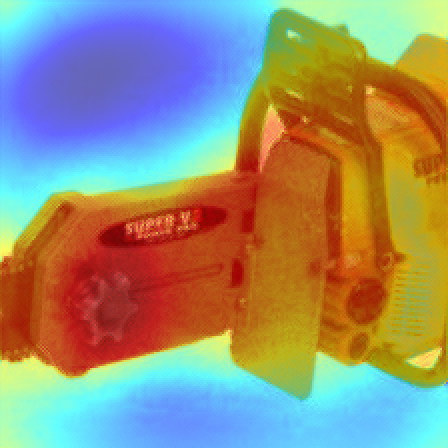} \\[2pt]

\rotatebox{90}{$\text{FEI}_{\text{AVM}}$} &
\includegraphics[width=.18\linewidth]{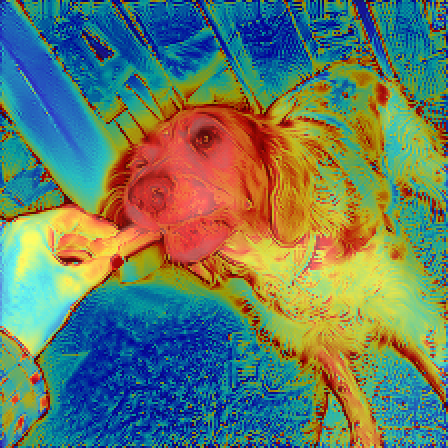} &
\includegraphics[width=.18\linewidth]{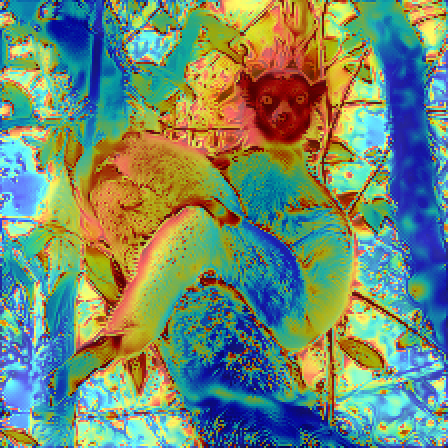} &
\includegraphics[width=.18\linewidth]{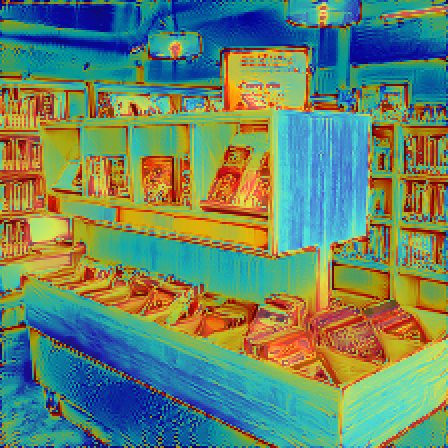} &
\includegraphics[width=.18\linewidth]{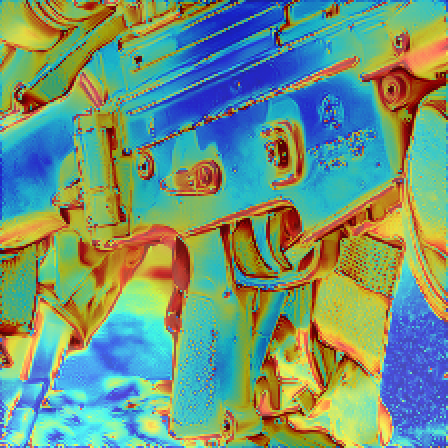} &
\includegraphics[width=.18\linewidth]{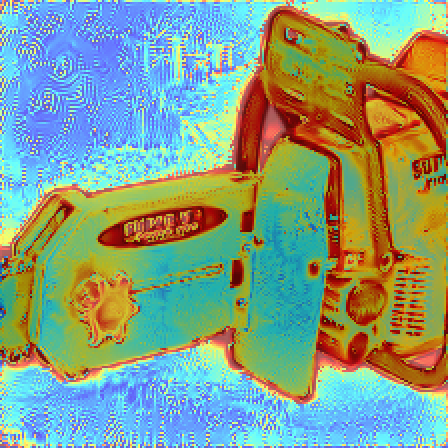} \\[2pt]

\rotatebox{90}{$\text{FEI}_{\text{VM}}$} &
\includegraphics[width=.18\linewidth]{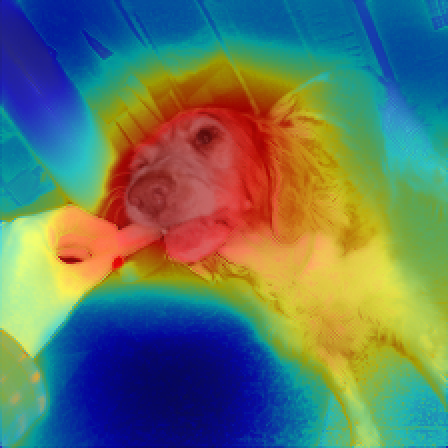} &
\includegraphics[width=.18\linewidth]{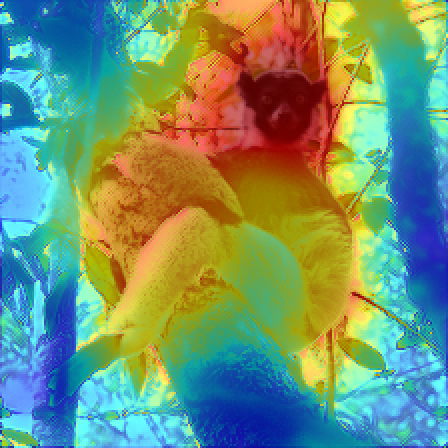} &
\includegraphics[width=.18\linewidth]{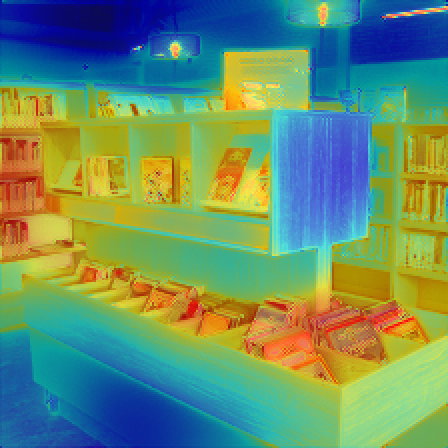} &
\includegraphics[width=.18\linewidth]{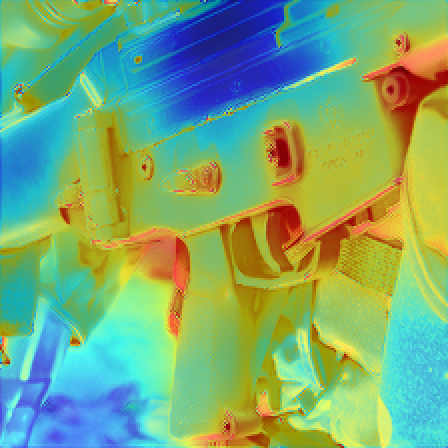} &
\includegraphics[width=.18\linewidth]{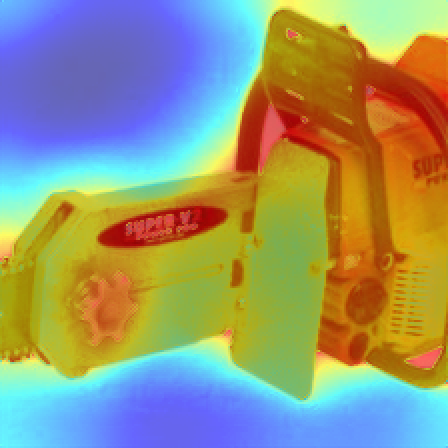} \\[2pt]

\rotatebox{90}{Unconstr. EQO} &
\includegraphics[width=.18\linewidth]{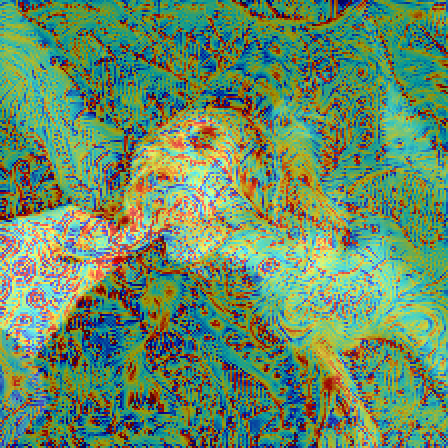} &
\includegraphics[width=.18\linewidth]{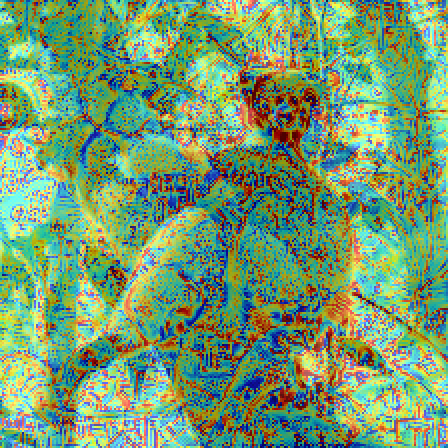} &
\includegraphics[width=.18\linewidth]{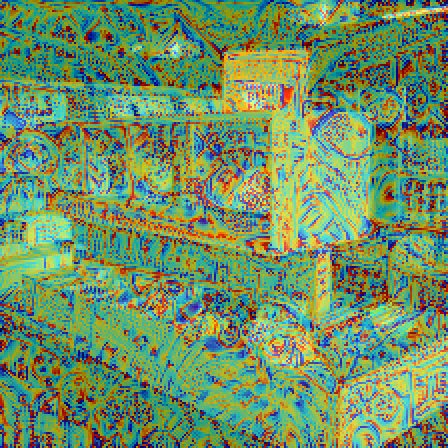} &
\includegraphics[width=.18\linewidth]{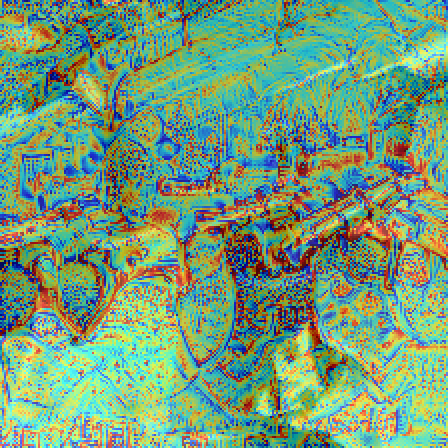} &
\includegraphics[width=.18\linewidth]{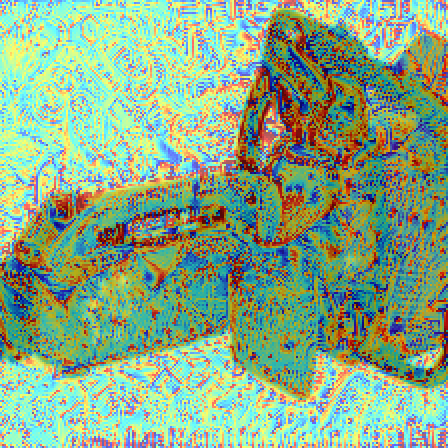} \\
\end{tabular}

\caption{\textbf{Additional visual comparison.} VGG16/ImageNet and
ResNet50/ImageNet examples. Continued on next page.}
\label{fig:supp_additional_vis}
\end{figure*}

% Page 2: VGG16 part 2 (6 rows with short caption)
\begin{figure*}[!t]
\ContinuedFloat
\centering
\setlength\tabcolsep{1pt}

\begin{tabular}{@{}c@{}ccccc@{}}
& Brittany spaniel & indri & library & assault rifle & chain saw \\
\rotatebox{90}{EP} &
\includegraphics[width=.18\linewidth]{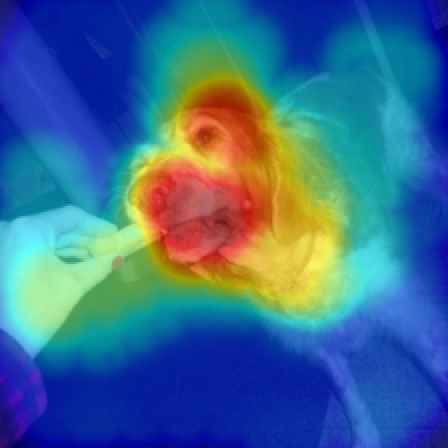} &
\includegraphics[width=.18\linewidth]{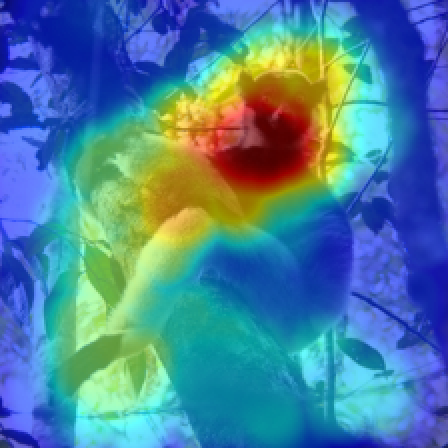} &
\includegraphics[width=.18\linewidth]{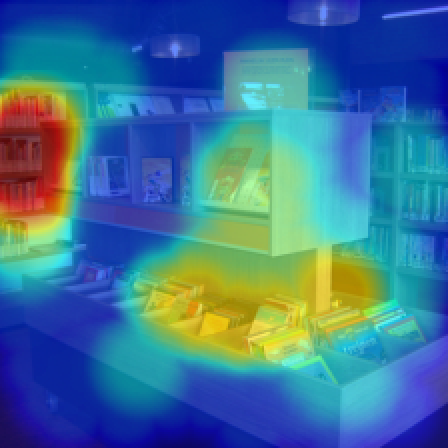} &
\includegraphics[width=.18\linewidth]{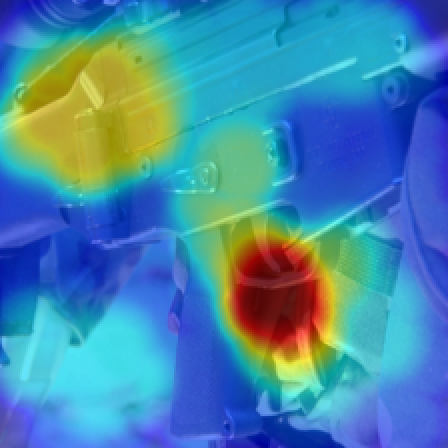} &
\includegraphics[width=.18\linewidth]{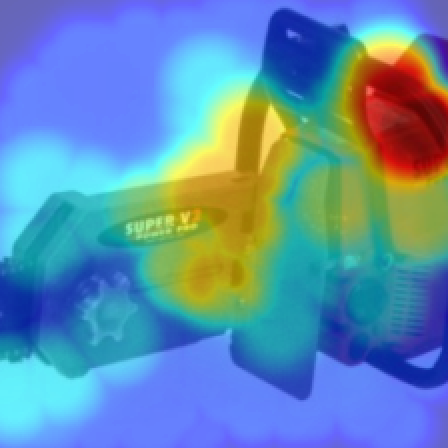} \\[2pt]

\rotatebox{90}{RISE} &
\includegraphics[width=.18\linewidth]{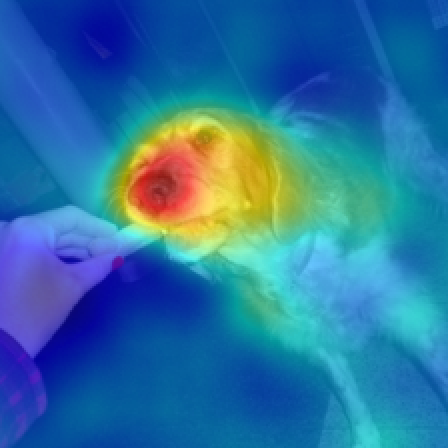} &
\includegraphics[width=.18\linewidth]{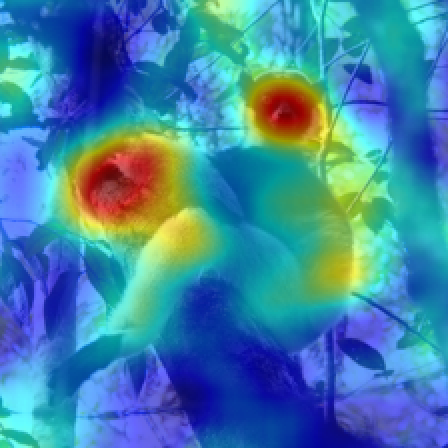} &
\includegraphics[width=.18\linewidth]{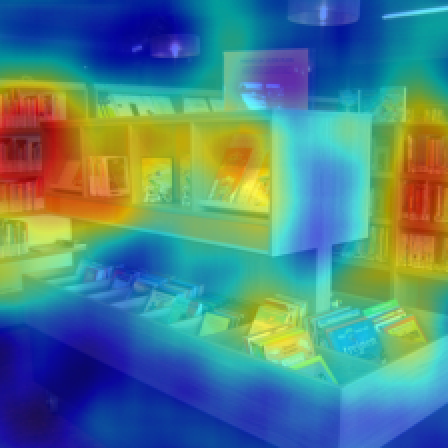} &
\includegraphics[width=.18\linewidth]{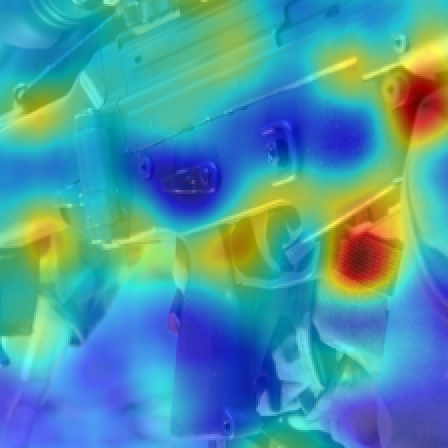} &
\includegraphics[width=.18\linewidth]{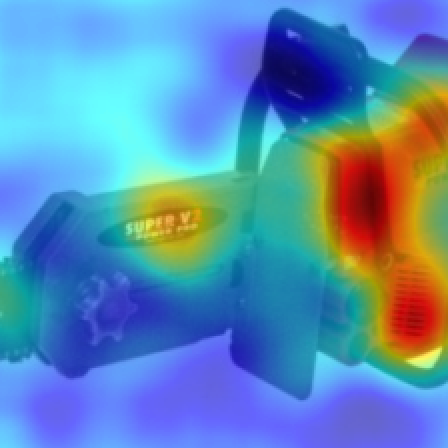} \\[2pt]

\rotatebox{90}{IIA} &
\includegraphics[width=.18\linewidth]{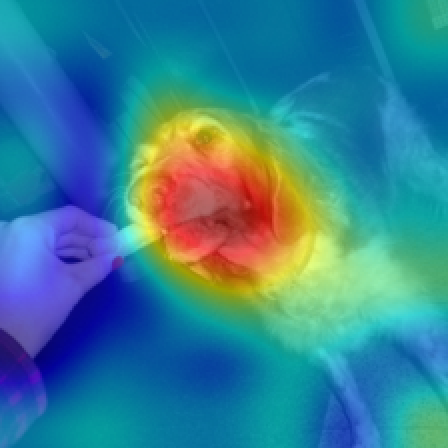} &
\includegraphics[width=.18\linewidth]{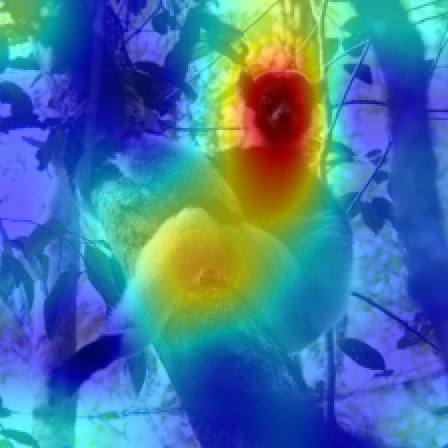} &
\includegraphics[width=.18\linewidth]{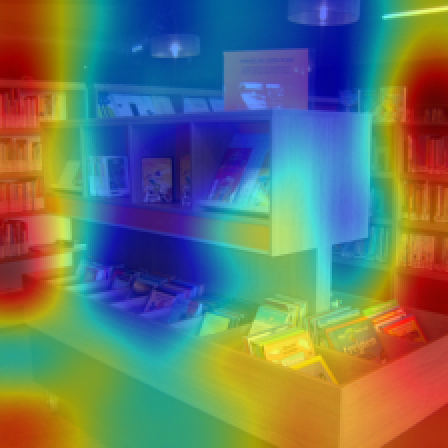} &
\includegraphics[width=.18\linewidth]{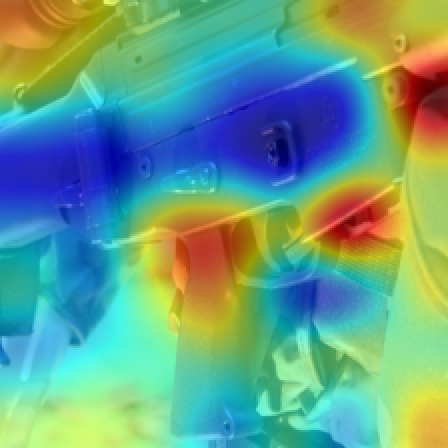} &
\includegraphics[width=.18\linewidth]{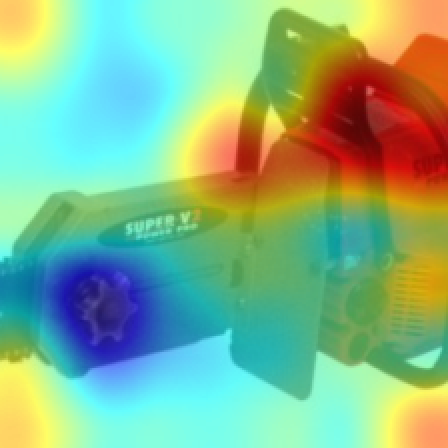} \\[2pt]

\rotatebox{90}{Lift-CAM} &
\includegraphics[width=.18\linewidth]{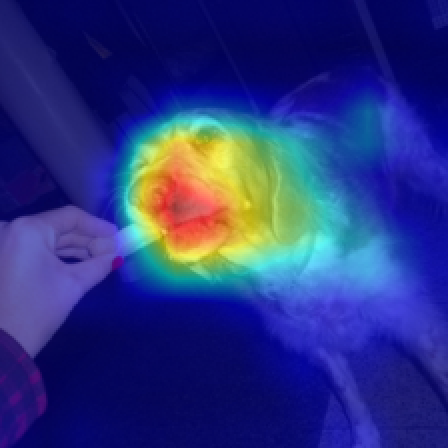} &
\includegraphics[width=.18\linewidth]{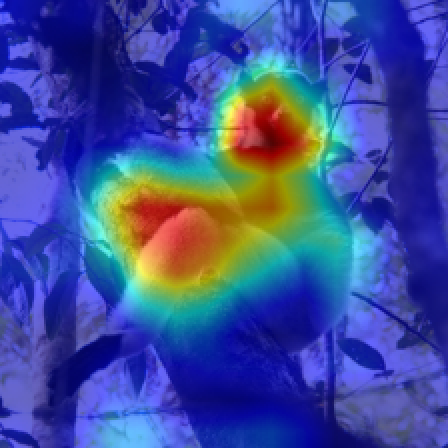} &
\includegraphics[width=.18\linewidth]{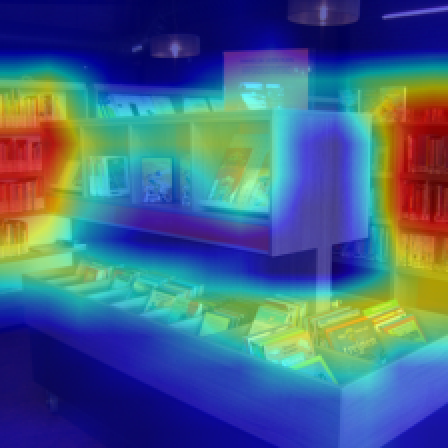} &
\includegraphics[width=.18\linewidth]{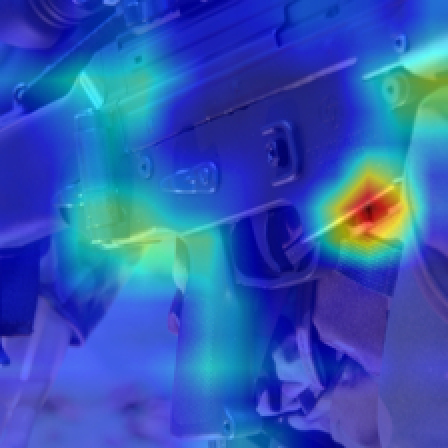} &
\includegraphics[width=.18\linewidth]{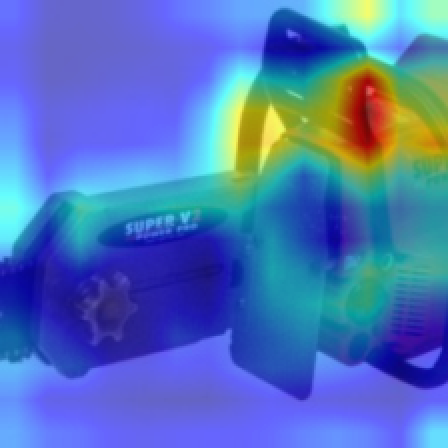} \\[2pt]

\rotatebox{90}{FG-VCE} &
\includegraphics[width=.18\linewidth]{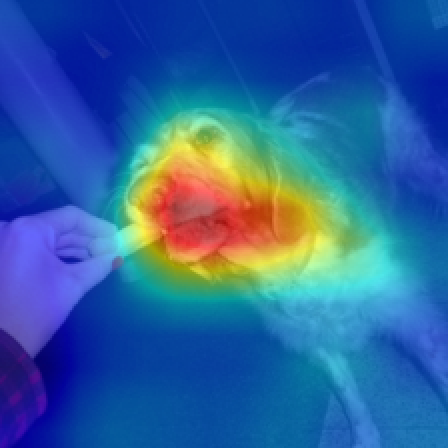} &
\includegraphics[width=.18\linewidth]{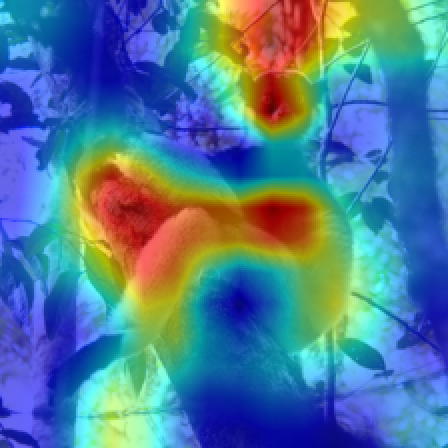} &
\includegraphics[width=.18\linewidth]{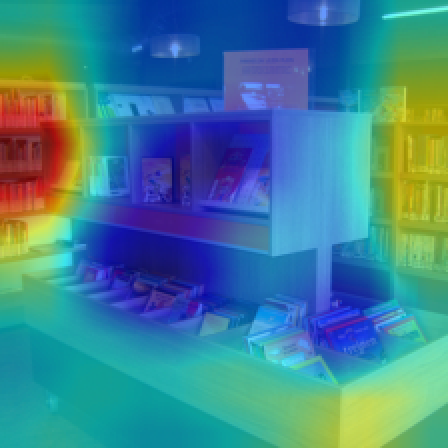} &
\includegraphics[width=.18\linewidth]{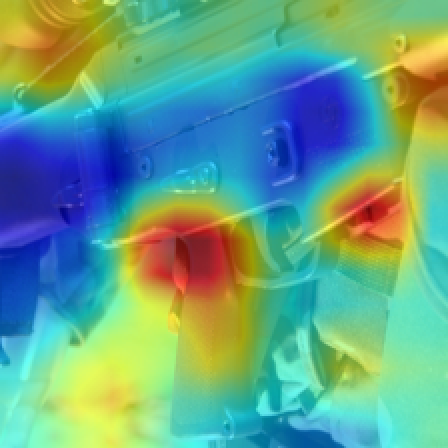} &
\includegraphics[width=.18\linewidth]{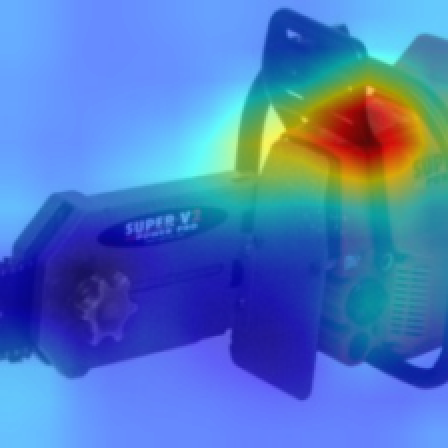} \\[4pt]

\multicolumn{6}{c}{\textbf{ResNet50-ImageNet}} \\
& wolf spider & reflex camera & pill bottle & terrapin & candle \\
\rotatebox{90}{Original Image} &
\includegraphics[width=.18\linewidth]{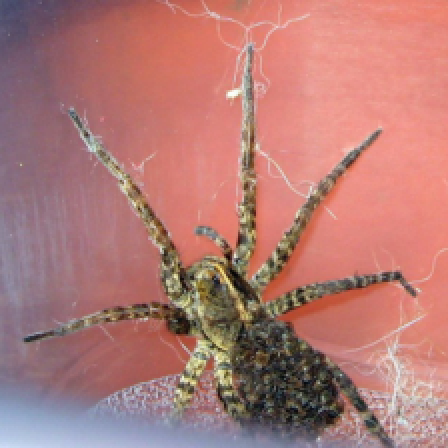} &
\includegraphics[width=.18\linewidth]{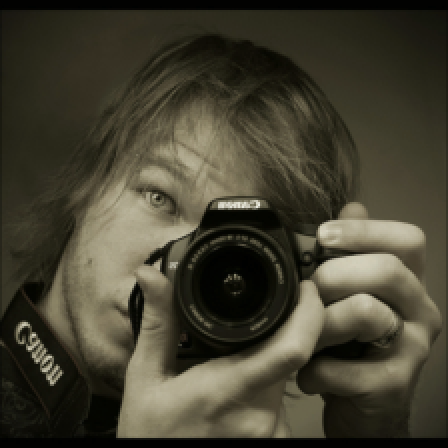} &
\includegraphics[width=.18\linewidth]{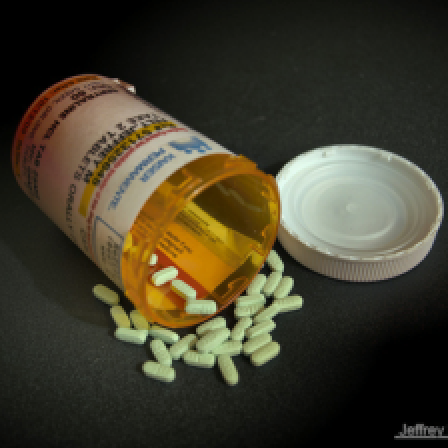} &
\includegraphics[width=.18\linewidth]{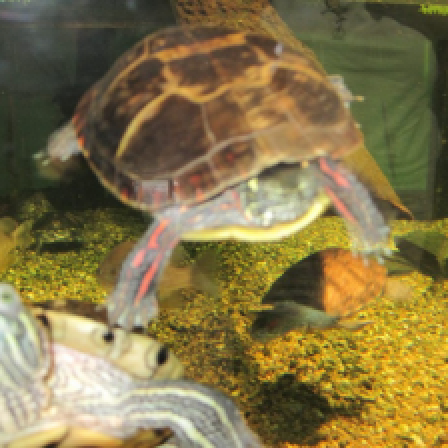} &
\includegraphics[width=.18\linewidth]{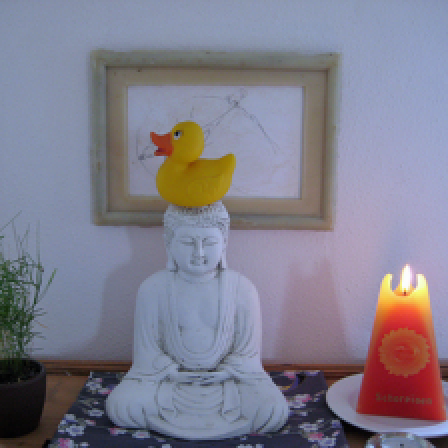}\\[2pt]

\end{tabular}

\caption{(Continued) Additional visual comparison.}
\end{figure*}

% Page 3: ResNet50 part 1 (6 rows with short caption)
\begin{figure*}[!t]
\ContinuedFloat
\centering
\setlength\tabcolsep{1pt}

\begin{tabular}{@{}c@{}ccccc@{}}
& wolf spider & reflex camera & pill bottle & terrapin & candle \\
\rotatebox{90}{$\text{FEI}_{\text{IBM}}$} &
\includegraphics[width=.18\linewidth]{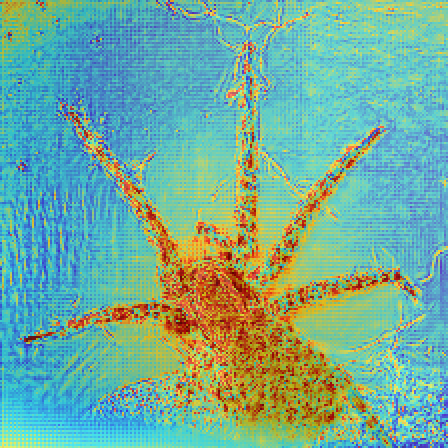} &
\includegraphics[width=.18\linewidth]{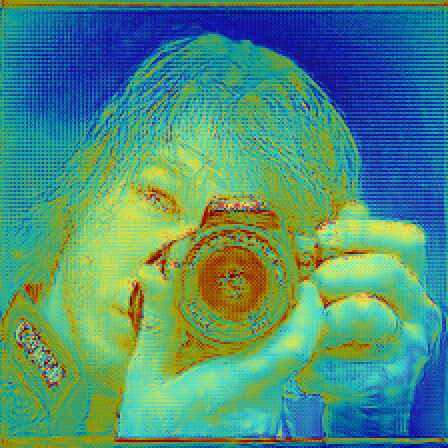} &
\includegraphics[width=.18\linewidth]{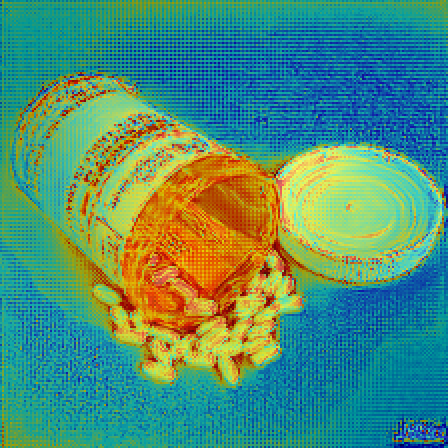} &
\includegraphics[width=.18\linewidth]{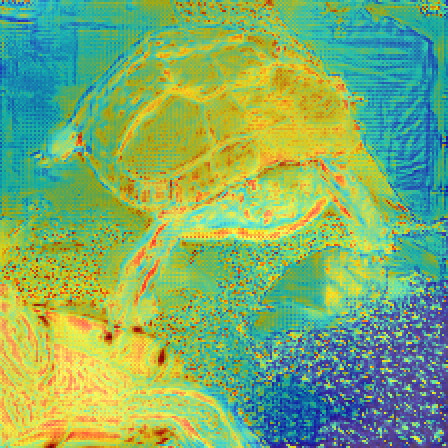} &
\includegraphics[width=.18\linewidth]{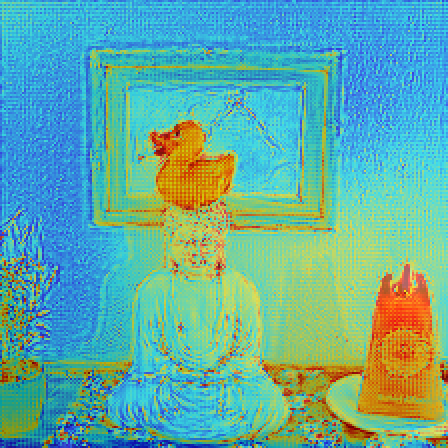} \\[2pt]

\rotatebox{90}{$\text{FEI}_{\text{IVM}}$} &
\includegraphics[width=.18\linewidth]{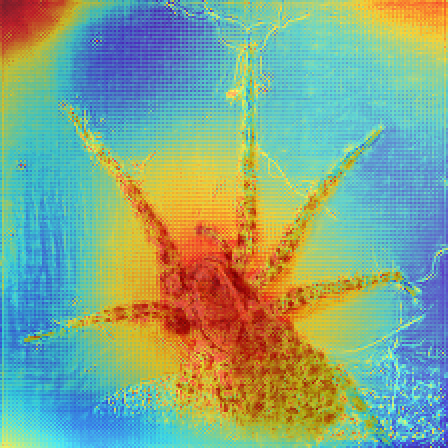} &
\includegraphics[width=.18\linewidth]{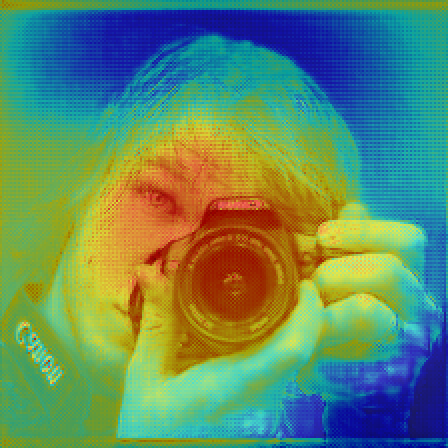} &
\includegraphics[width=.18\linewidth]{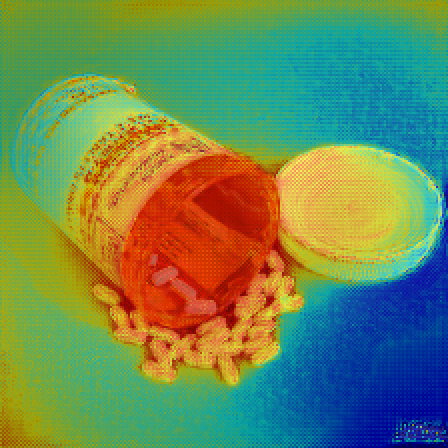} &
\includegraphics[width=.18\linewidth]{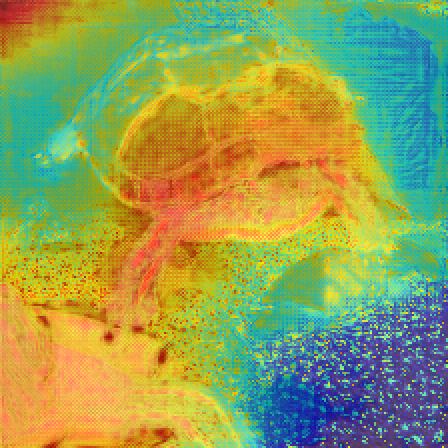} &
\includegraphics[width=.18\linewidth]{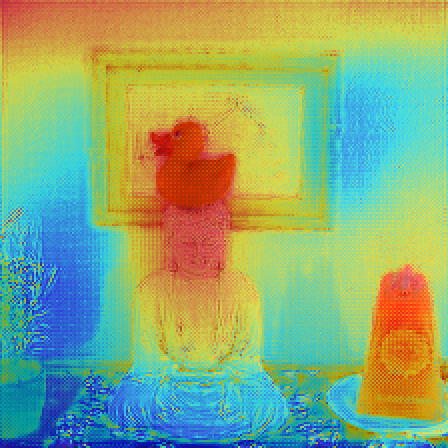} \\[2pt]

\rotatebox{90}{$\text{FEI}_{\text{AVM}}$} &
\includegraphics[width=.18\linewidth]{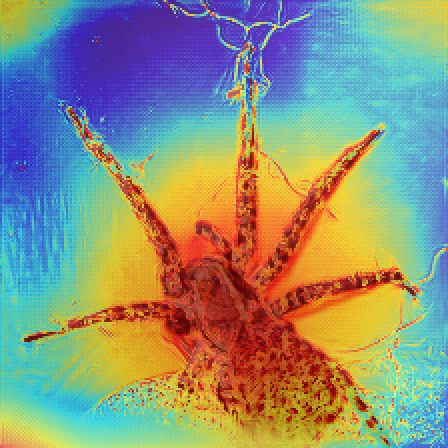} &
\includegraphics[width=.18\linewidth]{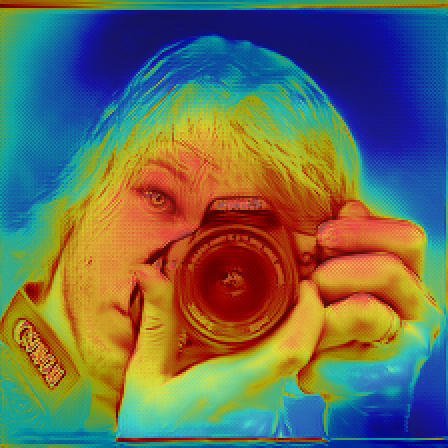} &
\includegraphics[width=.18\linewidth]{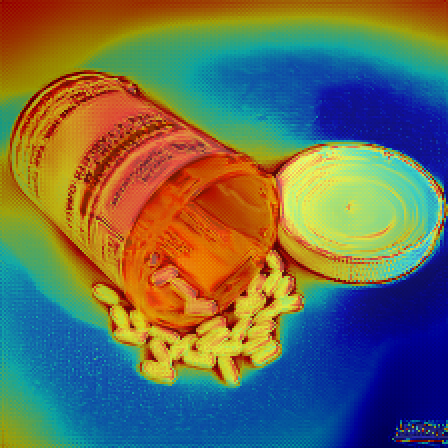} &
\includegraphics[width=.18\linewidth]{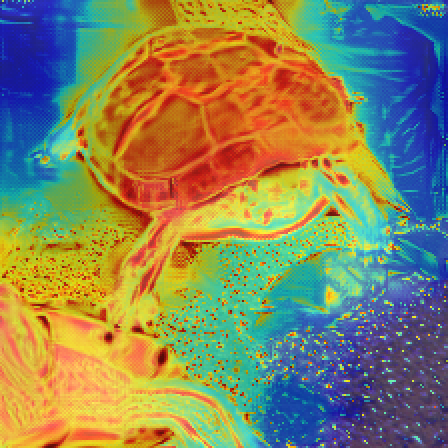} &
\includegraphics[width=.18\linewidth]{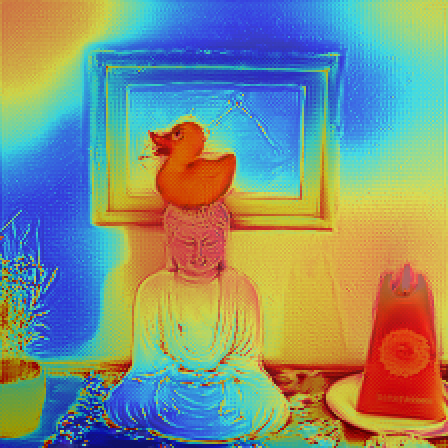} \\[2pt]

\rotatebox{90}{$\text{FEI}_{\text{VM}}$} &
\includegraphics[width=.18\linewidth]{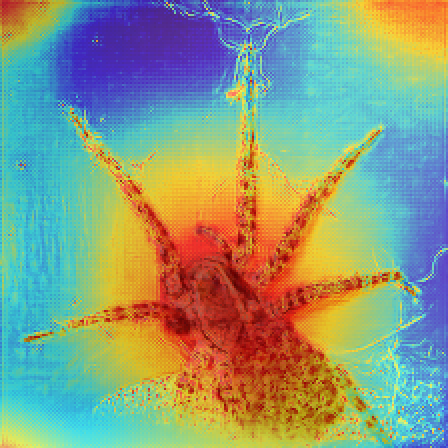} &
\includegraphics[width=.18\linewidth]{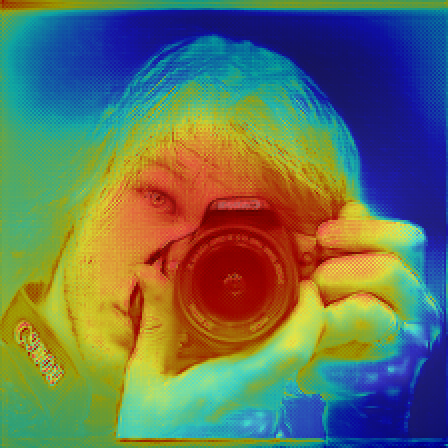} &
\includegraphics[width=.18\linewidth]{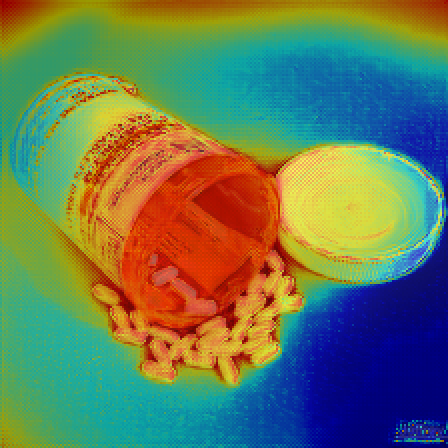} &
\includegraphics[width=.18\linewidth]{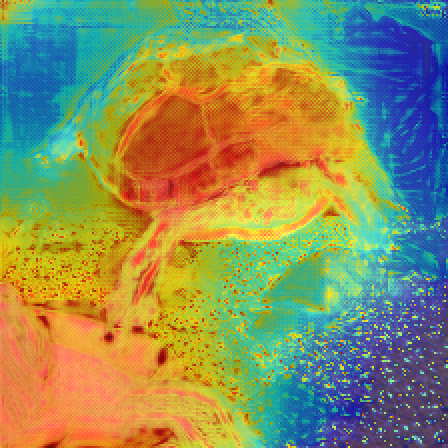} &
\includegraphics[width=.18\linewidth]{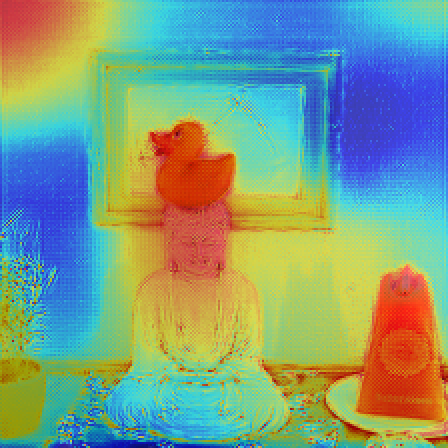} \\[2pt]

\rotatebox{90}{Unconstr. EQO} &
\includegraphics[width=.18\linewidth]{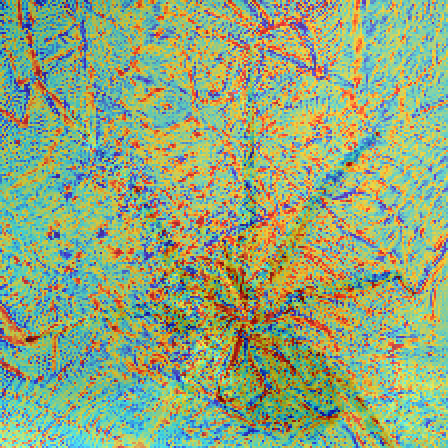} &
\includegraphics[width=.18\linewidth]{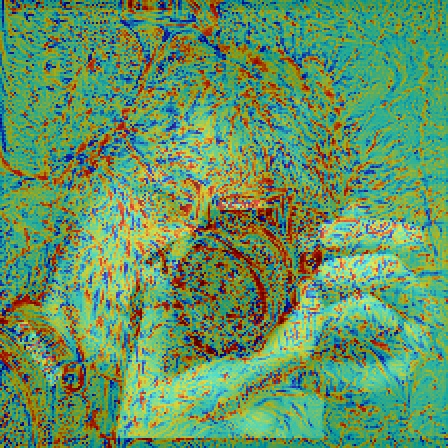} &
\includegraphics[width=.18\linewidth]{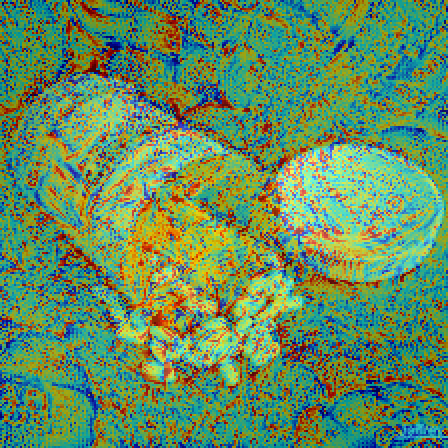} &
\includegraphics[width=.18\linewidth]{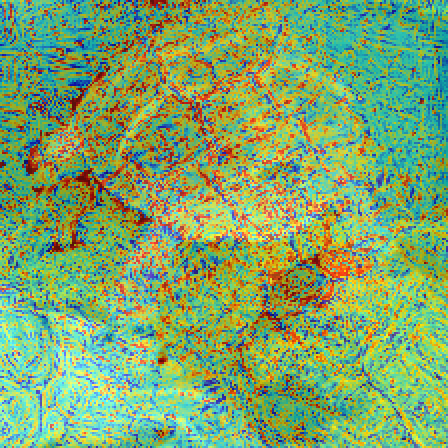} &
\includegraphics[width=.18\linewidth]{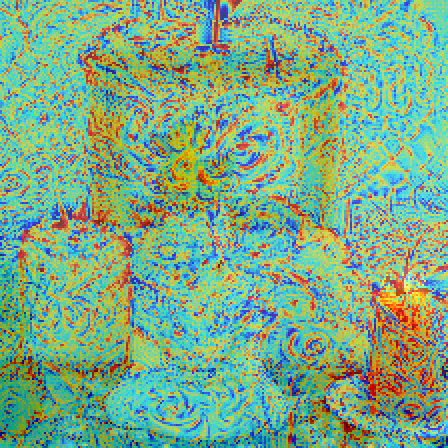} \\[2pt]

\rotatebox{90}{EP} &
\includegraphics[width=.18\linewidth]{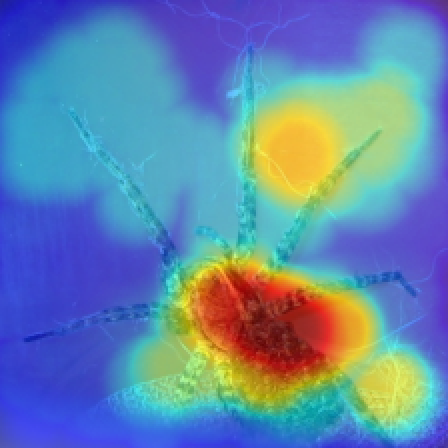} &
\includegraphics[width=.18\linewidth]{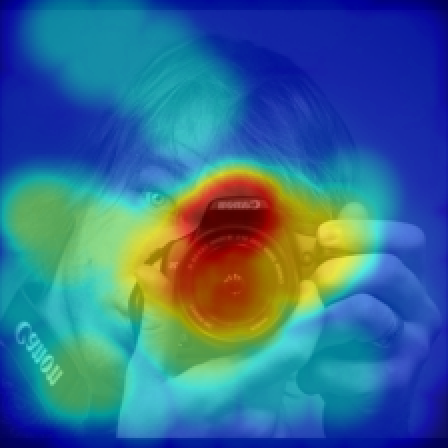} &
\includegraphics[width=.18\linewidth]{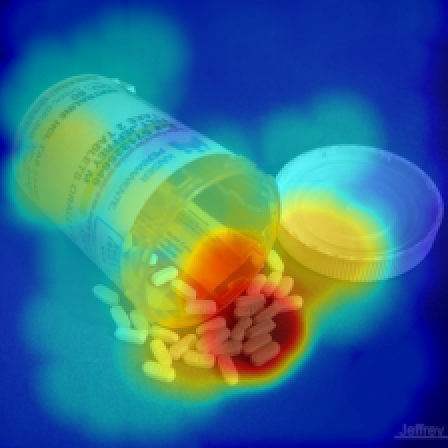} &
\includegraphics[width=.18\linewidth]{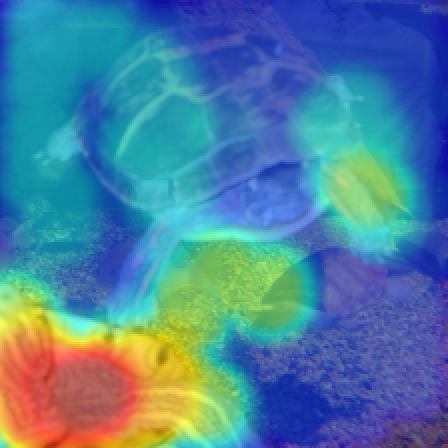} &
\includegraphics[width=.18\linewidth]{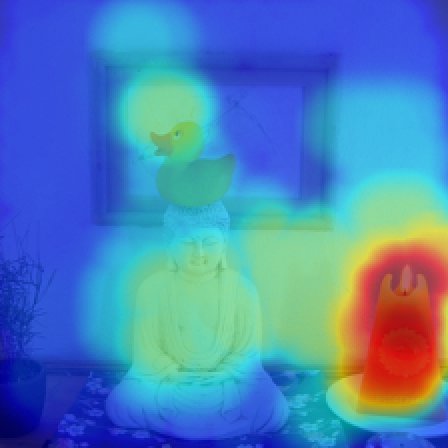} \\
\end{tabular}

\caption{(Continued) Additional visual comparison.}
\end{figure*}

% Page 4: ResNet50 part 2 (6 rows with short caption)
\begin{figure*}[!t]
\ContinuedFloat
\centering
\setlength\tabcolsep{1pt}

\begin{tabular}{@{}c@{}ccccc@{}}
& wolf spider & reflex camera & pill bottle & terrapin & candle \\
\rotatebox{90}{RISE} &
\includegraphics[width=.18\linewidth]{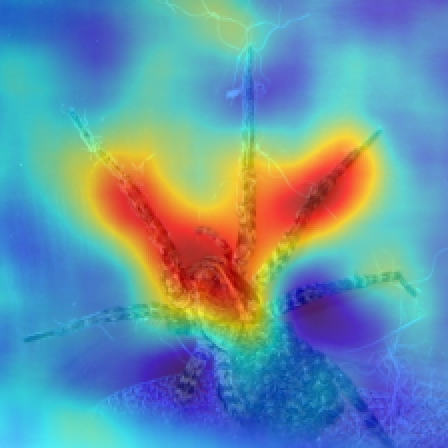} &
\includegraphics[width=.18\linewidth]{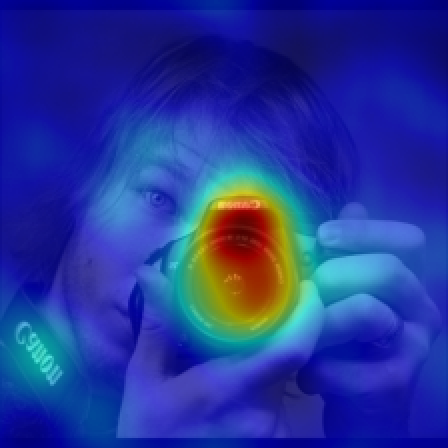} &
\includegraphics[width=.18\linewidth]{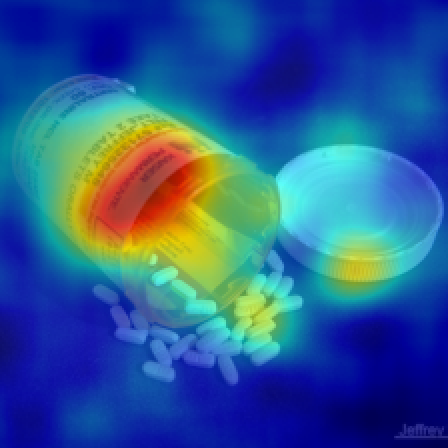} &
\includegraphics[width=.18\linewidth]{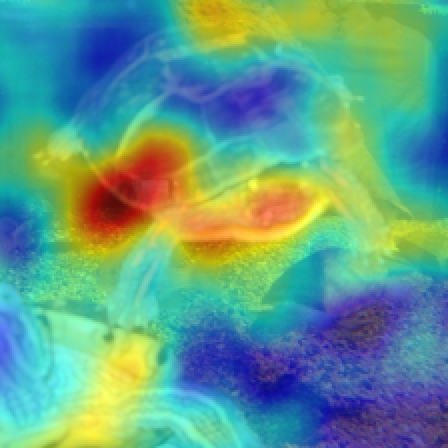} &
\includegraphics[width=.18\linewidth]{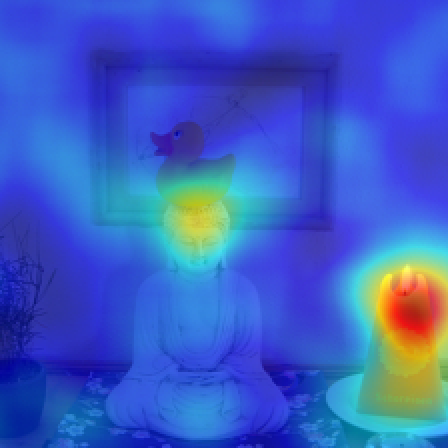} \\[2pt]

\rotatebox{90}{IIA} &
\includegraphics[width=.18\linewidth]{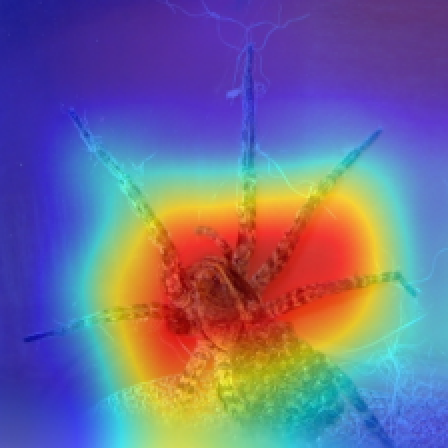} &
\includegraphics[width=.18\linewidth]{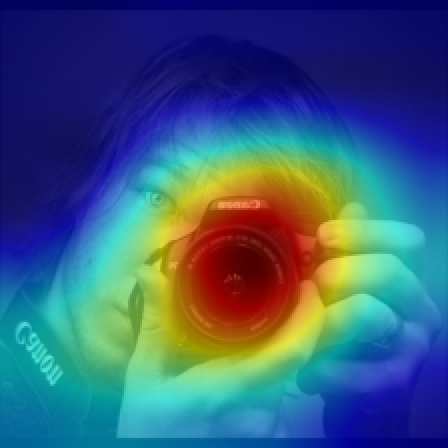} &
\includegraphics[width=.18\linewidth]{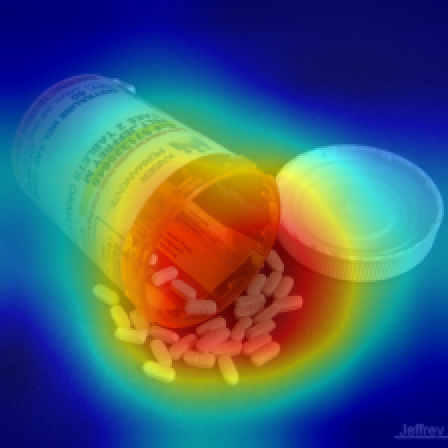} &
\includegraphics[width=.18\linewidth]{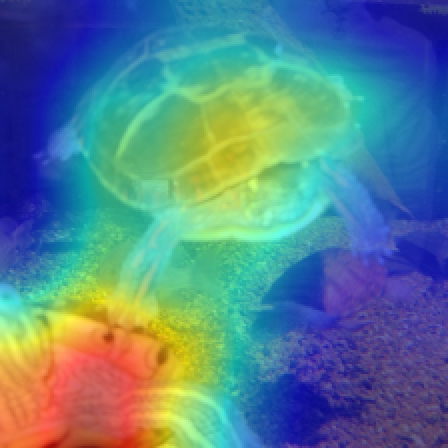} &
\includegraphics[width=.18\linewidth]{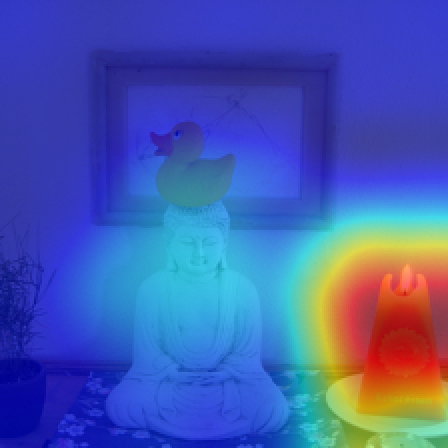} \\[2pt]

\rotatebox{90}{Lift-CAM} &
\includegraphics[width=.18\linewidth]{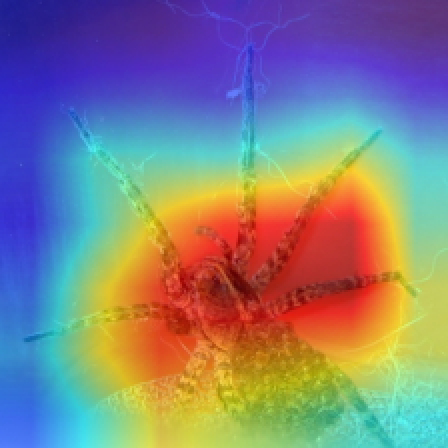} &
\includegraphics[width=.18\linewidth]{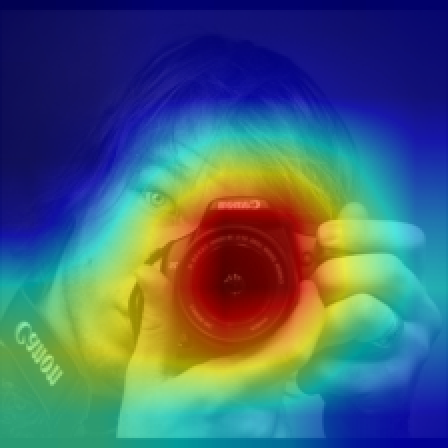} &
\includegraphics[width=.18\linewidth]{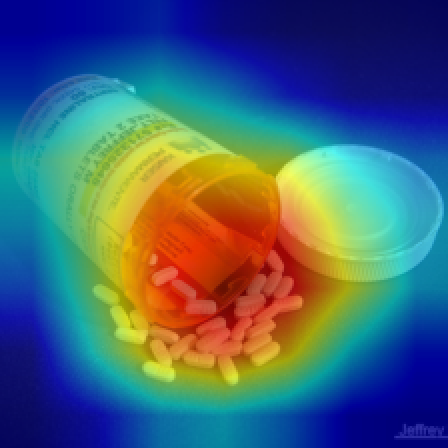} &
\includegraphics[width=.18\linewidth]{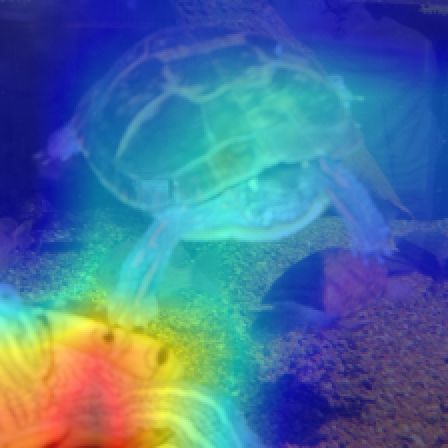} &
\includegraphics[width=.18\linewidth]{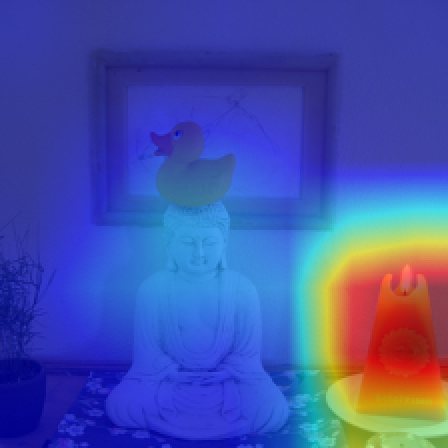} \\[2pt]

\rotatebox{90}{FG-VCE} &
\includegraphics[width=.18\linewidth]{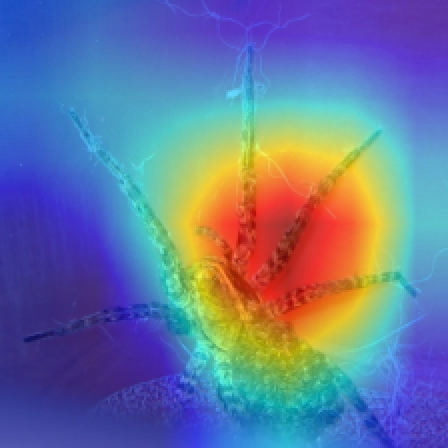} &
\includegraphics[width=.18\linewidth]{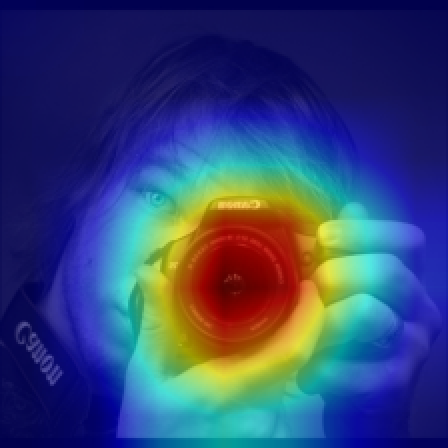} &
\includegraphics[width=.18\linewidth]{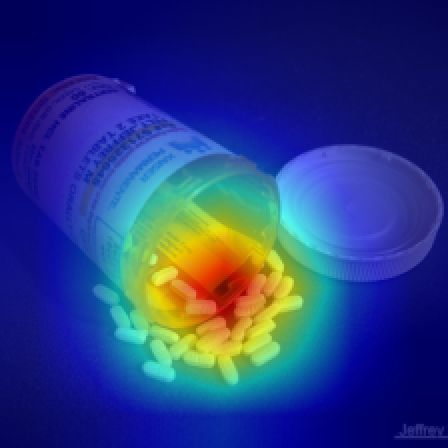} &
\includegraphics[width=.18\linewidth]{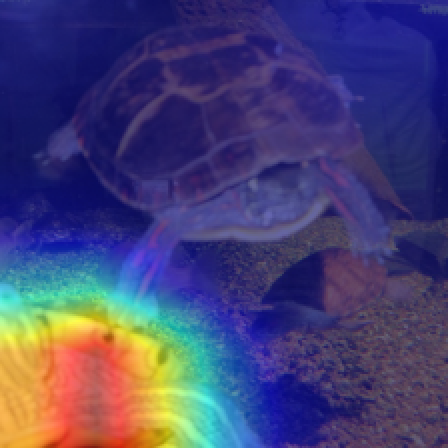} &
\includegraphics[width=.18\linewidth]{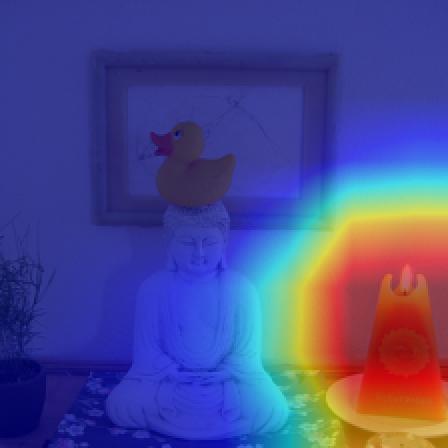} \\
\end{tabular}

\caption{(Continued) Additional visual comparison.}
\end{figure*}

\section{Additional Qualitative Comparisons}
\label{app:qual}
\label{sec:app_qualitative}
\Cref{fig:supp_additional_vis} shows additional attribution visualizations
across ImageNet samples on VGG16 and ResNet50.  These examples illustrate the
pattern in \cref{fig:motivation}: unconstrained EQO is fragmented, while the
constrained FEI variants reduce large-scale fragmentation on the displayed
images while retaining fine-scale pixel texture.

\begin{figure*}[t]
\centering
\begin{subfigure}[t]{0.98\textwidth}
\includegraphics[width=\linewidth]{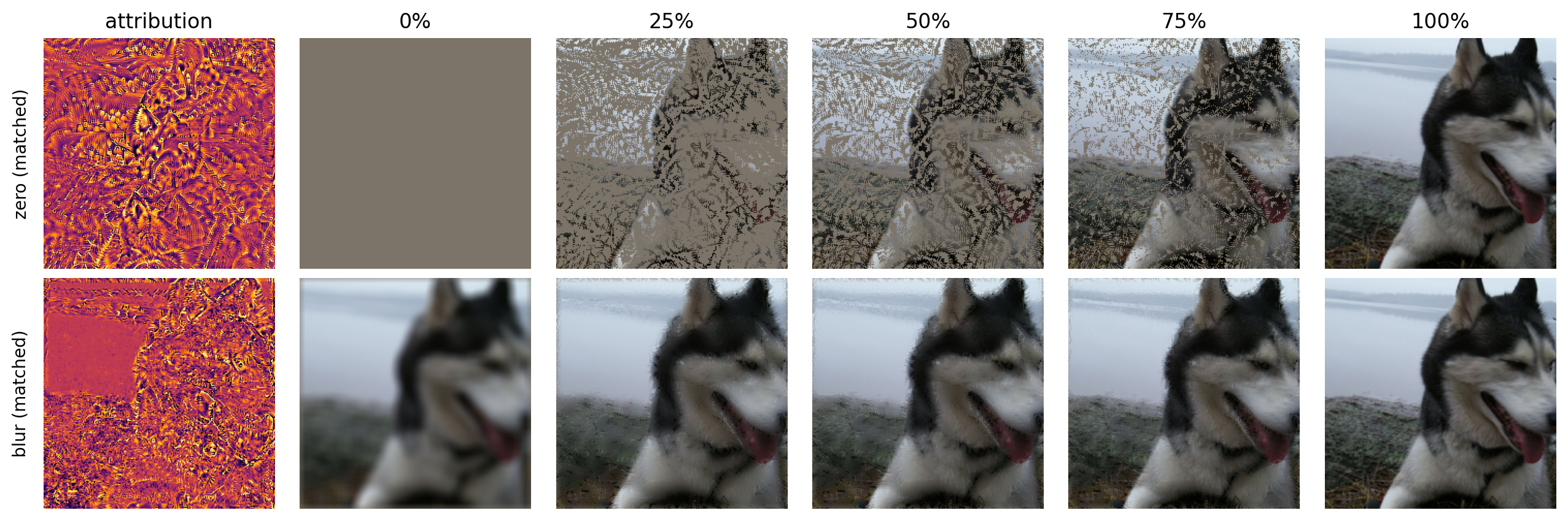}
\caption{Unconstrained EQO: matched zero-reference (top) and
blur-reference (bottom) insertion sequences.}
\end{subfigure}

\begin{subfigure}[t]{0.98\textwidth}
\includegraphics[width=\linewidth]{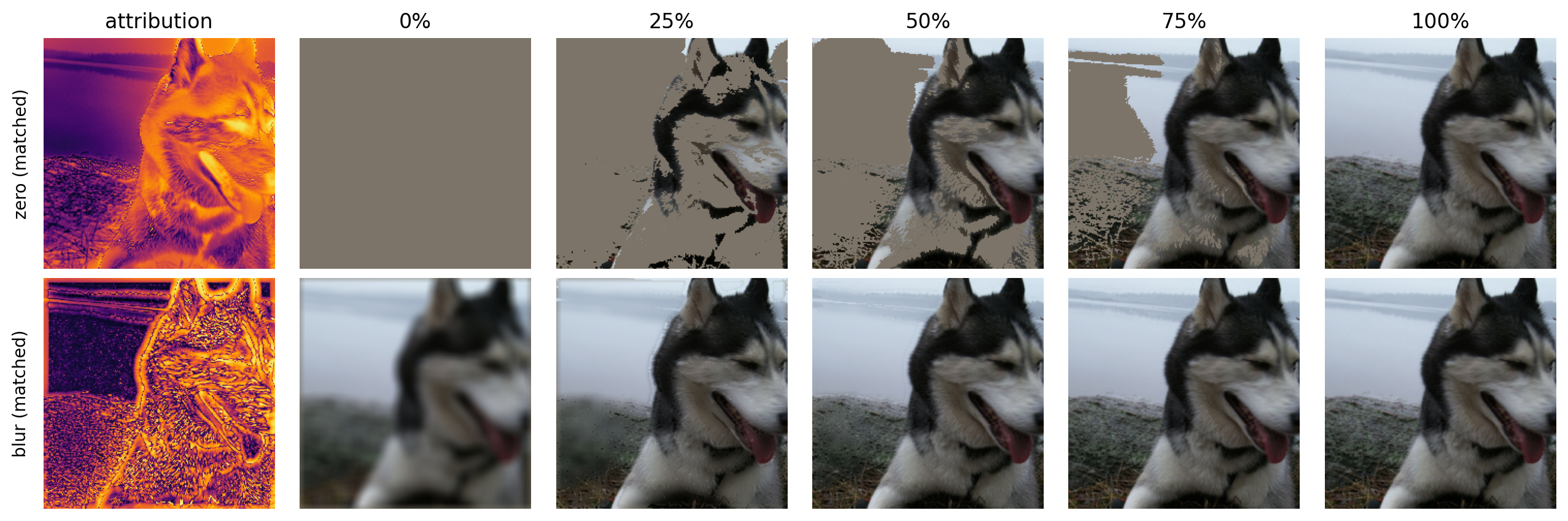}
\caption{FEI$_{\mathrm{VM}}$ under the same matched protocols.}
\end{subfigure}

\begin{subfigure}[t]{0.62\textwidth}
\includegraphics[width=\linewidth]{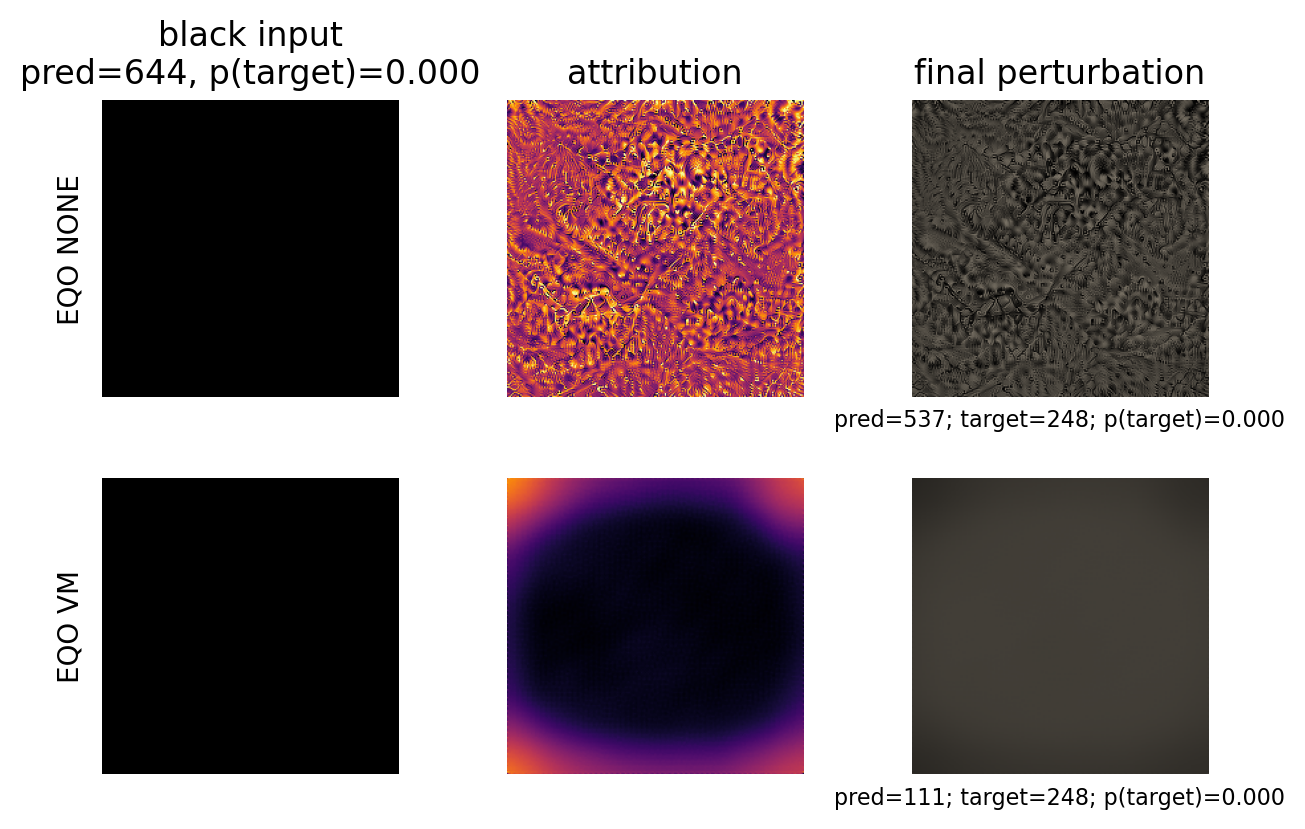}
\caption{Separate all-black-input sanity example.}
\end{subfigure}
\caption{Matched zero- and blur-reference insertion sequences with a separate
all-black-input check. Normalized zero decodes as gray; percentages denote
retained area.}
\label{fig:reference_examples}
\end{figure*}

\Cref{fig:reference_examples} provides the corresponding insertion sequences
for the matched zero- and blur-reference experiment in \cref{app:reference}.

\end{document}